%% file: main.tex
\documentclass[a4paper,12pt,notitlepage]{report}
\usepackage[left=1in, right=1in, top=1in, bottom=1in,footskip=.25in]{geometry}
\usepackage{times}

\usepackage{graphicx}
\usepackage{hyperref}
\usepackage{color}
\usepackage{subcaption}  %
\usepackage{xspace}

\usepackage{wrapfig}

\usepackage{tabularx}
\usepackage{booktabs}
\usepackage{multirow}

\newcolumntype{Y}{>{\centering\arraybackslash}X}

\usepackage{float}
\usepackage{multicol}

\usepackage[svgnames]{xcolor}

\usepackage{comment}

\usepackage[
backend=biber,
natbib=true,
style=alphabetic,
citestyle=numeric,
sorting=nty,
maxbibnames=99,
maxcitenames=1,
mincitenames=1,
bibstyle=ieee,
dateabbrev=false
]{biblatex}
\usepackage{tikz}
\usetikzlibrary{positioning,arrows.meta,fit,calc,backgrounds,decorations.pathmorphing}
\usetikzlibrary{shapes.geometric}
\usepackage{mathtools}
\input{math_commands}

\usepackage{algorithm}
\usepackage{algpseudocode}
\usepackage{amsmath}

\usepackage[export]{adjustbox}

\usepackage{setspace}

\usepackage[textsize=scriptsize, disable]{todonotes}

\usepackage{xspace}

\input{macro}

\begin{document}
\onehalfspacing

\pagenumbering{roman}

\definecolor{ttic_blue}{RGB}{55,93,137}
\begin{tikzpicture}[overlay,remember picture]
    \draw [line width=1mm] [ttic_blue]
        ($ (current page.north west) + (0.9in,-0.9in) $)
        rectangle
        ($ (current page.south east) + (-0.9in,0.9in) $);
\end{tikzpicture}

\begin{center}
\Large \MakeUppercase{\textbf{Generative Modeling Perspective}}\\
\Large \MakeUppercase{\textbf{for Control and Reasoning}}\\
\Large \MakeUppercase{\textbf{in Robotics}}\\
\normalsize
\vspace{0.25in}
by\\Takuma Yoneda\\
\vspace{1in}
A thesis submitted\\
in partial fulfillment of the requirements for\\
the degree of\\
\vspace{0.3in}
Doctor of Philosophy in Computer Science\\
\vspace{0.3in}
at the\\
\vspace{0.3in}
\MakeUppercase{Toyota Technological Institute at Chicago}\\
Chicago, Illinois\\
\vspace{0.3in}
September, 2024\\
\vspace{1.0in}
Thesis Committee:\\
Matthew R.\ Walter (Thesis Advisor)\\
Greg Shakhnarovich\\
Bradly C. Stadie \\
\end{center}
\thispagestyle{empty} \clearpage

\begin{abstract}\addcontentsline{toc}{chapter}{\abstractname}
Heralded by 
the initial success in speech recognition and image classification, learning-based approaches with neural networks, commonly referred to as deep learning, have spread across various fields.
A primitive form of a neural network functions as a deterministic mapping from one vector to another, parameterized by trainable weights.
This is well suited for point estimation in which the model learns a one-to-one mapping (e.g., mapping a front camera view to a steering angle) that is required to solve the task of interest.
Although learning such a deterministic, one-to-one mapping is effective,
there are scenarios where modeling \emph{multimodal} data distributions, namely learning one-to-many relationships, is helpful or even necessary.

In this thesis, we adopt a generative modeling perspective on robotics problems. Generative models learn and produce samples from multimodal distributions, rather than performing point estimation. We will explore the advantages this perspective offers for three topics in robotics.

The first topic is intuitive physics. Humans gain intuitive understanding of various properties of physics by interacting with and observing the world. Such intuition allows them to reason over safe and efficient handling of physical objects, without the need for explicit calculations.
Intuitive physics is a vague concept that encompasses different aspects.
As commonly used in previous works, we choose a stack of toy blocks as a subject.
While the majority of prior works predict stability scores of the stacks or block poses after collapse (i.e., forward dynamics), %
our work aims to obtain the notion of stability by learning joint distributions of block poses that compose stable stacks. 
After training, the model enables us to generate various block compositions that form stable structures, without the need for physics simulation.

Subsequently, we look at a long-studied domain of shared autonomy---a concept where an assistive agent helps a user operate on some system.
A key challenge is determining how to assist the user without explicit knowledge of their intent or goal.
Recent approaches train an assistive policy with model-free reinforcement learning (RL). It comes with a significant benefit that we do not need the knowledge of the environment dynamics, however, their formulation assumes that the assistive policy is paired with a user policy during training, rendering it an inefficient human-in-the-loop pipeline.
Instead, in our work, we train our model solely from task demonstrations. Diffusion models allow us to directly learn multi-modal action distribution where each mode loosely represents user's intent.
During inference, we use a modified reverse diffusion process to sample from a proper mode given user's original action.

Finally, we address the sim-to-real domain gap for visuo-motor control tasks.
Majority of learning-based methods for control policies are trained in simulation, preferring its efficiency and reproducibility (i.e., recreating a certain setup).
Direct deployment of simulation-trained models on real robots often results in performance degradation due to the gap between simulated and real environments.
When training a new model from scratch in the real environment is not a choice, a better alternative would be to adapt (i.e., fine-tune) the model that already performs well in simulation. We introduce a novel approach to unsupervised domain adaptation based on adversarial training objective.

\end{abstract}

\chapter*{Acknowledgements}
\addcontentsline{toc}{chapter}{Acknowledgements}

Words cannot adequately express my gratitude to my advisor Dr. Matthew R. Walter.
It was your warm and consistent support that kept me moving forward through difficult moments.
I'm especially grateful that you've always been hands-on and committed to understanding the details, contributing to projects at every level from discussing seed ideas and planning experiments to writing papers. Getting your input, digesting it, and learning from it has helped me grow as a researcher and I feel incredibly fortunate to have been under your guidance.

Throughout the projects, I also had a fantastic time getting advice from Dr. Bradly C. Stadie and Dr. Ge Yang.
I learned a lot of practical skills like how to organize experiments, write code, and approach paper writing.
I met you folks at the end of my second year when I was quite lost regarding research directions, without having experience with solid projects before.
Your strong guidance with micro-management and frequent communication was extremely helpful for me to pick up various skills and build confidence.

I was also very fortunate to be surrounded by wonderful peers and friends.
Onishi Takeshi san has been extremely kind to me even before I joined the institute, ever since I got interested in TTIC. Your warm support and advice that span from life in the States and connection with Japanese community to academic research have been crucial to survive here. 
I have also enjoyed hanging out with Kurihana Takuya. As a peer student, it was great that I can chat with you for random topics, consult various things, and have fun exploring the city together. Not to mention, this journey would not have been possible without cheerful, warm and creative-minded folks in our lab, Falcon Dai, Chip Schaff, Andrea F. Daniele, Shengjie Lin, Jiading Fang, David Yunis, Luzhe Sun, Tianchong Jiang, Teddy Ayalew, Davide Iafrate, Kevin Wu, Sam Wheeler, Ben Picker, and folks who stayed in our lab for short terms. Thank you so much for developing and maintaining a wonderful atmosphere in the lab to cultivate research and creativity.

I would also like to show my deep gratitude to professors and admins at Toyota Technological Institute in Nagoya (also known as TTIJ),
Miwa Makoto sensei, Sasaki Yutaka sensei, Sakaki Hiroyuki sensei, Ukita Norimichi sensei,
Hisamoto Masashi san, Masuda Yoshihiko san, and Hotate Kazuo sensei.
I have kept feeling like I am still part of the TTIJ community thanks to your warm welcome whenever I visit the institute,
your kind words to encourage me and your willingness to keep in touch. I definitely felt less lonely studying here because of your kind support.

I'd also like to express my sincere gratitude to Toyoda Tatsuro sensei, the founder of this very institute, for providing this fantastic educational environment to pursue a PhD as well as supporting me for my first year through your scholarship. Also, having conversations with Ayako-sama and Yumiko-sama was a precious experience for me. I truly appreciate your warm support and encouragement.

Finally, I thank you to my parents, sister, Dankichi (our cat) and the rest of my family for your unconditional love and support. It would have been impossible to come and study here all the way from the other side of the planet without your deep trust in me, warm support, and unwavering optimism.

\clearpage

\tableofcontents
\listoftables
\listoffigures

\pagenumbering{arabic}

\input{introduction}

\input{preliminary}

\input{block-stacking-with-diffusion/z_main}

\input{diffusha/z_main}

\input{ila/z_main}

\printbibliography

\end{document}

%% file: math_commands.tex
\usepackage{amsmath,amsfonts,bm}

\def\eqref#1{equation~\ref{#1}}

\def\1{\bm{1}}

\def\eps{{\epsilon}}

\DeclareMathAlphabet{\mathsfit}{\encodingdefault}{\sfdefault}{m}{sl}
\SetMathAlphabet{\mathsfit}{bold}{\encodingdefault}{\sfdefault}{bx}{n}

\DeclareMathOperator*{\argmin}{arg\,min}

%% file: macro.tex
\definecolor{mydarkblue}{rgb}{0,0.08,0.45}
\definecolor{es-blue}{rgb}{0,0.4,0.8}
\definecolor{darkgreen}{rgb}{0.0, 0.5, 0.0}
\definecolor{arsenic}{rgb}{0.23, 0.27, 0.29}
\newcommand{\srceq}[1]{\mathcolor{purple}{#1}}
\newcommand{\tgteq}[1]{\mathcolor{blue!60}{#1}}

\newcommand{\lcila}{invariance through latent alignment}
\newcommand{\src}{\textrm{src}}
\newcommand{\csrc}{\text{\color{purple}src}}
\newcommand{\tgt}{\textrm{tgt}}
\newcommand{\ctgt}{\text{\color{blue!60}tgt}}

\newcommand{\sw}{\textrm{sw}}
\newcommand{\fwr}{\gamma}

\newcommand{\bx}{\bm{x}}
\newcommand{\bxt}{\bm{x}_t}
\newcommand{\bxT}{\bm{x}_T}
\newcommand{\bxn}{\bm{x}_0}
\newcommand{\bxtm}{\bm{x}_{t-1}}
\newcommand{\bz}{\bm{z}}
\newcommand{\beps}{\bm{\eps}}

\newcommand{\bp}{\bm{p}}

%% file: introduction.tex
\chapter{Introduction}

Recent progress in (artificial) neural networks has brought tremendous benefits to the robotics community.
The primitive form of neural networks, multi-layer perceptron (MLP), is a simple stack of learnable linear maps interleaved with nonlinear activation functions such as sigmoid.
By design, an MLP maps a single input vector to an output vector \emph{deterministically}, parameterized by its weights.

This property aligns well with a training dataset whose underlying structure represents a one-to-one mapping from inputs to outputs.
Assuming that the output space is continuous, a common training strategy of an MLP for such a dataset is to compute a mean-squared error (MSE) between the predicted and the desired output, and then update the network weights in a direction that reduces the error. Iterating this process over a large number of data samples gradually pushes the weights to the point where the MLP achieves a small MSE.

This strategy faces a challenge, however, if the dataset exhibits a \emph{one-to-many} relationship between inputs and outputs. This means there are multiple, potentially contradictory desired outputs associated with a single input. Although such scenarios may seem unnatural, they are quite common in practice.
As an example, let us consider a driving task where the agent predicts a steering angle at each state, while the vehicle moves forward at a constant speed. A dataset for this task might come from expert demonstrations, containing pairs of states and desired steering angles. In a state where an obstacle is directly ahead of the vehicle, the corresponding desired steering angle may not be unique: one expert might dodge it by steering \emph{left}, while another might choose to go \emph{right}. 
This gives rise to the aforementioned one-to-many relationship, where a single state has two opposite desired steering values. 
Training an MLP on this dataset with MSE as described above would result in the model fitting the average over these opposite values (i.e., heading straight into the obstacle!), which was never a desired output in the demonstrations.
To address this issue, we aim for the model to learn a \emph{multimodal} distribution and pick from the two modes, \emph{left} or \emph{right}, in such ambiguous states.

We employ generative models throughout the thesis to achieve this, especially diffusion models and generative adversarial networks (GANs). These models are capable of modeling a multimodal distribution as well as generating samples from it.
The ability to generate a diverse set of samples also
offers practical benefits over point estimates.
When the model is integrated into a pipeline,
downstream modules do not need to commit to a single predicted sample; rather, it can potentially consider multiple candidate samples and pick the one that is suitable based on additional preferred metrics.
As a practical example, in chapter \ref{chap:block-stacking-with-diffusion}, we consider the task of generating a stable block composition that matches a given silhouette.
We will see that our generative model can produce a diverse set of candidate compositions for a single silhouette, potentially offering users the flexibility to reject those candidates that are difficult to construct.

The main topic throughout the thesis is to have a generative modeling perspective on robotics problems, and discuss the benefits and unique characteristics it offers.
After preliminary material~(chapter \ref{chap:preliminary}), in chapter \ref{chap:block-stacking-with-diffusion} and \ref{chap:diffusha}, we present our works that leverage the characteristics of diffusion models for each robotics task setting, namely block stacking and shared autonomy, and describes the unique benefits brought by the capability of modeling and sampling from multimodal distributions. 
In chapter \ref{chap:ila}, we discuss an approach to leverage adversarial training objective for bridging gaps in two different visuo-motor control domains.

\section{Contributions}

In chapter \ref{chap:block-stacking-with-diffusion}, we develop an approach that aims to learn a specific aspect of \emph{intuitive physics}, especially the notion of stability for a stack of toy blocks. 
Previous works on the same subject predict a stability score of a stack or the object positions in the future timestep after it collapses. 
Instead, our work takes a novel approach by learning a joint distribution over block poses that form stable structures.
Using the silhouette of a stack as a conditional information, the model learns to generate various stable block compositions that align with the reference silhouette.

Chapter \ref{chap:diffusha} discusses shared autonomy:
a concept in which a user controls a system with the help of an assistive agent. Many  recent learning-based approaches for shared autonomy rely on an expensive human-in-the-loop training pipeline.
To circumvent this, we propose a new approach leveraging diffusion models that only requires task demonstrations for training. 
A critical challenge in shared autonomy is assisting the user while facing uncertainty about their goals or intentions.
Diffusion models enable us to naturally learn multimodal action distributions across various intentions.
To sample from an appropriate mode that reflects the user's intent during inference, we develop a modified reverse diffusion process. This process interpolates between the user's action distributions and those of the demonstrations.

In chapter \ref{chap:ila}, we shift our focus to bridging the domain gap using a distribution matching objective inspired by GANs.
We consider a scenario where we have two visuo-motor control domains: \emph{source} and \emph{target}. While these domains share tasks, their (pixel-based) observations differ.
Given an agent that performs well in the \emph{source} domain, we propose a novel approach to unsupervised domain adaptation that aims to make the agent work equally well in the \emph{target} domain.

\section{Background}
In this section, I would like to give a brief overview and short history of the field from my perspective,
especially focusing on learning-based approaches in robot control and reasoning. Please note that this is not meant for a comprehensive review of the field, rather it may be largely biased by my personal experience and interests.

Although neural networks and related approaches have been studied from very early, the field of deep learning gained explosive popularity from 2012 when AlexNet~\citep{alexnet} won a major image classification competition, ILSVRC2012~\citep{ILSVRC15}, by a large margin.
In 2013, Deep Q-network (DQN)~\citep{dqn} showed a significant result that a model-free RL with neural networks has the potential to solve various games in Atari 2600 Games from the raw pixel inputs. After DQN, application of deep learning in RL, referred to as deep RL, became widely popular in the field.
Atari games became a popular benchmark for testing the performance of deep RL methods using pixel inputs. 
Researchers explored various approaches with model-free~\citep{rainbow, impala, R2D2} and model-based RL~\citep{simple, dreamer-v2, muzero} on Atari.
Agent57~\citep{agent57} achieved a score that is above the human baseline on all 57 games. Deep RL has shown significant success on other video games \citep{Vinyals2019GrandmasterLI, dota2, DBLP:conf/iclr/JaderbergMCSLSK17, doi:10.1126/science.aau6249} and table games like Go~\citep{alphago, alphago-zero}, Shogi~\citep{alphazero, muzero}, Poker~\citep{NEURIPS2020_c61f571d} and Hanabi~\citep{Lerer2019ImprovingPV}.

Deep RL has been also applied in motor control tasks, environments built on MuJoCo simulator~\citep{mujoco}, notably DeepMind Control suite~\citep{deepmindcontrolsuite2018} provides a benchmark for various dynamic control tasks in simulation, and it has served as a go-to benchmark for RL methods with continuous action space. 
Around this time, a lot of the ``robotics'' works using deep RL were trained and evaluated only within simulation and oftentimes did not have accompanying experiments in the real environment. This is because the sample complexity (loosely equivalent to the amount of required data) of deep RL approaches were so high that training in the real environment was not practical.
For the papers that tackle real environments, the models are first trained in simulation and then often fine-tuned in the real environment with a relatively small amount of environment interactions.
Some of the notable results in deploying a deep RL agent in the real world include legged locomotion~\citep{RMA, doi:10.1126/scirobotics.aau5872, doi:10.1126/scirobotics.abc5986, margolisyang2022rapid, DBLP:conf/rss/HaarnojaHZTTL19} and dexterous manipulation~\citep{pmlr-v87-kalashnikov18a, doi:10.1177/0278364917710318, Andrychowicz2018LearningDI}.

Although these works demonstrated surprising effectiveness in specific domains, the training instability and unexpected behaviors of the policy when facing unseen scenarios made them less practical for application in new domains. As a result, deep RL did not remain a mainstream topic for real-robot control.

Instead, in the past few years, there have been increasing interest in imitation learning, which aims to learn behaviors from demonstrations without exploration by the agent nor pre-defined environment rewards. The demonstrations are collected by experts teleoperating the robot to perform a task.
In imitation learning, the aforementioned ``one-to-many mapping'' issue in the datasets is even more pronounced. This is because when experts teleoperate a robot, say approaching to a specific object to grasp, it is very common that different attempts do not result in the exact same trajectories. This gives rise to the scenario where a single state is annotated with different actions. Thus, recent approaches use generative models such as diffusion models and variational auto-encoders to learn a policy \citep{anonymous2023imitating, diffusion-policy, Zhao-RSS-23}, or discretize the action space to handle multimodal action distribution~\citep{behavior-transformer, pmlr-v235-lee24y}.

In imitation learning, a policy typically aims to learn a single or just a handful of tasks at once. Given the success of large language models (LLMs), showing their extremely general capability to understand and generate text on diverse set of topics, there is a growing interest in the robotics community to build an embodied equivalent to LLMs that can perform diverse set of tasks in unstructured environments. 
The economic impact such an embodied agent brings would easily surpass that of LLMs. Thus industry has also been attracted to this concept of embodied large action models, leading to a tremendous number of startups launched in the past few years related to this topic. 
However, we have not seen clear signs of success that matches what we saw in LLMs yet.

\clearpage

%% file: preliminary.tex
\chapter{Preliminary}
\label{chap:preliminary}

In this section, we briefly review key concepts assumed throughout the thesis. We cover Markov decision processes (MDPs), which represent interactions between an agent and an environment, as well as two families of generative models: generative adversarial networks (GANs) and diffusion models.

\section{Markov Decision Process and Reinforcement Learning}
Markov Decision Processes (MDPs) are a classical formalization of sequential decision making.
It serves as a straightforward framing of the learning problem where a decision maker called \emph{agent} interacts with an \emph{environment}, an entity comprising everything outside the agent.

As the agent takes an \emph{action} $a_t \in \mathcal{A}$ in the environment, the environment presents a new \emph{state} $s_{t+1} \in \mathcal{S}$ that describes the current situation, as well as a \emph{reward} $r_{t+1} \in \mathbb{R}$ that the agent seeks to maximize over the course of actions. %
Controlling a robot arm on a tabletop as an example, the agent's action may be joint torques, while state may contain joint angles, velocities, and information of the surroundings such as object locations and their shapes.
The continuous interactions between the environment and the agent gives rise to a \emph{trajectory} $\tau$
\[
\tau \coloneqq (s_0, a_0, r_1, s_1, a_1, r_2, s_2, a_2, r_3, \ldots)
\]

The \emph{dynamics} of Markov decision process is defined by the transition and reward probability distributions
\begin{eqnarray}
s_{t+1} \sim P_\textrm{tr}(\cdot |s_t, a_t) \\
r_{t+1} \sim P_\textrm{r}(\cdot | s_t, a_t).
\end{eqnarray}
From this definition, the probability of possible next state $s_{t+1}$ and reward $r_{t+1}$ depend solely on the immediately preceding state and action ($s_t, a_t$), but not on earlier states or actions ($s_0, a_0, \ldots, s_{t-1}, a_{t-1}$). This restriction on the state is called \emph{Markov property}.
Equivalently, under Markov property, given an action $a_t$ it means that a single state $s_t$ contains all the information that is necessary to determine (distribution over) the next state $s_{t+1}$ and $r_{t+1}$. This has an important consequence that an agent can determine an optimal action only looking at the current state, without worrying about any other information in the past.

Reinforcement learning (RL)~\citep{Sutton1998} is a suite of algorithms that trains a \emph{policy} that instantiates the agent in an MDP. %
A policy is a mapping from states to probabilities distribution over actions, functioning as a \emph{strategy} on how the agent behaves in the environment
\[
\pi(a_t | s_t) \coloneqq P_\pi(a_t | s_t).
\]
The goal of an RL algorithm is to obtain a policy $\pi^{*}$ that maximizes total rewards over a trajectory.
By running a policy $\pi$ in an environment for $T$ steps, we obtain a trajectory $\tau$
The initial state $s_0$ is sampled from an initial state distribution $\rho_0$
\[
s_0 \sim \rho_0(\cdot),
\]
and the next state as well as reward are sampled according to the dynamics of the MDP.
from the corresponding distributions

With slight abuse of notation, a probability to obtain a trajectory $\tau$ is
\[
P_\pi(\tau) = \rho_0(s_0) \prod_{t=0}^{T-1} P_\textrm{r}(r_{t}|s_t, a_t) P_\textrm{tr}(s_{t+1}|s_t, a_t) \pi(a_t | s_t).
\]
Using this, we define an \emph{expected return} of the policy $J(\pi)$
\[
J(\pi) = \int_\tau P_\pi(\tau)R(\tau) = \mathbb{E}_{\tau \sim \pi}\big[ R(\tau) \big],
\]
where $R$ simply computes a discounted sum over rewards
\[
R(\tau) = \sum_{t=0}^T \gamma^t r_t~~~~(\gamma \in (0, 1]).
\]

The overall objective of RL to obtain an optimal policy $\pi^*$ is then expressed by
\[
\pi^{*} = \arg \max_{\pi} J(\pi).
\]

\section{Generative Adversarial Networks}
Generative adversarial networks (GANs) \citep{gans} are a family of generative models that have two distinct models: generator and discriminator.
A generator $G$ produces a synthetic sample $\hat{\bm{x}}$ from a noise vector $\bm{z}$. On the other hand, discriminator $D$ takes a sample that is either 1) synthesized by the generator; or 2) drawn from the dataset, and classifies where it originates from. In practice, $D$ outputs a single scalar that represents the probability that the input comes from the dataset.
These two models are trained jointly, creating an adversarial setting: the generator aims to produce an authentic sample that can trick the discriminator, and the discriminator tries to detect the \emph{fake} samples.
Formally, the objective of this adversarial training is written as a minimax game:
\begin{equation}
\min_{\phi} \max_{\theta} \mathbb{E}_{\bm{x} \sim \mathcal{P}, \bm{z} \sim \mathcal{N}}[\log \big(1 - D_\theta(G_\phi(\bm{z})) \big) + \log D_\theta(\bm{x})],
\end{equation}
where the subscripts $\theta$ and $\phi$ denote their model parameters respectively.
This objective has a unique solution that the generator recovers the data distribution and the discriminator outputs a probability of $1/2$ for every sample.
Algorithm \ref{alg:gans} details how this optimization proceeds step by step.

\begin{algorithm}[t]
\caption{Minibatch training of generative adversarial nets. $k$ is a hyperparameter for the number of steps to update the discriminator. OptimStep refers to a single step of any optimizer based on a stochastic gradient.}
\label{alg:gans}
\algrenewcommand\alglinenumber[1]{}
\begin{algorithmic}[1]
\For{number of training iterations}
    \For{$k$ steps}
        \State $\{z^{(1)},\ldots,z^{(m)}\} \sim p_g(z)$ \Comment{Sample noise}
        \State $\{x^{(1)},\ldots,x^{(m)}\} \sim p_\textrm{data}(x)$ \Comment{Sample examples}

        \[
        \theta \leftarrow \text{OptimStep}\bigg(- \nabla_{\theta} \frac{1}{m} \sum_{i=1}^m \left[\log D_\theta\left(x^{(i)}\right) + \log\left(1 - D_\theta\left(G_\phi\left(z^{(i)}\right)\right)\right)\right]\bigg).
        \]
        \State \Comment{Update the discriminator parameters}
    \EndFor
    \State $\{z^{(1)},\ldots,z^{(m)}\} \sim p_g(z)$ \Comment{Sample noise}
    \[
    \phi \leftarrow \text{OptimStep}\bigg(\nabla_{\phi} \frac{1}{m} \sum_{i=1}^m \log\left(1 - D_\theta\left(G_\phi\left(z^{(i)}\right)\right)\right)\bigg).
    \]
    \State \Comment{Update the generator parameters}
\EndFor
\end{algorithmic}
\end{algorithm}

\section{Diffusion Models}
Diffusion models \citep{thermo} are another family of generative models that produces a sample by iteratively denoising a noisy input.
At high level, diffusion models define two stochastic processes: forward and reverse processes. The forward process involves adding noise on a data sample at each time step, transforming the input data distribution into a tractable noise distribution in the end. The reverse process simulates the opposite direction where the model gradually removes noise from an input at each step that leads to a synthetic sample in the end.
There have been various formulation, interpretation and implementation of this high level idea of diffusion models, and the theory and math behind them have been fairly convoluted.
\citet{song2021scorebased} focused on two major classes of approaches that instantiate this idea, and provided a framework that unifies the two different formulations. \emph{Score matching with langevin dynamics} (SMLD) \citep{song-and-ermon-19} trains a model to estimate a \emph{score} (i.e., $\nabla \log p_\textrm{data}(\bx; \sigma)$) for each noise scale $\sigma$ and generate a sample via Langevin dynamics,
and \emph{denoising diffusion probablistic models} (DDPM) \citep{ddpm} trains a \emph{denoiser} that predicts a noise in the input given the noise scale, and uses an analytical form of the reverse process to generate a sample from predicted noises.
\citet{song2021scorebased} notes that the training objective of DDPM can be also seen as predicting the score, and uses stochastic differential equation (SDE) %
to provide a general framework that encompasses these two classes of approaches. On top of this, \citet{Karras2022edm} further reformulates the unified framework in favor of practicality and describes how different approaches can be described under the new simplified formulation. %

We consistently use DDPM formulation for diffusion models throughout the thesis. %
In this section, we provide an overview of DDPM first, and the rest of it covers a high-level summary of the unified framework by \citet{song2021scorebased} and \citet{Karras2022edm}.%

\subsection{Denoising Diffusion Probabilistic Models (DDPM) \citep{ddpm}}

Given a noise schedule $\alpha_{1:T}$ and $\bar{\alpha}_t \coloneqq \prod_{s=1}^t \alpha_s$, the distribution over noisy input at timestep $t \in [0, T]$ for a data sample $\bm{x}_0$ is expressed as
\[
q(\bm{x}_t | \bm{x}_0) = \mathcal{N}(\bm{x}_t; \sqrt{\bar{\alpha}_t} \bm{x}_0, (1 - \bar{\alpha}_t) \mathbf{I}).
\]
This gives the result of forward diffusion process from $\bx_0$ at arbitrary time step $t$. Equivalently, given a noise $\bm{\epsilon} \sim \mathcal{N}(\bm{0}, \mathbf{I})$, a noisy sample $\bx_t$ is obtained with%
\begin{equation}
\label{eq:ddpm-forward-closed-form}
\bx_t = \sqrt{\bar{\alpha}_t} \bx_0 + \sqrt{1 - \bar{\alpha}_t} \beps.
\end{equation}
With this noisy sample $\bxt$ and the corresponding timestep $t$, we train a \emph{denoiser} $\bm{\epsilon}_\theta$ that predicts the noise $\bm{\epsilon}$ that was added to the original data sample $\bm{x}_0$. This training procedure is summarized in Algorithm \ref{alg:ddpm-training}.
\begin{algorithm}[t]
\caption{Training process in denoising diffusion probabilistic models.}
\label{alg:ddpm-training}
\begin{algorithmic}[1]
\Repeat
    \State $\bm{x}_0 \sim p_\textrm{data}$
    \State $t \sim \text{Uniform}(\{1,\ldots,T\})$
    \State $\bm{\epsilon} \sim \mathcal{N}(\mathbf{0},\mathbf{I})$
    \State $\tilde{\bm{x}}_t \leftarrow \sqrt{\bar{\alpha}_t} \bm{x}_0 + \sqrt{1 - \bar{\alpha}_t}\bm{\epsilon}$ \Comment{Add a noise to the data sample}
    \State Take gradient descent step on
    \State $\nabla_\theta \|\boldsymbol{\epsilon} - \bm{\epsilon}_\theta(\tilde{\bm{x}}_t, t)\|^2$
\Until{converged}
\end{algorithmic}
\end{algorithm}

The reverse process $q(\bx_{t-1}|\bx_t)$ is not tractable, however, it turns out that including $\bx_0$ in the condition enables us to write it down in a closed form:
\begin{align}
q(\bxtm | \bxt, \bxn) &= \mathcal{N}(\bxtm; \textcolor{blue}{\tilde{\mu}_t(\bxt, \bxn)}, \textcolor{magenta}{\tilde{\beta}} \mathbf{I}) \label{eq:ddpm-reverse-mu}
\\
\text{where}~~ \textcolor{blue}{\tilde{\mu}_t (\bxt, \bxn)} &\coloneqq \frac{\sqrt{\bar{\alpha}_{t-1}}\beta_t}{1 - \bar{\alpha}_t} \bxn + \frac{\sqrt{\alpha_t}(1 - \bar{\alpha}_{t-1})}{1 - \bar{\alpha}_t} \bxt \\
\text{and}~~ \textcolor{magenta}{\tilde{\beta}} &\coloneqq \frac{1 - \bar{\alpha}_{t-1}}{1 - \bar{\alpha}_t} \beta_t.
\end{align}
We aim to learn parameters $\theta$ that approximates the distribution
\[
p_\theta(\bxtm | \bxt) = \mathcal{N}\left(\bxtm; \textcolor{blue}{\mu_\theta (\bxt, t)}, \textcolor{magenta}{\Sigma_\theta (\bxt, t)} \right).
\]
\citet{ddpm} chooses to simply train $\textcolor{blue}{\mu_\theta(\bxt, t)}$ to match $\textcolor{blue}{\tilde{\mu}_t (\bxt, \bxn)}$. Rewriting Eq. \ref{eq:ddpm-forward-closed-form} for $\bxn$ gives
\begin{equation}
\bxn = \frac{1}{\sqrt{\bar{\alpha_t}}} (\bxt - \sqrt{1 - \bar{\alpha}_t} \bm{\epsilon})
\end{equation}
and replacing $\bx_0$ in Eq. \ref{eq:ddpm-reverse-mu} with this, we get
\[
\tilde{\mu}_t = \frac{1}{\sqrt{\alpha_t}} \left( x_t - \frac{1-\alpha_t}{\sqrt{1 - \bar{\alpha}_t}} \beps \right),
\]
where in practice $\beps$ is predicted with a neural network $\beps_\theta (\bxt, t)$:
\begin{equation}
\mu_\theta (\bxt, t) = \frac{1}{\sqrt{\alpha_t}} \left(\bxt - \frac{1-\alpha_t}{\sqrt{1 - \bar{\alpha}_t}} \beps_\theta(\bxt, t) \right).
\end{equation}
The sampling algorithm is designed by iteratively computing $\mu_\theta(\bxt, t)$ and sampling from the corresponding distribution, as shown in Algorithm \ref{alg:ddpm-sampling}.

\begin{algorithm}[h]
\caption{Sampling process in denoising diffusion probabilistic models}
\begin{algorithmic}[1]
\State $\bxT \sim \mathcal{N}(\mathbf{0},\mathbf{I})$
\For{$t = T,\ldots,1$}
    \State $\bz \sim \mathcal{N}(\mathbf{0},\mathbf{I})$ if $t > 1$, else $\bz = \mathbf{0}$
    \State $\bxtm = \frac{1}{\sqrt{\alpha_t}}\left(\bxt - \frac{1-\alpha_t}{\sqrt{1-\bar{\alpha}_t}}\beps_\theta(\bxt, t)\right) + \sigma_t\bz$
\EndFor
\State \Return $\bxn$
\end{algorithmic}
\label{alg:ddpm-sampling}
\end{algorithm}

\section{A unified framework by \citet{song2021scorebased}}
\citet{song2021scorebased} considers a diffusion process $\{\bx(t)\}_0^T$ indexed by a continuous time variable $t \in [0, T]$, where $\bx(0)$ follows the data distribution and $\bx(T)$ follows a tractable noise distribution (e.g., Gaussian distribution).
They consider modeling the process as a solution to an Itô SDE:
\begin{equation}
\label{eq:ito-sde}
d\bx = \bm{f}(\bx, t)dt + g(t) d\bm{w},
\end{equation}
where $\bm{w}$ is a standard Wiener process, which can be considered as a continuous variant of random walk, $\bm{f}(\cdot, t): \mathbb{R}^d \rightarrow \mathbb{R}^d$ is called \emph{drift} coefficient of $\bx(t)$, and the scalar function $g(t)$ is referred to as \emph{diffusion} coefficient.
Since a simplified drift coefficient $\bm{f}(\bx, t) = f(t)\bx$ is enough in most of the diffusion models, we use:
\begin{equation}
d\bx = f(t) \bx dt + g(t) d \bm{w},
\end{equation}
where $f(t)$ is a scalar function.
This SDE corresponds to the forward diffusion process. 
Under this formulation, the forward process of DDPM can be recovered by setting $f(t) = -\frac{1}{2}\beta(t), g(t) = \sqrt{\beta(t)}$, and that of SMLD can be recovered with $f(t) = 0, g(t) = \sqrt{2\dot{\sigma}(t) \sigma(t)}$, aside from necessary discretization, where both $\beta(t)$ and $\sigma(t)$ define (continuous) noise schedules.

To generate a sample, we need a reverse diffusion process. \citet{ANDERSON1982313} shows that the reverse of a diffusion process is also a diffusion process, given by the reverse-time SDE:
\begin{equation}
    \label{eq:sde-reversed}
    d\bx = [f(t)\bx - g(t)^2 \nabla_{\bx} \log p_t (\bx)]dt + g(t) d\bar{\bm{w}},
\end{equation}
where $\bar{\bm{w}}$ is a standard Wiener process running backward in time, $p_t(\bx)$ is a probability density of $\bx(t)$, and $dt$ is an infinitesimal negative timestep. Following this reverse-time SDE allows us to generate a sample, as long as we have estimate of the score $\nabla_{\bx} \log p_t(\bx)$.

The estimation of the score can be achieved by training a time-conditioned model $\bm{s}_\theta\left(\bx(t), t \right)$ with a score matching objective \citep{JMLR:v6:hyvarinen05a}.
Let us use $p_{0t}(\bx(t)| \bx(0))$ to denote the transition kernel from time $t=0$ to $t=t$, the training objective for score matching is
\begin{equation}
    \label{eq:sde-training-objective}
    \mathbb{E}_t \bigg[ \lambda(t) \mathbb{E}_{\bx(0)} \mathbb{E}_{\bx(t) | \bx(0))} \big[\| \bm{s}_\theta(\bx(t), t) - \nabla_{\bx(t)} \log p_{0t} (\bx(t) | \bx(0))\|^2_2 \big] \bigg].
\end{equation}
It is common to set $\lambda(t)$ to scale inversely proportional to the scale of the score, in order to avoid incurring unfairly large loss for a large score:
\[
\lambda(t) \propto 1 / \mathbb{E}_{\bx(0)}\mathbb{E}_{\bx(t) | \bx(0)} \left[\| \nabla_{\bx(t)} \log p_{0t} (\bx(t) | \bx(0))\|^2_2 \right]
\]
We note that the log probability of transition kernel is typically tractable.

In addition to this unified framework, \citet{song2021scorebased} shows another significant observation: for all diffusion processes, there is a corresponding \emph{deterministic process} whose trajectories share the same marginal probability distributions $\{p_t(\bx)\}^T_{t=0}$ with the SDE. The deterministic process is an ordinary differential equation (ODE) and they call it \emph{probability flow ODE}:
\begin{equation}
    \label{eq:song-ode}
    d\bx = \big[\textcolor{blue}{f(t)} \bx - \textcolor{darkgreen}{\frac{1}{2}g(t)^2} \nabla_{\bx} \log p_t(\bx) \big]dt.
\end{equation}
Notably, under this process the only source of \emph{noise} that affects the generation process is the initial value $\bx(T)$.
Expressing the reverse process in this way brings a lot of benefits. \citet{song2021scorebased} demonstrates that it allows us to compute exact likelihood on any input data, encode a datapoint $\bx(0)$ into a latent space $\bx(T)$ in \emph{uniquely identifiable} way, meaning that under some conditions the encoding is uniquely determined by the data distribution, and allows us to balance the efficiency and quality of samples with the help of black-box ODE solvers.

\section{Further simplifications by \citet{Karras2022edm}}
When it comes to exploring different designs of diffusion models, equation \ref{eq:song-ode} is not quite convenient, since $f(t)$ and $g(t)$ do not immediately correspond to any key concept in diffusion models such as noise schedule or scaling of the data. Thus \citet{Karras2022edm} reparameterizes the equation preferring the practicality of the formulation. In addition to noise scale $\sigma(t)$, they introduce $s(t)$ as an additional scale schedule that scales the original variable $\hat{\bx}$ with $\bx = s(t) \hat{\bx}$, and they always assume Gaussian noise rather than the general formulation in \citet{song2021scorebased}. Thus, the transition kernel is expressed as:
\begin{equation}
    p_{0t}\left(\bx(t) | \bx(0)\right) = \mathcal{N}\left(\bx(t);\,  s(t)\bx(0),\, s(t)^2 \sigma(t)^2 \mathbf{I}\right).
\end{equation}
With these, they transform the \emph{probability flow} ODE (equation \ref{eq:song-ode}) into:
\begin{align}
    d\bx = \left[ {\color{blue}\frac{\dot{s}(t)}{s(t)}} \bx - {\color{darkgreen} s(t)^2 \dot{\sigma}(t) \sigma(t)} \nabla_{\bx} \log p \left( \frac{\bx}{s(t)}; \sigma(t) \right) \right]dt, \\
    \text{where}~~p(\bm{y}; \sigma) = \int_{\mathbb{R}^d} p_\textrm{data}(\bx_0)~\mathcal{N}(\bm{y}; \bx_0, \sigma(t)^2 \mathbf{I})~d\bx_0. \label{eq:karras-marginal}
\end{align}
We note that equation \ref{eq:karras-marginal} is nothing but the convolution of the data distribution and Gaussian. For notational brevity, it is also expressed as
$p(\bm{y}; \sigma) = p_\textrm{data} \ast \mathcal{N}(\bm{0}, \sigma(t)^2 \mathbf{I})$.

They also simplified the training objective for score estimation. A \emph{denoiser} $D(\cdot; \sigma): \mathbb{R}^d \rightarrow \mathbb{R}^d$ can be trained by minimizing the following objective:
\begin{equation}
    \mathbb{E}_{\bm{y} \sim p_\textrm{data}} \mathbb{E}_{\bm{n} \sim \mathcal{N}(\bm{0}, \sigma^2 \mathbf{I})} \| D(\bm{y} + \bm{n}; \sigma) - \bm{y}\|^2_2,
\end{equation}
and then the score can be estimated by:
\begin{equation}
    \nabla_{\bx} \log p(\bx; \sigma) = \left(D(\bx; \sigma) - \bx\right) / \sigma^2.
\end{equation}
This seemingly nontrivial connection between the denoiser and the score matching is derived for a finite dataset in Appendix B.3 in \citet{song2021scorebased}.

%% file: block-stacking-with-diffusion/z_main.tex
\chapter{Block Stacking with Diffusion Models}
\label{chap:block-stacking-with-diffusion}

Humans gain an intuitive understanding of physics by interacting with and observing the world. This intuition helps them handle physical objects in a safe and efficient manner without ever performing explicit calculations. %
Equipping with such intuitions would be a great help for embodied agents that aim to work alongside with humans in unstructured environments.

Intuitive physics is a vague term that encompasses multiple aspects.
As a testbed, one of the subjects that previous works adopted is a set of toy blocks.
\citet{Li2016ToFO} takes in the visuals of a stack of blocks and predicts its stability score. \citet{pmlr-v48-lerer16} attempts to learn forward dynamics by predicting where each block will end up in the pixel space, after the stack collapses.
These approaches %
address interesting aspects of intuitive physics, however, their relevance to practical applications may not be immediately clear.
Learning forward dynamics model may be useful as it can serve as a surrogate of a simulator, exposing different modalities of input and output.
For instance, \citet{DBLP:conf/iclr/JannerLFTFW19} trains a forward dynamics model that operates directly in the pixel space, leveraging object-centric representations. They then use the model to 
plan block placements and build a stack that matches a reference image.
Although this approach shows some success in their task,
it is inherently inefficient as it falls in the category of \emph{rejection-based} approaches, where a sampler keeps producing a diverse and large number of candidates until the one that satisfies specific conditions is found. In their approach, the sampler picks a random position to drop a block at, and each drop is evaluated by the forward dynamics module, rejecting those that do not increase the alignment with the reference image.

\begin{figure}[t]
    \centering
    \includegraphics[width=\linewidth]{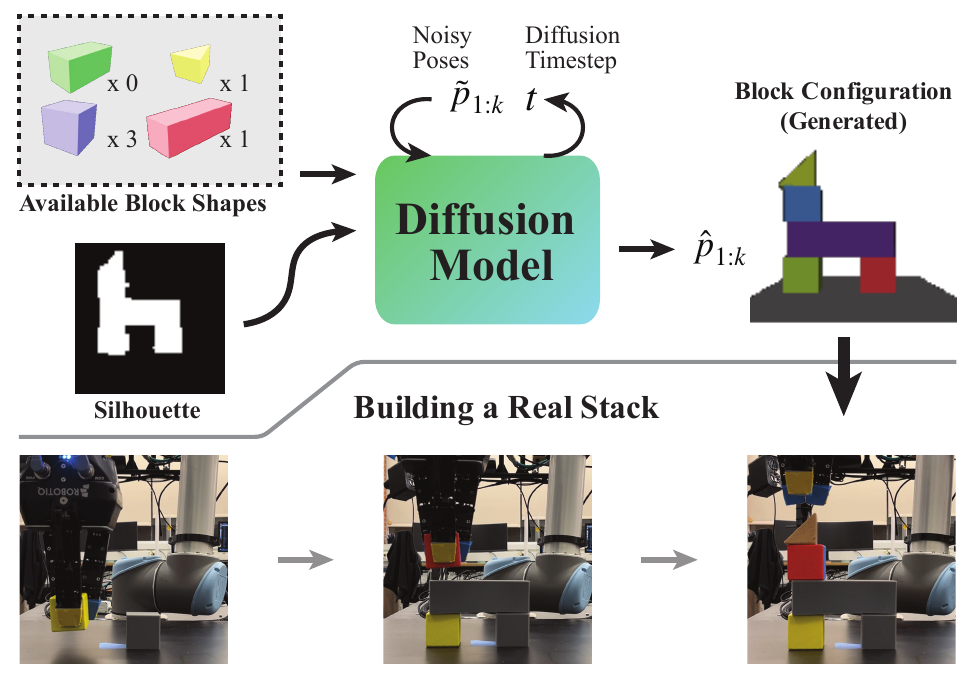}
    \caption{The overall pipeline of our approach. We design a diffusion model that takes a silhouette of a structure as well as available block shapes, and generate a set of block poses $\hat{\bp}_{1}, \ldots \hat{\bp}_k$ that makes up a stable stack matching the silhouette. We further demonstrate that we can apply our approach in the real block stacking task.}
    \label{fig:motivation}
\end{figure}

Instead of predicting stability scores or learning forward dynamics, in this work, we aim to learn joint distributions over the block poses that form a stable structure, and then \emph{generate} a stable block composition that meets certain user preference during inference.
There have been numerous approaches of generative modeling that aim to learn underlying data distribution from data samples.
Regarding practical tasks, generating realistic images has been one of the important topics in computer vision community.
The initial success was on generating images in a single domain, such as a face of a person~\citep{gans}.
Researchers also investigated approaches to \emph{condition} the image generation, with a class, other type of images and language descriptions.
It was diffusion models that enabled learning and generation of a largely diverse set of images, conditioned on various formats such as image categories, sketches, or text descriptions~\citep{NEURIPS2021_dhariwal, rombach2021highresolution, pmlr-v162-nichol22a, pmlr-v139-ramesh21a, Ho2022ClassifierFreeDG}.

We are specifically inspired by those that use a sketch as its condition for image generation.
A sketch serves as a great medium to specify the composition and semantics of a target sample to generate, while being simple and relatively easy to prepare.
In this work, we adopt a similar interface of specifying a target to generate, but instead of images we generate a composition (i.e., poses) of toy blocks that form a desired stack.
A stark difference from image generation works is that our method focuses on generating a \emph{physically stable} stack of blocks, rather than an arbitrary composition of blocks that matches the silhouette.
We adopt a simple binary silhouette rather than a sketch for conditioning, as it gives rise to more flexibility in generated designs. For example, a silhouette of a rectangle can be matched with either a single rectangle or a pair of cubes placed next with each other. Such arbitrariness in target specification makes the setup less trivial and more interesting.

In the following sections, we describe how we prepare the training data for a diverse set of stable stack of blocks~(Section \ref{subsec:method-data-generation}) our model architecture and training details~(Section \ref{subsec:method-model-arch-and-training}). In experiments, we first evaluate the generated stack of blocks in simulation~(Section \ref{subsec:exp-eval-in-sim}), followed by generation with a silhouettes extracted from stacks of real blocks and construction of the produced compositions with UR5 arm~(Section \ref{subsec:exp-in-real}).

\section{RELATED WORKS}%
\subsection{Intuitive Physics}
As humans efficiently interact with objects in the world without complicated calculations based on intuitive understanding of physics, embedding such intuition on a robotic system has a potential to improve its capability.
Multiple works \citep{Li2016ToFO, pmlr-v48-lerer16, DBLP:conf/iclr/JannerLFTFW19} consider a similar domain to ours that involve stability or dynamics around toy blocks.
One form of intuitive physics that humans have is a set of expectations about how the future unfolds given a setup, which includes concept of continuity and object permanence.
There have been focus on setting up benchmarks and measuring models' ability around such capability called violation-of-expectation (VoE) \citep{RePEc:nat:nathum:v:6:y:2022:i:9:d:10.1038_s41562-022-01394-8, NEURIPS2019_e88f243b, DBLP:journals/corr/abs-1803-07616, Piloto2018ProbingPK}.
For those that involve intuitive physics for robotics, 
\citet{DBLP:journals/corr/AgrawalNAML16} collected video sequence of poking objects with a robot arm and trained a forward and inverse dyanmics model from pixels, further showing that it can be deployed to generate a sequence of pokes to achieve a goal image.
\citet{unsupervised-learning-for-phys-interaction} is similar in spirit but has a rather scaled-up setup. They use $10$ robotic arms and collected long hours of videos interacting with various objects. With the dataset they trained an action-conditioned frame prediction model. They later use the model for model-predictive control to perform push task across various objects \citep{Finn2016DeepVF}.

\subsection{Diffusion Models for Pose Generation}
Given their impressive capability of learning multi-modal distributions, diffusion models \citep{ho2020ddpm} have been applied to learning distribution over SE(3) poses \citep{urain2022se3dif, simeonov2023rpdiff, Yoneda20236DoFSF}. 

\citet{urain2022se3dif} learns to generate a feasible gripper pose given a pointcloud of a mug cup. With the trained model they demonstrate that it allows joint optimization of trajectory and grasp pose.
\citet{structdiffusion2023} and \citet{Xu2024SetIU} consider language description such as ``set up a table" as an input to a model and generate placement poses of objects.

\section{METHOD}
\label{sec:method}

In this section, we describe our approach to generate the training dataset that contains a diverse set of stable stack of blocks in simulation, our model architecture that takes a variable number of block poses and silhouette, and the training procedure.

\subsection{Generating Data}
\label{subsec:method-data-generation}
\begin{figure}[t]
    \centering
    \includegraphics[width=0.9\linewidth]{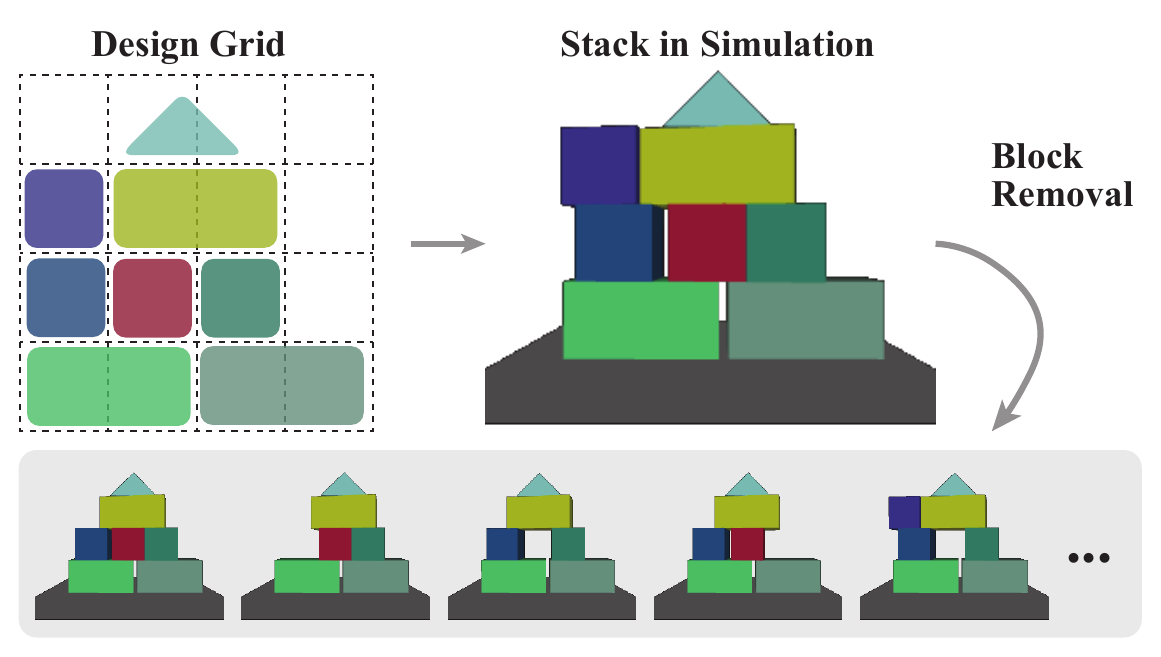}
    \caption{Our strategy to generate diverse set of stable stacks. After filling the design grid with shapes, we add a small horizontal displacement (sampled at random) at each layer and spawn them in simulation, verify its stability, and from there we try removing each block to create its variations.}
    \label{fig:removal}
\end{figure}

To train a model that can generate a diverse set of stable block poses, the quality and diversity of the dataset is crucial.
We need to have an algorithm that synthetically samples various stable block configurations to produce such a dataset at scale.
If we place excessive emphasis on diversity of the block stacks, a naive and general approach could be to spawn and drop a randomly selected shape at a random pose in simulation, wait until it settles and repeat this process to see if there's any meaningful stack constructed in the scene (by checking their height or collisions between blocks, for example).
In the case this process does not end up in a stack, we could reject it and start over again, repeating the process. This could potentially lead to a very general and extremely diverse dataset of block stacking, however, it would be excessively inefficient and thus impractical.

As an alternative approach, we employ a simple idea of designing a stack according to a pre-defined grid, followed by iterative \emph{block removal} while the stack stays stable. Although samples following a grid does not lead to a large number of unique compositions, the block removal process adds a lot of diversity.

Concretely, we consider a $4 \times 4$ grid that serves as a frame of a block stack design. For each layer starting at the bottom, we sample a shape one by one. Throughout this chapter we consider four different shapes: triangle, cube, rectangle and long rectangle, as shown in Figure~\ref{fig:motivation}.
A cube exactly fits in a single cell, rectangle and long rectangle takes two and three horizontal cells respectively. We stop and move onto the layer above once $3$ or $4$ cells are filled in the current layer. The configurations that goes over $4$ cells in a layer are rejected, and we do not consider placing rectangles or long rectangles vertically.
This procedure gives us various templates of block stacks. We convert each template to corresponding SE(3) poses, add a small noise on the horizontal position per layer and spawn them in simulation. After this, we run forward dynamics to verify that the stack does not collapse, render its silhouette, and add them to the dataset. We simply reject the design if the stack falls.

For each stack of blocks we obtain from the process described above, we perform \emph{block removal}, in which we remove blocks that do not collapse the structure. More concretely, from the initial stack of blocks, we try removing each block and test if the removal would make the structure fall by running forward dynamics and checking displacement of the existing blocks in the stack. If the stack stays still, we proceed to try removing another block. We run this process recursively, just like expanding a tree where the depth corresponds to the number of removed blocks, up to removing four blocks in total. We note that the block at the top is excluded from removal, and thus data samples always have the height of four cubes. In order to increase the diversity of stacks, we added all the intermediate nodes during this removal process to the dataset, rather than only considering the leaf nodes in the tree. This procedure is visualized in Figure~\ref{fig:removal}.

Following the process, we collected $191$k instances of stable block stacks, and split it into training and test dataset with a 9:1 ratio.

\subsection{Training and Sampling with DDPM}
The goal of our model is to learn a distribution over set of block poses, conditioned on a binary silhouette that shows the overall shape of the structure.
At test time we aim to use the trained model to sample various block poses that compose stable structure, given a silhouette.
We adopt denoised diffusion probablistic models (DDPM) \citep{ho2020ddpm} as the core framework of our model. 

Rather than describing the details of derivations or theory behind DDPM, in this section we only discuss our training objective as well as the sampling procedure during inference.
Following DDPM, we start with a forward diffusion process that adds noise to the state space. In our case, given a set of block poses $\bm{p}_1, \bm{p}_2, \ldots, \bm{p}_k \in \mathbb{R}^d$ that compose a stable stack and a diffusion timestep $t \in [1, T]$ that specifies a noise scale, we add noises following
\[
\tilde{\bm{p}}^t_i = \sqrt{\bar{\alpha}_t} \bm{p}_{i} + \sqrt{1 - \bar{\alpha}_t} \beps_i,~~ \beps_i \sim \mathcal{N}(\mathbf{0}, \mathbf{I}),
\]
where $\bar{\alpha}_t$ is a coefficient determined by a noise schedule.
The noisy poses $\tilde{\bm{p}}^t_i~(i = 1, 2, \ldots, k)$ are then fed to our \emph{denoising} network $D_\theta$ along with the diffusion timestep $t$, shapes $s_1, \ldots, s_k$ and the silhouette of the blocks $S$.
\begin{equation}
\label{eq:denoiser}
\hat{\beps}_{1:k} = D_\theta(\tilde{\bp}_{1:k}, t, s_{1:k}, S),    
\end{equation}
where the notation $X_{1:k}$ is equivalent to $\{X_1, \ldots, X_k\}$,
and $\beps_i$ is a predicted pose-noise for the $i$-th block. With the predicted noises, the training objective to minimize for a single sample is
\[
\frac{1}{k} \sum_{i=1}^{k} \| \beps_i - \hat{\beps}_i \|^2.
\]
Diffusion timestep $t$ is sampled uniformly random from $[1, T]$ at each training step.

Once the denoising network is trained, the sampling procedure starts with sampling a noisy pose from Gaussian distribution
\[
\tilde{\bp}^{T}_i \sim \mathcal{N}(\mathbf{0}, \mathbf{I}).
\]
From this initial noises, we iterate the following step from $t=T$ to $1$
\begin{align}
\tilde{\bp}^{t-1}_{i} = \frac{1}{\sqrt{\alpha}_t}\left( \tilde{\bp}^t_i - \frac{1-\alpha_t}{\sqrt{1 - \bar{\alpha}_t}}\hat{\beps}_i \right) + \sigma_t \bm{z},
\end{align}
where $\hat{\beps}_i$ is given by Eq. \ref{eq:denoiser} and $\bm{z} \sim \mathcal{N}(\mathbf{0}, \mathbf{I})$ if $t > 1$, otherwise $\bm{z} = \bm{0}$. A set of generated poses is obtained as $\tilde{\bp}^{0}_{1:k}$.

\subsection{Model Architecture}
\label{subsec:method-model-arch-and-training}
\begin{figure*}[t]
    \centering
    \includegraphics[width=\textwidth]{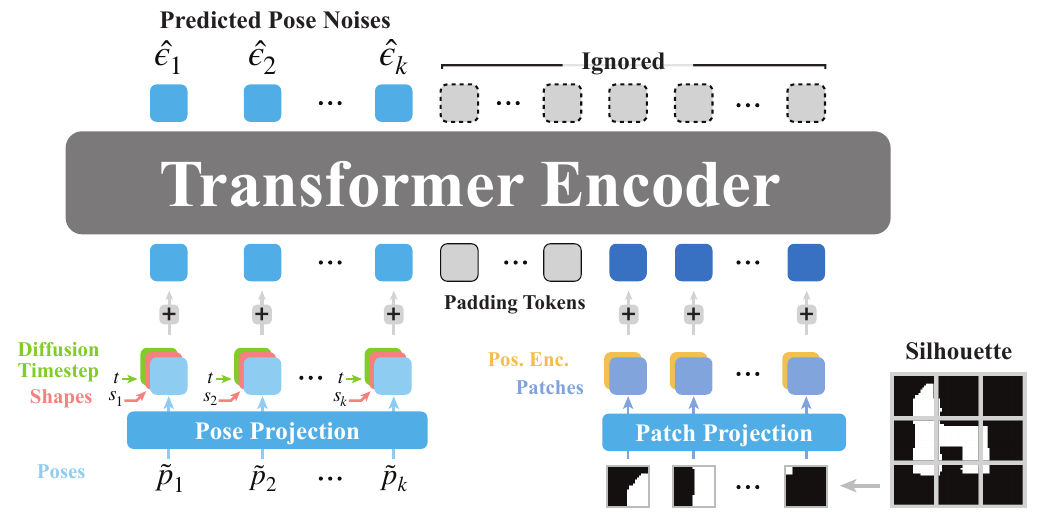}
    \caption{Our architecture for diffusion models.}
    \label{fig:arch}
\end{figure*}
Challenging requirements for the model come from the nature of the task that 1) the model must be able to work with \emph{variable number of} block poses 
and shapes of blocks; and that 2) the model must process the inputs from different modalities, including poses, shapes, diffusion timesteps and silhouettes that have spatial information.

We adopt Transformer \citep{NIPS2017_3f5ee243} as an architecture that can process inputs originating from different modalities in the form of input tokens. %
Given a scene that contains stable stack of $k~(\leq N)$ blocks, we extract a list of their poses $\bm{p}_1, \bm{p}_2, \ldots, \bm{p}_k \in \mathbb{R}^{6}$
and shape embeddings $\bm{s}_1, \bm{s}_2, \ldots, \bm{s}_k \in \mathbb{R}^{d}$. 
We used $6$ dimensional pose representation consisting of Cartesian coordinates for its translation and exponential coordinate for orientation.
A shape embedding is looked up from a codebook that stores a unique trainable embedding for each shape.
The poses are further projected onto $\mathbb{R}^d$ space with an MLP independently applied for each, and
the diffusion timestep $t \in [1, T]$ %
is also converted to an embedding $\bm{t} \in \mathbb{R}^{d}$.
With this, a pose, shape and diffusion timestep are all embedded in $\mathbb{R}^d$. We sum these up per object to get $k$ object tokens.
We make the number of input tokens to the Transformer encoder constant by padding the remaining $N-k$ object tokens with zero vector, resulting in $N$ object tokens independent of $k$.

Meanwhile, a silhouette of the block structure is given as a binary image of size  $64 \times 64$. Similarly to \citet{dosovitskiy2020vit}, we first split this into $16 \times 16$ patches and use a two-layer MLP to encode each independently to get silhouette tokens. Sinusoidal positional embeddings \citep{NIPS2017_3f5ee243} are added on the silhouette tokens to retain its spatial information.

The object and silhouette tokens are then added and fed to the Transformer encoder together.
At the last layer of the encoder, we project each contextualized block token back to the pose space ($\mathbb{R}^6$) and supervise it with the original noise added on the corresponding pose, following the framework of DDPM. Figure~\ref{fig:arch} summarizes this process and the architecture.

\section{EXPERIMENTS}
\label{sec:experiments}

\begin{figure}
    \centering
    \begin{subfigure}{\columnwidth}
        \centering
        \begin{subfigure}{0.23\columnwidth}
            \includegraphics[width=\textwidth]{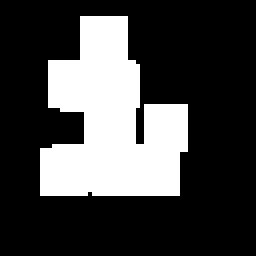}
        \end{subfigure}%
        \hspace{0.005\columnwidth}%
        \begin{subfigure}{0.23\columnwidth}
            \includegraphics[width=\textwidth]{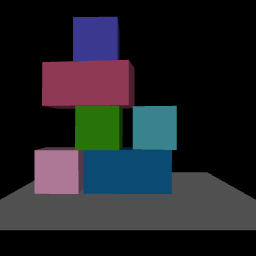}
        \end{subfigure}%
        \hspace{0.03\columnwidth}%
        \begin{subfigure}{0.23\columnwidth}
            \includegraphics[width=\textwidth]{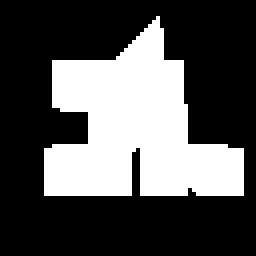}
        \end{subfigure}%
        \hspace{0.005\columnwidth}%
        \begin{subfigure}{0.23\columnwidth}
            \includegraphics[width=\textwidth]{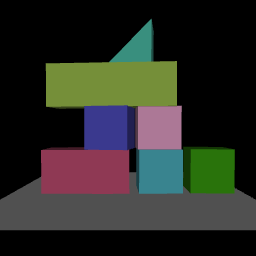}
        \end{subfigure}
    \end{subfigure}

    \vspace{0.5em}

    \begin{subfigure}{\columnwidth}
        \centering
        \begin{subfigure}{0.23\columnwidth}
            \includegraphics[width=\textwidth]{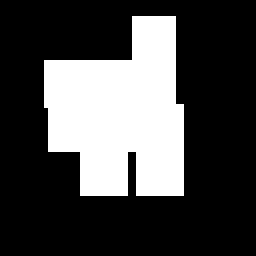}
        \end{subfigure}%
        \hspace{0.005\columnwidth}%
        \begin{subfigure}{0.23\columnwidth}
            \includegraphics[width=\textwidth]{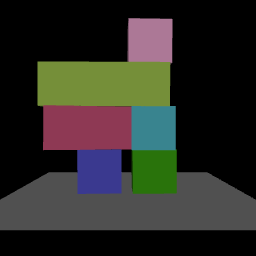}
        \end{subfigure}%
        \hspace{0.03\columnwidth}%
        \begin{subfigure}{0.23\columnwidth}
            \includegraphics[width=\textwidth]{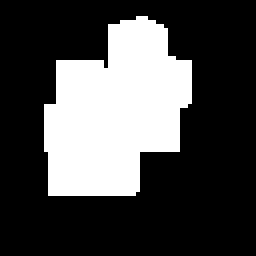}
        \end{subfigure}%
        \hspace{0.005\columnwidth}%
        \begin{subfigure}{0.23\columnwidth}
            \includegraphics[width=\textwidth]{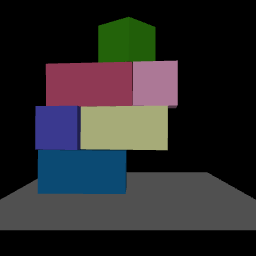}
        \end{subfigure}
    \end{subfigure}

    \caption{Silhouettes from the heldout dataset and rendering of block poses generated by our model.}
    \label{fig:silhouettes-vs-generated-stack}
\end{figure}

\begin{figure}
    \centering
    \begin{subfigure}{\columnwidth}
        \centering
        \begin{subfigure}{0.196\columnwidth}
            \includegraphics[width=\textwidth]{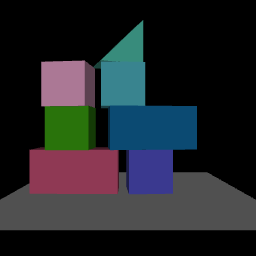}
        \end{subfigure}%
        \hfill%
        \begin{subfigure}{0.196\columnwidth}
            \includegraphics[width=\textwidth]{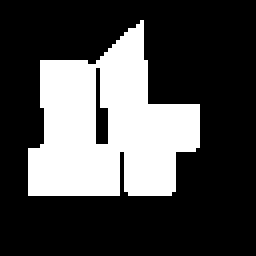}
        \end{subfigure}%
        \hspace{0.018\columnwidth}%
        \begin{subfigure}{0.196\columnwidth}
            \includegraphics[width=\textwidth]{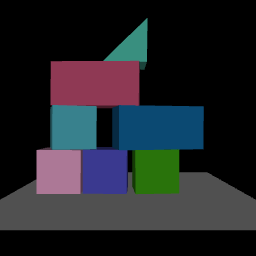}
        \end{subfigure}%
        \hfill%
        \begin{subfigure}{0.196\columnwidth}
            \includegraphics[width=\textwidth]{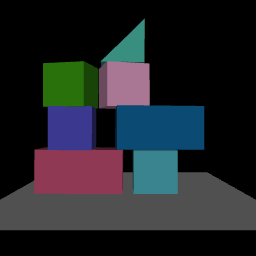}
        \end{subfigure}%
        \hfill%
        \begin{subfigure}{0.196\columnwidth}
            \includegraphics[width=\textwidth]{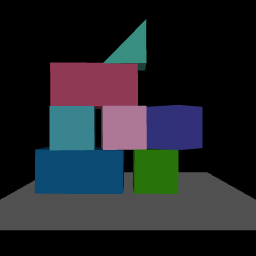}
        \end{subfigure}
    \end{subfigure}

    \vspace{0.5em}

    \begin{subfigure}{\columnwidth}
        \centering
        \begin{subfigure}{0.196\columnwidth}
            \includegraphics[width=\textwidth]{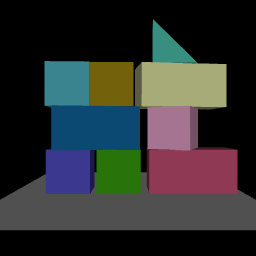}
        \end{subfigure}%
        \hfill%
        \begin{subfigure}{0.196\columnwidth}
            \includegraphics[width=\textwidth]{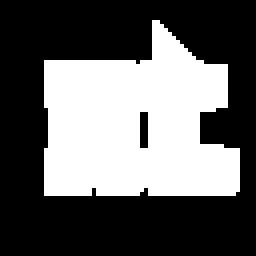}
        \end{subfigure}%
        \hspace{0.018\columnwidth}%
        \begin{subfigure}{0.196\columnwidth}
            \includegraphics[width=\textwidth]{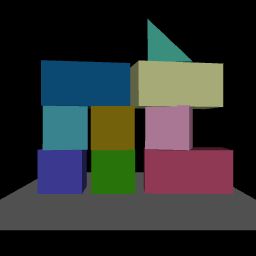}
        \end{subfigure}%
        \hfill%
        \begin{subfigure}{0.196\columnwidth}
            \includegraphics[width=\textwidth]{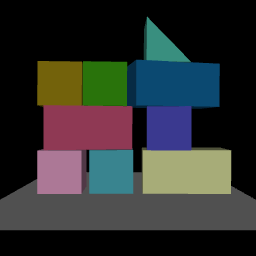}
        \end{subfigure}%
        \hfill%
        \begin{subfigure}{0.196\columnwidth}
            \includegraphics[width=\textwidth]{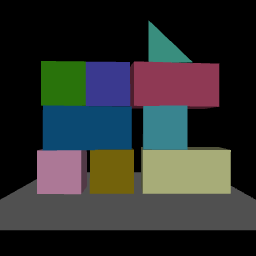}
        \end{subfigure}
    \end{subfigure}

    \caption{A reference stack (ground truth) with its silhouette (left), and a diverse set of structures generated from the silhouette by our model (right).}
    \label{fig:silhouettes-vs-generated-stack--diversity}
\end{figure}

We aim to evaluate our approach from the following three perspectives: 1) how often our approach can generate block poses that indeed form a stable stack; 2) how diverse the generated stacks are; and 3) how consistent a generated stack is with the silhouette provided as a condition.
After evaluating the model in simulation, we deploy it on the real world setting, where a user gives a reference silhouette by building a real stack of blocks, followed by a UR-5 arm building a stack that matches the silhouette.

\subsection{Evaluation in simulation}
\label{subsec:exp-eval-in-sim}
We evaluate our model using the heldout test dataset.
Figure~\ref{fig:silhouettes-vs-generated-stack} shows generated stacks paired with the corresponding silhouettes.
In Figure~\ref{fig:silhouettes-vs-generated-stack--diversity}, we present a diverse set of stacks the model produced for a single silhouette, showing its capability to model multimodal distributions.

We compute intersection over union (IoU) to quantify silhouette consistency, and runs forward dynamics simulation for stability testing. More concretely on the stability check, we first spawn blocks of the specified shapes at the generated poses in simulation, run forward dynamics, and then check if displacement of blocks remains small. We compute these metrics over $500$ generated samples, where we generate and use a single sample per test silhouette. %
With this, the reported IoU was $0.884 \pm 0.084$ and the ratio of stable stack was $0.862$~($431$ out of $500$).%

To provide a baseline against our approach, we implemented a heuristic pose sampler. Given a silhouette and a set of available shapes, the baseline algorithm searches for a potential placement pose of each block by maximizing the alignment score. The alignment score is computed from the intersection between the silhouette of the candidate placement and the reference silhouette, how well the placement is aligned with the edges of the reference silhouette, and collisions with previously placed blocks. To find a placement that has a large alignment score, we sample $64$ placement positions randomly and perform linear search from each point seeking for a (local) maximum. The height is sampled from a pre-defined set of values, corresponding one of four levels in the stack.
Notably, this baseline sampler does not consider stable placement and only aims to maximize the alignment with the given reference silhouette. Running this baseline against the test silhouette, we get IoU of $0.952 \pm 0.038$ and the ratio of stable stack $0.724~(362 / 500)$. As expected, the IoU on this baseline is higher since it was literally the objective metric of the algorithm, however, there are many scenarios that simply aligning blocks to match the silhouette does not result in a stable composition. This explains the lower number on the ratio of stable stack, and demonstrates the capability of our approach to implicitly learn stability.

\subsection{Block Stacking in the Real World}
\label{subsec:exp-in-real}

\begin{figure}[t]
    \centering
    \begin{subfigure}[b]{0.48\linewidth}
        \includegraphics[width=\linewidth, height=0.9\textheight, keepaspectratio]{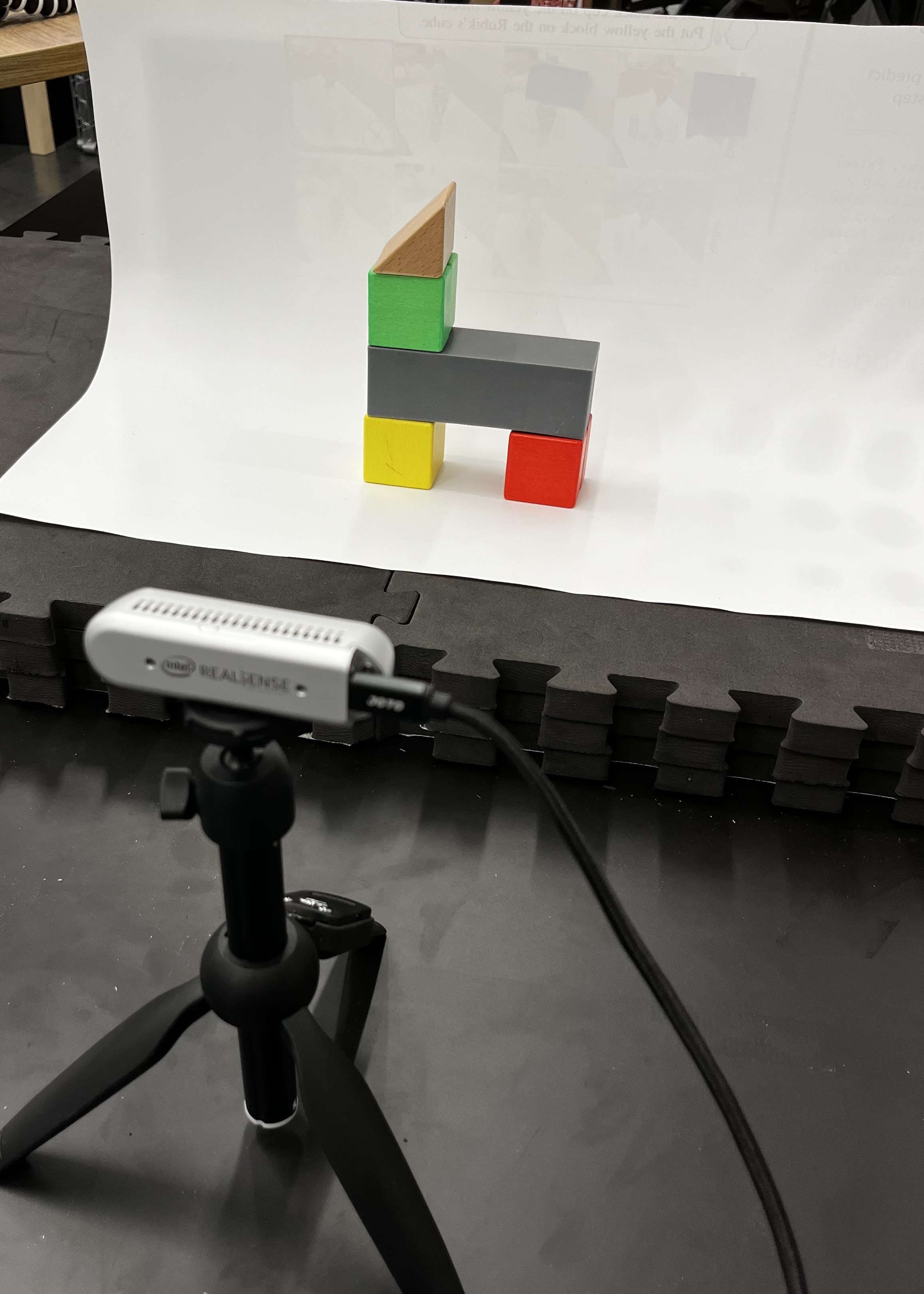}
    \end{subfigure}
    \begin{subfigure}[b]{0.48\linewidth}
        \begin{subfigure}[b]{\linewidth}
            \includegraphics[width=0.7\linewidth]{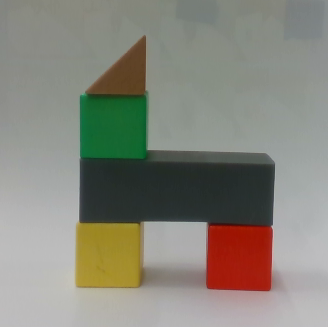}
        \end{subfigure}
        \vfill
        \begin{subfigure}[b]{\linewidth}
            \includegraphics[width=0.7\linewidth]{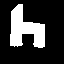}
        \end{subfigure}
    \end{subfigure}
    \caption{A simple rig to capture real block silhouette. Our setup with toy blocks and Realsense camera (left), captured RGB image (top right) and extracted silhouette (bottom right).}
    \label{fig:real-block-setup}
\end{figure}

\begin{figure}
    \centering
    \begin{subfigure}{\columnwidth}
        \centering
        \begin{subfigure}{0.236\columnwidth}
            \includegraphics[width=\textwidth]{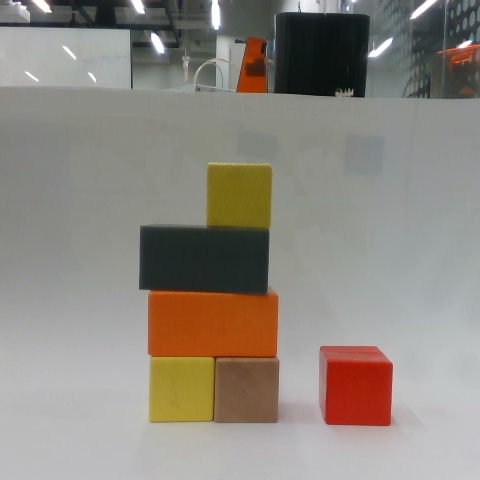}
        \end{subfigure}%
        \begin{subfigure}{0.236\columnwidth}
            \includegraphics[width=\textwidth]{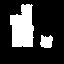}
        \end{subfigure}%
        \begin{subfigure}{0.236\columnwidth}
            \includegraphics[width=\textwidth]{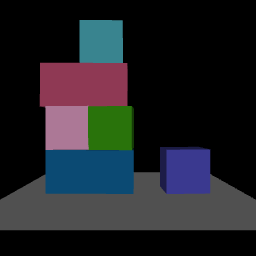}
        \end{subfigure}%
        \begin{subfigure}{0.236\columnwidth}
            \includegraphics[width=\textwidth]{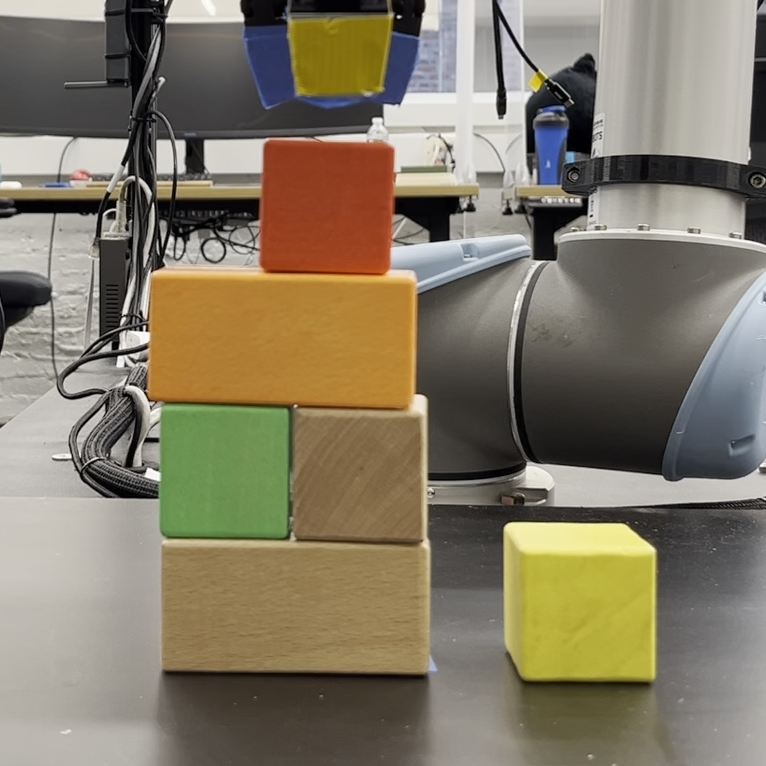}
        \end{subfigure}
    \end{subfigure}

    \vspace{0.5em}

    \begin{subfigure}{\columnwidth}
        \centering
        \begin{subfigure}{0.236\columnwidth}
            \includegraphics[width=\textwidth]{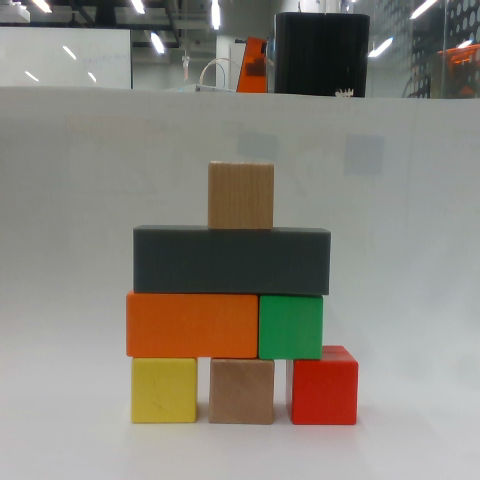}
        \end{subfigure}%
        \begin{subfigure}{0.236\columnwidth}
            \includegraphics[width=\textwidth]{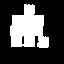}
        \end{subfigure}%
        \begin{subfigure}{0.236\columnwidth}
            \includegraphics[width=\textwidth]{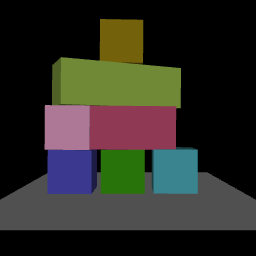}
        \end{subfigure}%
        \begin{subfigure}{0.236\columnwidth}
            \includegraphics[width=\textwidth]{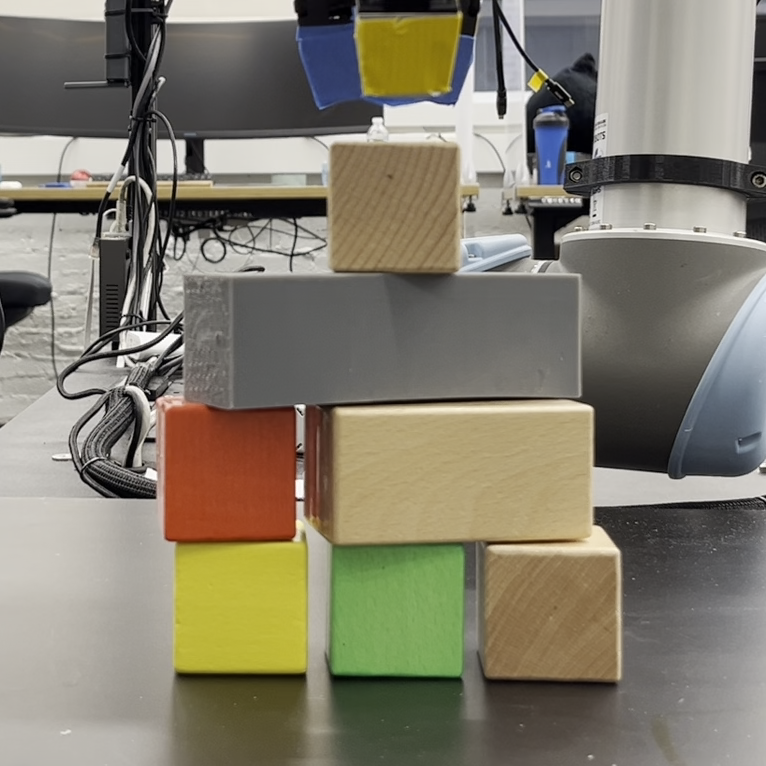}
        \end{subfigure}
    \end{subfigure}
    
    \vspace{0.5em}
    
    \begin{subfigure}{\columnwidth}
        \centering
        \begin{subfigure}{0.236\columnwidth}
            \includegraphics[width=\textwidth]{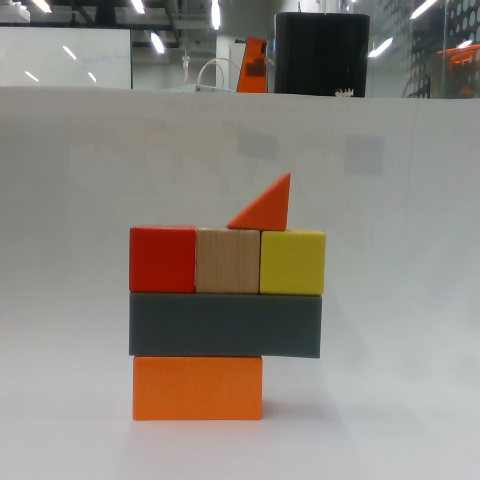}
        \end{subfigure}%
        \begin{subfigure}{0.236\columnwidth}
            \includegraphics[width=\textwidth]{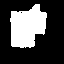}
        \end{subfigure}%
        \begin{subfigure}{0.236\columnwidth}
            \includegraphics[width=\textwidth]{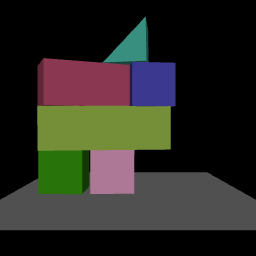}
        \end{subfigure}%
        \begin{subfigure}{0.236\columnwidth}
            \includegraphics[width=\textwidth]{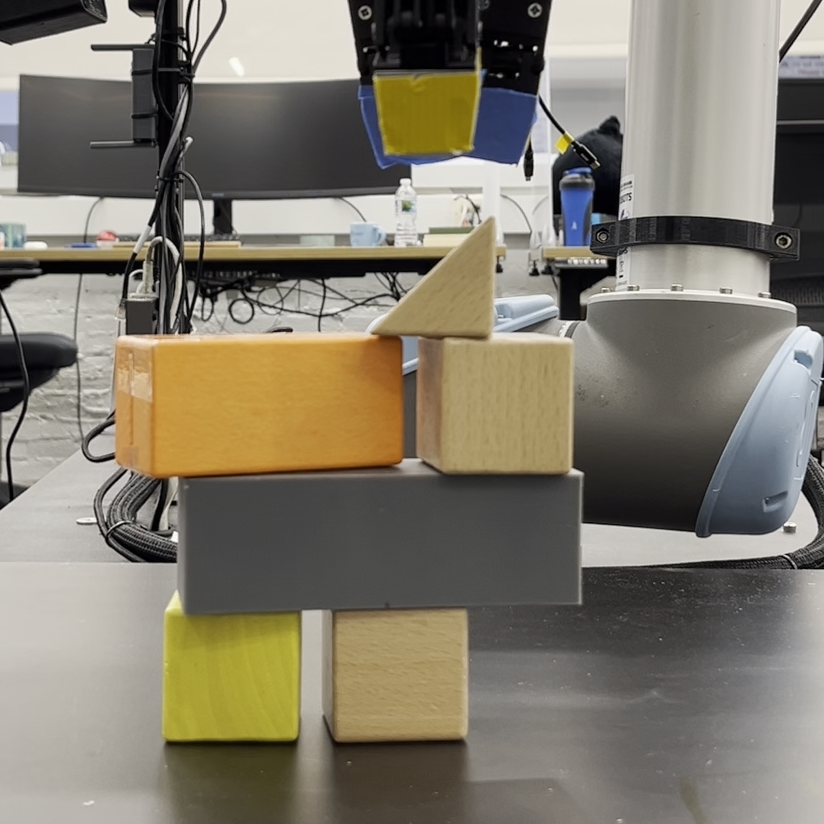}
        \end{subfigure}
    \end{subfigure}
    
    \vspace{0.5em}
    
    \begin{subfigure}{\columnwidth}
        \centering
        \begin{subfigure}{0.236\columnwidth}
            \includegraphics[width=\textwidth]{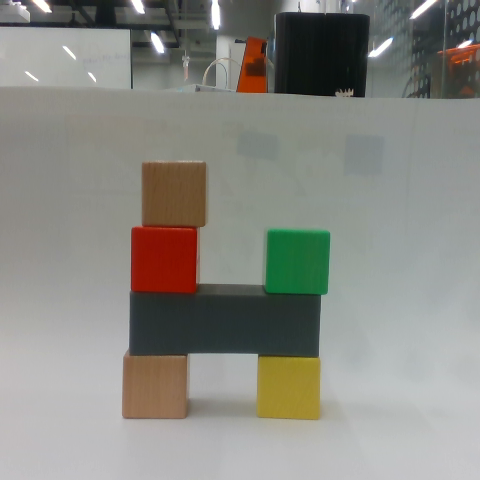}
        \end{subfigure}%
        \begin{subfigure}{0.236\columnwidth}
            \includegraphics[width=\textwidth]{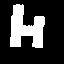}
        \end{subfigure}%
        \begin{subfigure}{0.236\columnwidth}
            \includegraphics[width=\textwidth]{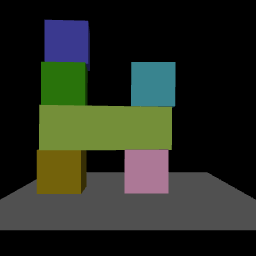}
        \end{subfigure}%
        \begin{subfigure}{0.236\columnwidth}
            \includegraphics[width=\textwidth]{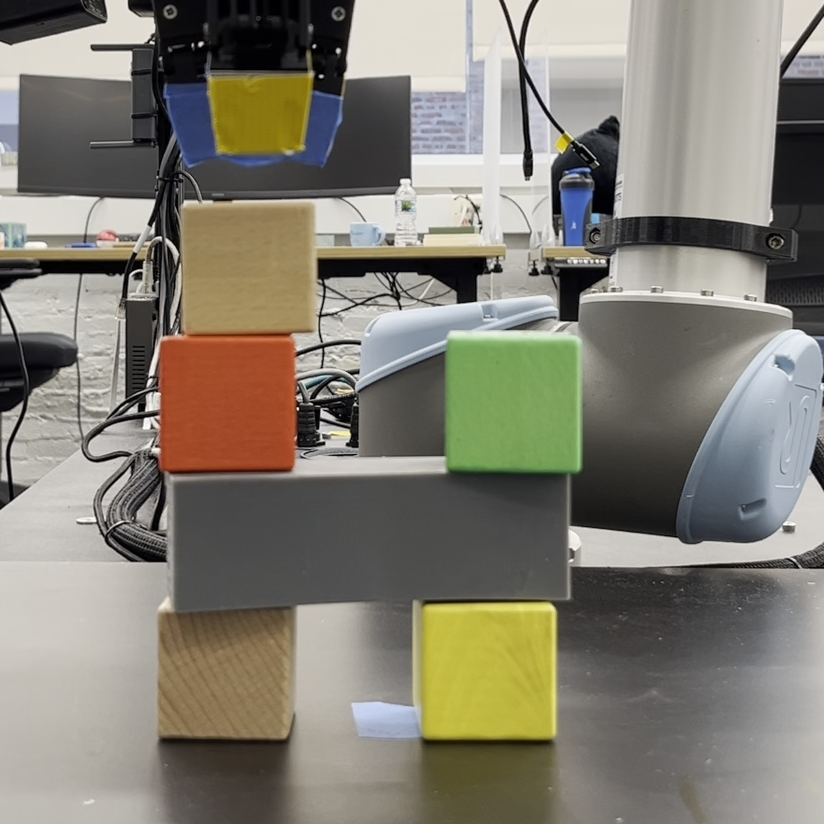}
        \end{subfigure}
    \end{subfigure}

    \caption{Generated samples from the extracted silhouette and the real stack built with UR-5 arm.}
    \label{fig:real-stack-vs-generated-and-reconstructed}
\end{figure}

In order to show that our method can also perform well in a real environment, we conduct an experiment with toy blocks and a UR-5 arm.
Our goal here is to build a pipeline as follows. First, a user provides a silhouette by presenting a reference stack of toy blocks. After extracting silhouette from the stack, we run our model to generate a stable composition of blocks that matches the silhouette. Finally we use a UR-5 arm to build the generated stack on the table with real blocks.

A silhouette is extracted from a stack with a simple rig we prepared.
It consists of an RGBD camera (Realsense 435D), toy blocks and the white background as shown in Figure~\ref{fig:real-block-setup}.
Given a stack of blocks that a user has built, our goal with this rig is to obtain a binary silhouette of size $64 \times 64$.
From the depth reading, we first filter out the background pixels based on distance,
apply a median filter to fill tiny ``holes'' on the block pixels, and remove the white pixels. We finally resize the extracted silhouette appropriately and obtain the silhouette shown at the bottom right in Figure~\ref{fig:real-block-setup}.
Other example pairs of block stack and extracted silhouette can be seen from the left two columns in Figure~\ref{fig:real-stack-vs-generated-and-reconstructed}.

With this setup, we build $10$ reference stacks of blocks and run the pipeline.
To ensure the diversity and validity of the stacks for the evaluation purpose, 
we sampled examples from the test dataset and manually built real block stacks mimicking their reference rendering in simulation.

With the extracted silhouettes paired with the shapes used in each stack, we use our pretrained model to generate candidate block poses.
For each set of generated block poses, we identify which layer each block belongs to based on a heuristic,
and execute pick-and-place onto the corresponding generated pose with a UR-5 arm.
We note that our model aims to match the given silhouette with the generated stack, but not to recreate the exact stack that a user has built, as you can see in the structural difference between the leftmost and the rightmost columns in Figure \ref{fig:real-stack-vs-generated-and-reconstructed}.

Figure~\ref{fig:real-stack-vs-generated-and-reconstructed} summarizes the original user's stack, extracted silhouette, a stack generated by the model (rendered in simulation) and the real stack built with UR-5.

Out of the ten cases we tried, this pipeline succeeded building all of the stacks stably, with mostly negligible discrepancy from the silhouettes of the original stack.
We clarify that this does not mean our system is perfect. As we saw in Section~\ref{subsec:exp-eval-in-sim}, there are cases where the model generates unstable block compositions.
The results of this real-robot experiment rather demonstrate that the model works robustly well enough against potentially out-of-distribution silhouettes, extracted from user's reference stack.%

\section{CONCLUSIONS}
In this work, we presented a new approach toward acquiring a specific aspect of intuitive physics, namely distributions over block poses that form stable structures. 
We developed a diffusion model that is trained from a set of block poses that compose stable stacks, paired with their silhouettes. During inference, the model generates diverse set of stable block poses that aligns with the reference silhouette given by a user. We showed that our approach outperforms a strong heuristics-based baseline by its ability to produce stable structures, as well as demonstrating its applicability in the real environment.

While it is still a long way to obtain various forms of intuitions and understanding  that humans possess for the physical world, we hope this work serves as a proof-of-concept for how to approach intuitive physics from a generative modeling perspective.

%% file: diffusha/z_main.tex
\chapter{Shared Autonomy with Diffusion Models}
\label{chap:diffusha}
\begin{figure}
    \centering
    \includegraphics[width=0.49\linewidth]{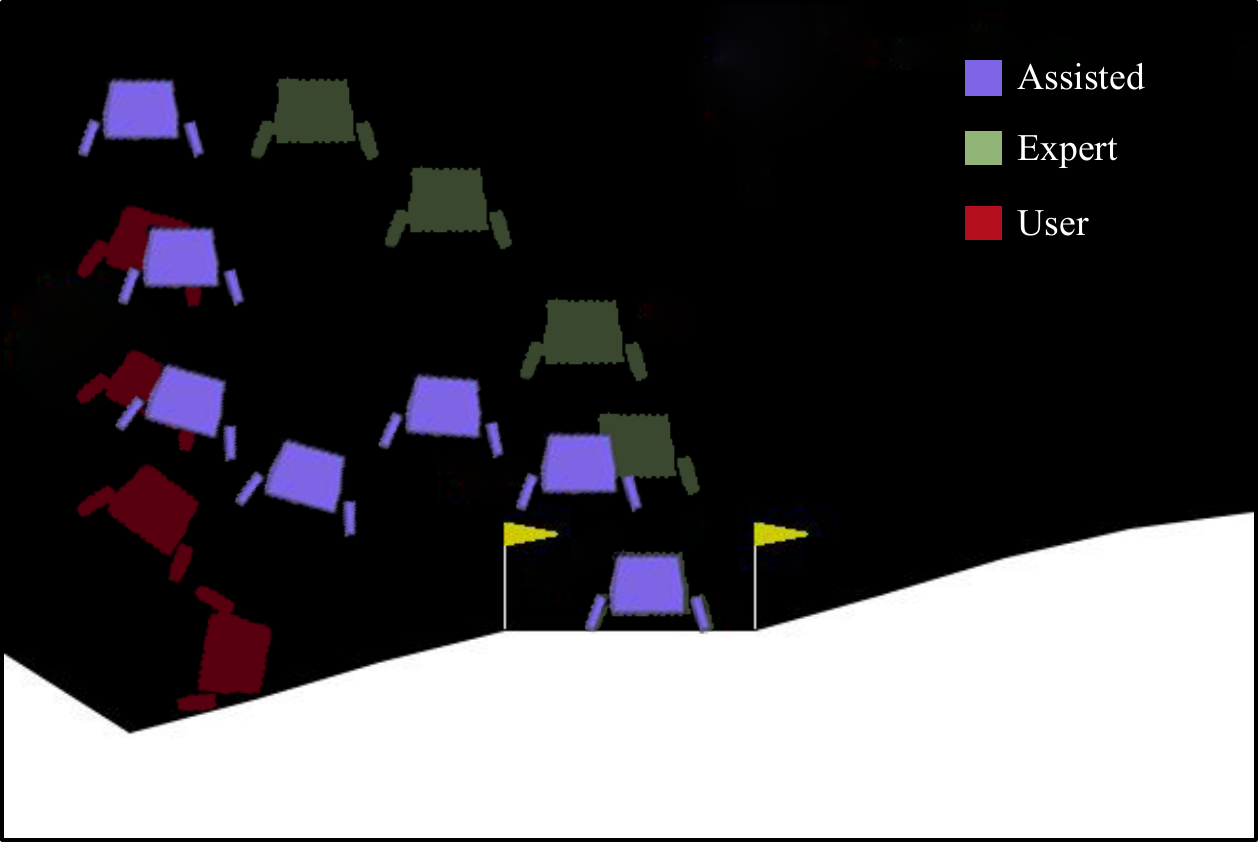}%
    \includegraphics[width=0.49\linewidth]{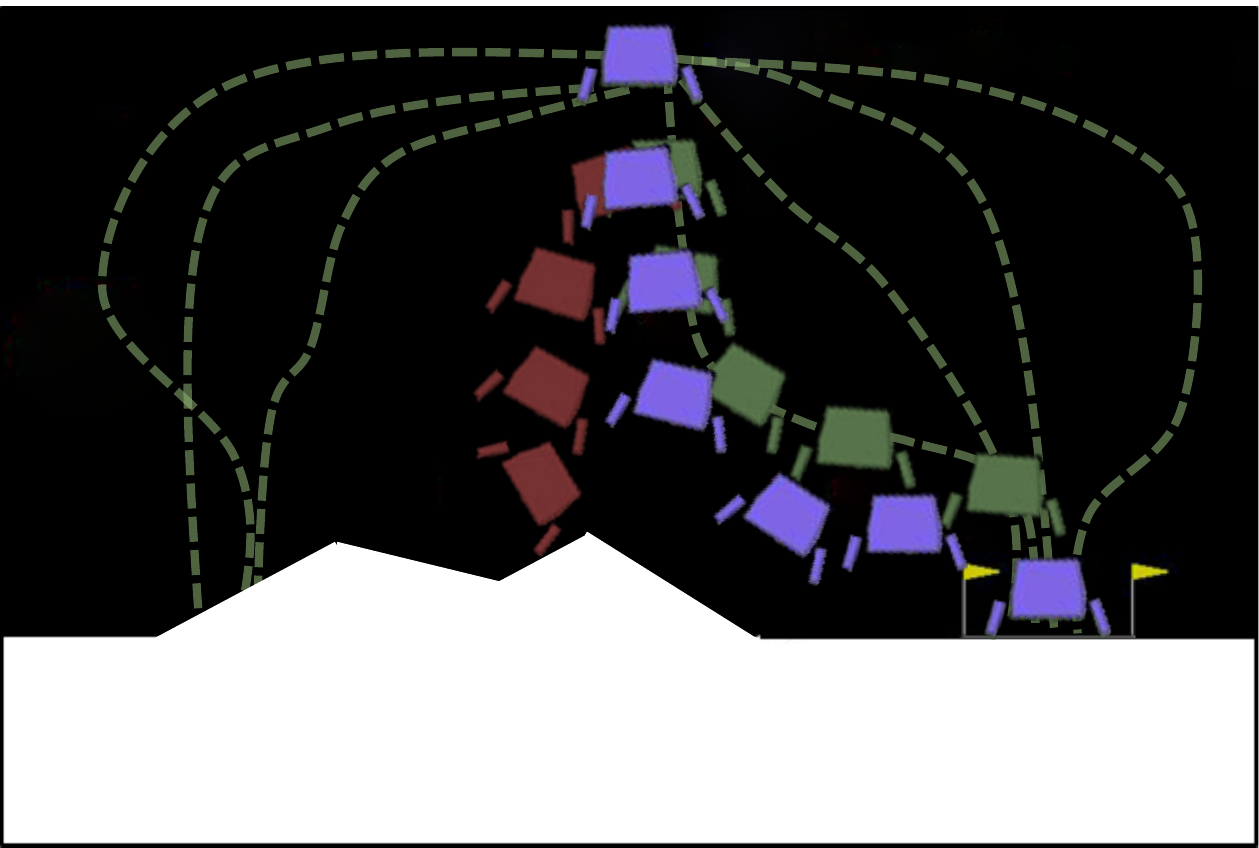}
    \caption{Our framework utilizes a diffusion model to adapt a user's action (red) to those from a demonstration distribution (green) in a manner (blue) that balances a user's desire to maintain control authority with the benefits (e.g., safety) of conforming to the desired (demonstration) distribution. Without knowledge of the user's specific goal (e.g., the landing location), the demonstration distribution reflects different goals that the demonstration trajectories previously reached.}
    \label{fig-diffusha:model-intuition}
\end{figure}

Contemporary robots primarily operate in one of two different ways---full teleoperation or full autonomy. Teleoperation is common in unstructured environments (e.g., underwater), where the proficiency with which robots are able to understand their surroundings is insufficient for fully autonomous robots to operate reliably. However, direct teleoperation requires users to  interpret the robot's environment observations while simultaneously controlling its low-level actions, a responsibility that is particularly challenging for highly dynamic tasks. This operational gap motivates a setting in which a human and an autonomous agent collaborate and share control of the robot.

Shared autonomy~\cite{abbink18} is a framework in which a human user (also referred to as the \textit{pilot}) performs a task with an assistance of an autonomous agent (also referred to as the \textit{copilot})~\citep{goertz63, rosenberg93, Aigner97, dragan12, dragan13, gombolay14, billings2021towards}. 
The role of the agent is to complement the control authority of the user, whether to improve the robot's  performance on the current task or to encourage/ensure safe behavior. An important consideration when providing assistance via shared autonomy is the degree to which the agent balances the user's preference for maintaining control authority (i.e., the \emph{fidelity} of the assisted behavior relative to the user's actions), and the potential benefits of endowing more control to the agent (i.e., the \emph{conformity} of the assisted behavior to that of an autonomous agent).

A core difficulty of shared autonomy lies in the fact that the user's goal (intent) is typically not known. Many approaches to shared autonomy assume that there is a fixed, discrete set of candidate goals and seek to infer the user's specific goal at test-time based on observations, including the user's control input~\cite{muelling17, javdani15, perez15, hauser13, dragan13}. Such assumptions may be reasonable in structured environments (e.g., in the context of a manipulation task when there is a small number of graspable objects sitting on a table). However, they can be limiting in unstructured environments that lack well-defined goals or that have a very large set of potential goals.

Bootstrapped by function approximation with neural-networks, recent approaches based on deep reinforcement learning (RL) algorithms seek to learn assistive policies without assumptions on the
knowledge of the goal space, or the assumption that the environment dynamics are known.
\textcite{sha-via-deeprl} propose a deep RL approach to shared autonomy for domains with discrete actions, nominally relying on reward feedback from the user as an alternative to assuming that the goal space is known. In an effort to balance the user's control authority with task performance, the assistant chooses the action most similar to that of the user while also satisfying a state-action value constraint.
\textcite{rsa} treat the \text{copilot} as providing a residual that is added to the user's actions to correct for unsafe behavior. They train their model to minimize the norm of the residual, subject to a goal-agnostic reward constraint that represents safe behavior.

These methods treat the pilot as a part of the environment, using an augmented state that includes the user's action. %
Framing the problem in this way has a clear and significant advantage---it enables the direct utilization of the modern suite of tools for deep RL. However, these methods have two notable limitations. First, they require human-in-the-loop interaction during training in order to generate user actions while learning the assistant's policy. Since the sample complexity of deep RL makes this interaction intractable, these methods replace the human with a \emph{surrogate policy}. If this surrogate is misspecified or invalid, this approach can lead to copilots that are incompatible with actual human pilots~\parencite{rsa}. Second, these methods require access to task-specific reward during training, which may be difficult to obtain in practice.

In light of these limitations, we propose a model-free approach to shared
autonomy that interpolates between the user's action and an action sampled from
a generative model that provides a distribution over desired behavior
(Fig.~\ref{fig-diffusha:model-intuition}). %
Our approach has the distinct advantage that it does not require knowledge of or access to the user's policy or any reward feedback during training, thus eliminating the need for reward engineering.
Instead, our training process, which involves learning the generative model, only requires access to
trajectories that are representative of desired behavior.

The generative model that underlies our approach is a diffusion model~\parencite{vincent11,thermo,song19,ddpm}, which has proven highly effective for complex generation tasks including image synthesis~\parencite{dhariwal2021diffusion, saharia2022photorealistic}. Diffusion models consist of two key processes: the \textit{forward process} and the \textit{reverse process}.
The forward process iteratively adds Gaussian noise to the input with an increasing noise scale, while the reverse process is trained to iteratively denoise a noisy input in order to arrive at the target distribution. 
As part of this denoising process, the model produces the gradient direction in which the likelihood of its input increases under the target distribution.
Once the model is trained, generating a sample from the (unknown) target distribution involves running the reverse process on a sample drawn from a zero-mean isotropic Gaussian distribution.

As we will see in the following sections, a direct use of diffusion models for shared autonomy ends up in generating an action that ignores user's intent (i.e., low \emph{fidelity} to user intent), even though the action would be consistent with the desired behaviors (i.e., high \emph{conformity} to the target behaviors). To address this, we propose a new algorithm that controls the effect of the forward and reverse process through a \emph{forward diffusion ratio} $\fwr$ %
that regulates the balance between the fidelity and the conformity of the generated actions. The forward diffusion ratio provides a formal bound on the extent to which the copilot's action deviates from that of the user.

We evaluate our shared autonomy algorithm using a series of continuous control
tasks. In each case, we demonstrate that our algorithm significantly improves
the performance of a variety of different pilots, and we analyze the effects of
a range of different forward diffusion ratios.

\section{Method}

\begin{figure}[!t]
    \centering
    \includegraphics[width=\linewidth]{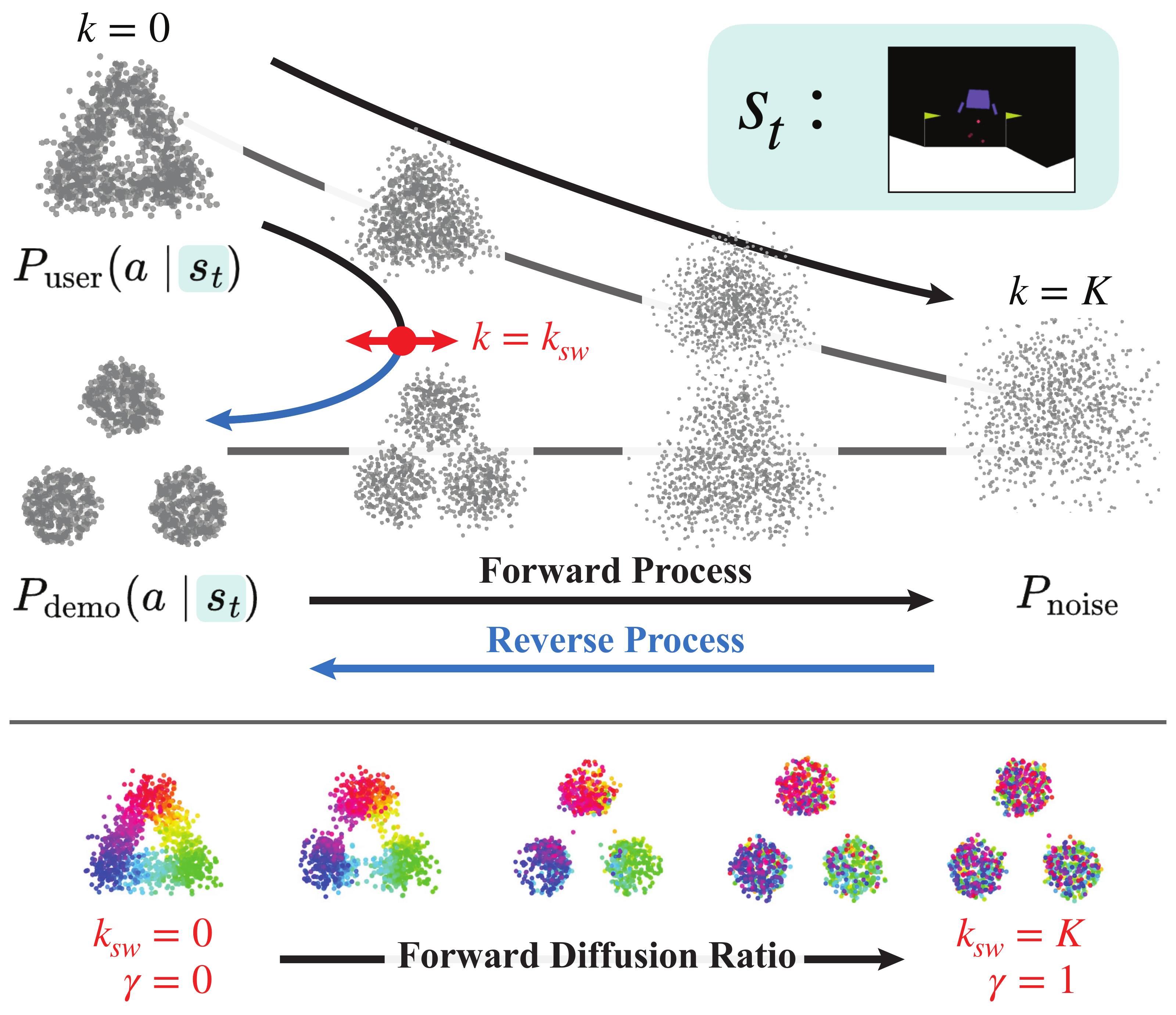}
    \caption{(Top) A visualization of diffusion processes of action distributions at state $s_t$. The black arrow at the top shows the forward diffusion of a source user distribution $P_\textrm{user}$ and the blue arrow below shows the reverse diffusion to the target demonstration distribution $P_\textrm{demo}$. We switch these two processes in the intermediate step $k=k_\textrm{sw}$ to achieve partial forward and reverse diffusion shown in the black and blue arrow on the left. (Bottom) The result of forward and reverse diffusion for different switching times $k_\textrm{sw}$, where standard reverse diffusion process corresponds to $k_\textrm{sw} = K$.}
    \label{fig-diffusha:diffusion-visualization}
\end{figure}

An integral part of our shared autonomy approach is the utilization of diffusion models as a generative model that serves to correct the user actions.
We elaborate our methodology in the sections below.

\subsection{Diffusion model guidance for distribution transformation}
\label{subsec-diffusha:fwd-and-rev}
Consider generally the problem of transforming one distribution $P_\src$ onto another distribution $P_\tgt$. This problem is challenging, because finding such a correspondence between distributions has an inherent dependence on the intrinsic geometrical match between the two distributions. However, there is an enticing shortcut that presents itself in the form of diffusion processes. In particular, we note that the forward diffusion process should allow us to map $P_\src$ into a noise distribution $P_\text{noise}$ (i.e., zero-mean isotropic Gaussian noise). Drawing samples from this noise distribution, we can execute the reverse diffusion process to guide the sample to being drawn from $P_\tgt$, assuming of course that the reverse process was trained on data samples from $P_\tgt$.

Although this diffusion process will ultimately take points $\bm{x}_\src$ from $P_\src$ and map to samples drawn from $P_\tgt$, resulting in $\hat{\bm{x}}_\tgt$, it will do so in a destructive manner. In particular, as the forward diffusion process is applied to $\bm{x}_\src$, the distribution of these points becomes progressively more random (by design). In the end, its distribution reaches an isotropic Gaussian distribution, completely corrupting the information in the original sample. At this point, there is no relationship between the source points $\bm{x}_\src$ and the derived points $\hat{\bm{x}}_\tgt$. We may as well have simply started from Gaussian noise and run reverse diffusion onto $P_\tgt$.

We would like to learn a transformation $\mathcal{F}: \bm{x}_\src \to \hat{\bm{x}}_\tgt$ that preserves information. Such a transformation between distributions can be achieved by running the forward diffusion process on $P_\src(x)$ \textit{partway through}, up to $k=k_\textrm{sw}$, followed by the reverse diffusion process for the same number of steps ($k_\textrm{sw}$), in order to transform these partially diffused points onto $P_\tgt$.
Crucially, this process trades off the amount of information present in $\bm{x}_\src$ that is preserved as a result of forward diffusion, and the consistency (in terms of likelihood) of the sample generated via reverse diffusion with respect to the target distribution $P_\tgt$ (i.e., the distribution over desired behaviors).

Very recently, \textcite{sdedit} proposed a similar partial diffusion process for the task of generating realistic images based upon a user-provided sketch. The authors derive a bound on the distance between $\hat{\bm{x}}_\tgt$ and $\bm{x}_\src$ as follows\footnote{Their bound is for a variance exploding (VE) formulation of diffusion, whereas we employ DDPM that uses a variance preserving (VP) formulation. Despite the difference, they share the same mathematical intuition.}.
Assuming that
$\lVert \bm{\epsilon}_\theta(\bm{x}, k) \rVert \leq \mathcal{K}$
for all $\bm{x}\in X$ and $k \in [0, K]$, then for all $\delta \in (0, 1)$ with probability at least $(1-\delta)$,
\begin{equation}\label{eqn:displacement-bound}
    \lVert \bm{x}_\src - \hat{\bm{x}}_\tgt \rVert^2_2 \leq \sigma_{k_\text{sw}}^2 (\mathcal{K} \sigma_{k_\text{sw}}^2 + d + 2 \sqrt{-d \cdot \log \delta} - 2 \log \delta),
\end{equation}
where $d$ is the dimensionality of $\bm{x}$. %
From this bound, we can see that the distance between $\bm{x}_\src$ and $\hat{\bm{x}}_\tgt$ increases with $k_\textrm{sw}$, since $\sigma_{k_\textrm{sw}}$ increases with $k_\textrm{sw}$ while other terms are not affected. This expression allows us to bound the difference between the action generated by the assistant $\hat{\bm{x}}_\tgt$ and that of the user $\bm{x}_\src$.

Figure~\ref{fig-diffusha:diffusion-visualization} visualizes this property in two dimensions. 
Here, temporarily ignoring state condition, samples from $P_\textrm{user}$ (i.e., source distribution $P_\src$) form the shape of a triangle, and 
samples from $P_\textrm{demo}$ (i.e., target distribution $P_\tgt$) follow a distribution with three modes, each centered on one of the vertices of a triangle. The black and blue arrow represents the idea of executing partial diffusion---running the forward (black) and reverse (blue) processes for $k_\textrm{sw}$ steps.
At the bottom of Figure~\ref{fig-diffusha:diffusion-visualization}, we visualize the result of partial forward and reverse diffusion for different values of $k_\textrm{sw}$. Here, we color points based on their original spatial location in $P_\src$ (e.g., green points in the lower-right, red at the top, and blue at the lower-left), and track their location over different degrees of forward and reverse diffusion. Based on the visualization of the resulting distributions, we see that the distance between the initial points $\bm{x}_\src$ and $\hat{\bm{x}}_\tgt$ increases with $k_\textrm{sw}$, consistent with Equation~\ref{eqn:displacement-bound}. More generally, we find that:

\noindent\emph{\textbf{When $k_\textrm{sw}$ is small}}
\begin{itemize}
    \item The original information is well-preserved (small displacements; high \emph{fidelity})%
    \item The obtained distribution is far from $P_\tgt$ (low \emph{conformity})
\end{itemize}

\noindent\emph{\textbf{When $k_\textrm{sw}$ is large}}
\begin{itemize}
    \item The original information is corrupted (large displacements; low \emph{fidelity})%
    \item The obtained distribution is close to $P_\tgt$ (high \emph{conformity})
\end{itemize}

To discuss the effect of $k_\textrm{sw}$ independent of the number of diffusion steps $K$, we herein define the \textit{Forward Diffusion Ratio} as 
$\fwr \coloneqq k_\textrm{sw} / K$, and will refer to this throughout this chapter.

\input{diffusha/alg1}

A similar idea in diffusion models has been adopted in image generation,%
manipulation, and 3D geometry generation~\parencite{sdedit, sjc, dreamfusion,
ddnm}. %

\subsection{Distribution transformation for shared autonomy}
\label{diffusha-subsec:task-setting}

\begin{figure*}[!th]
    \centering
    \begin{subfigure}[b]{0.24\textwidth}
        \includegraphics[height=2.7cm]{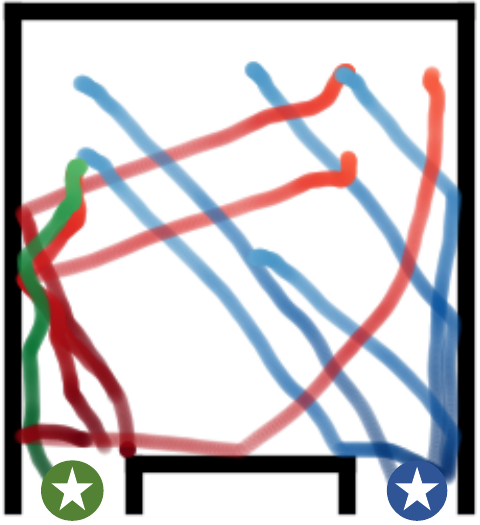}
        \caption{2D Control}
        \label{fig:maze_env_render}
    \end{subfigure}
    \hfill
    \begin{subfigure}[b]{0.24\textwidth}
        \includegraphics[height=2.7cm]{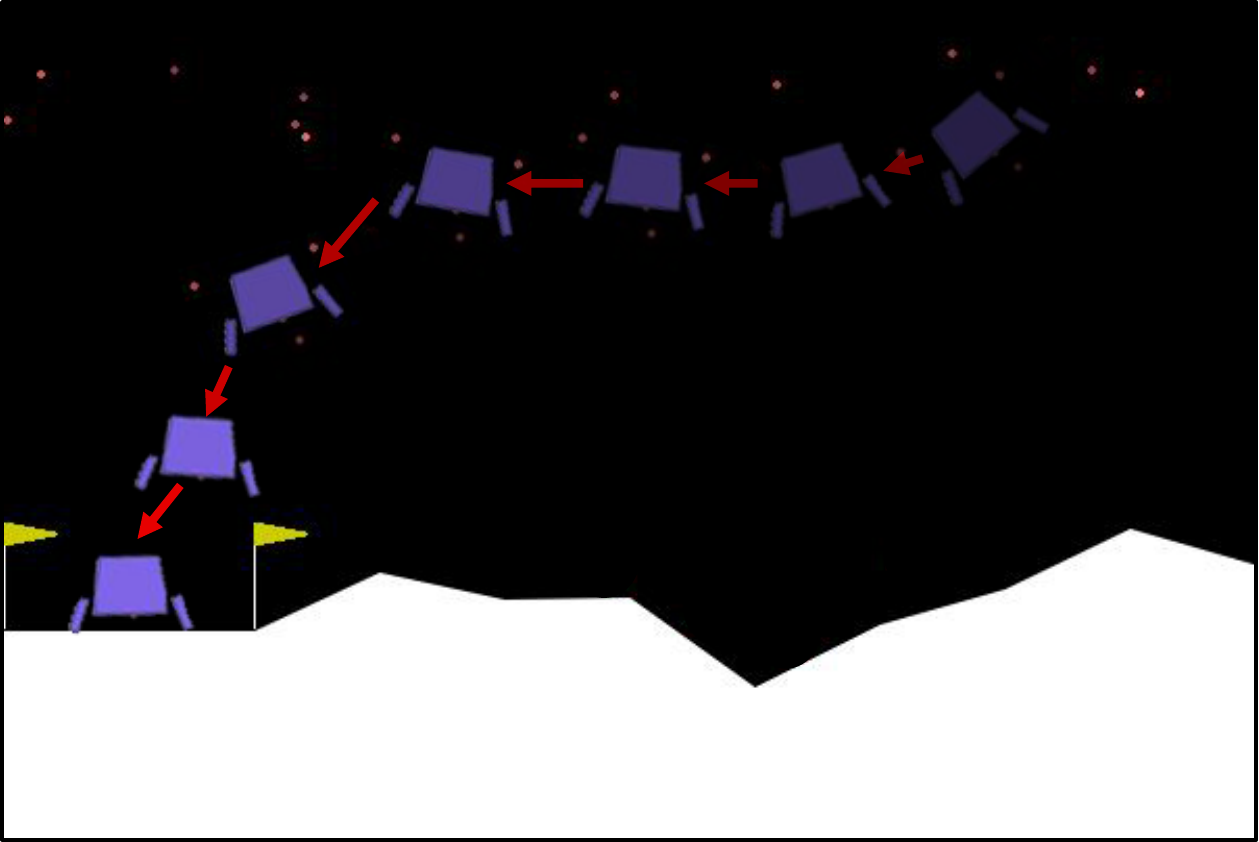}
        \caption{Lunar Lander}
        \label{fig:ll-lander_env_render}
    \end{subfigure}
    \hfill
    \begin{subfigure}[b]{0.24\textwidth}
        \includegraphics[height=2.7cm]{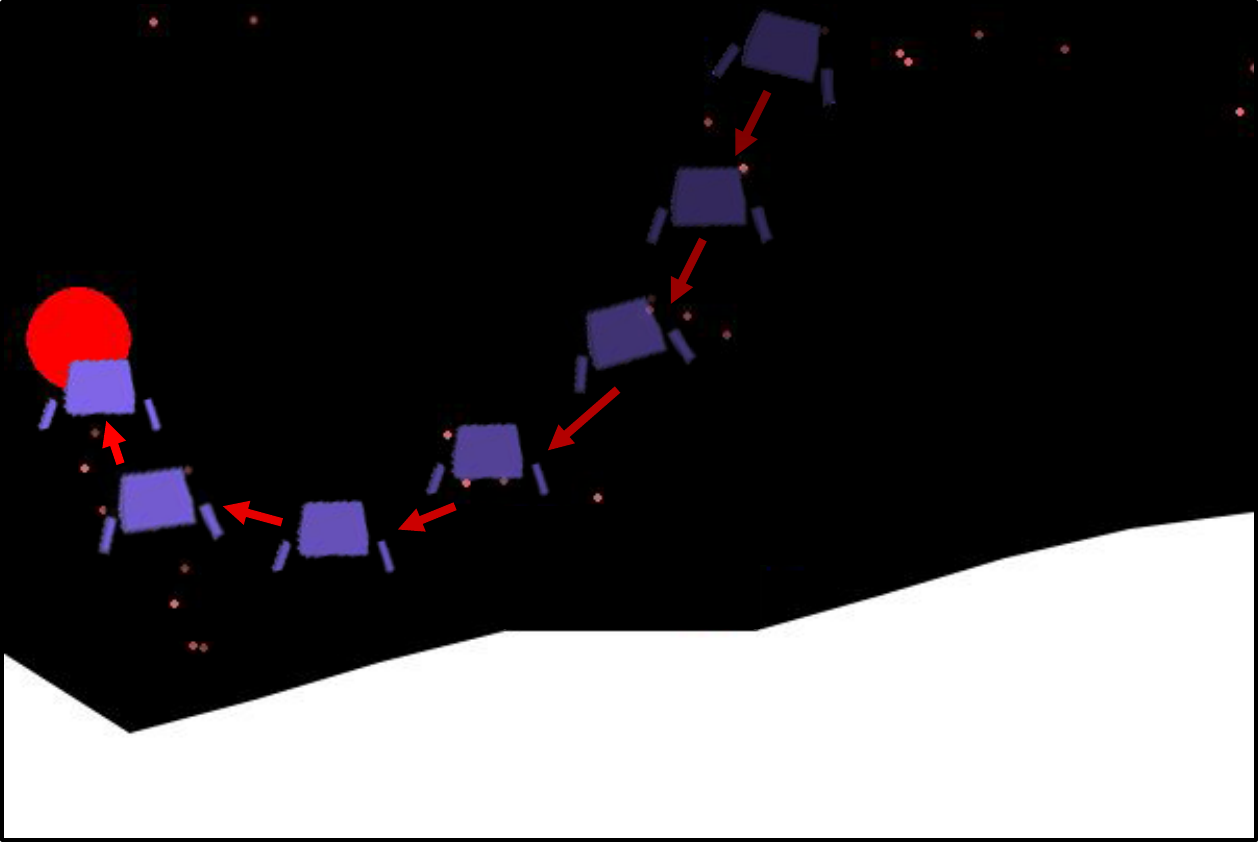}
        \caption{Lunar Reacher}
        \label{fig:ll-reacher_env_render}
    \end{subfigure}
    \hfill
    \begin{subfigure}[b]{0.24\textwidth}
        \includegraphics[height=2.7cm]{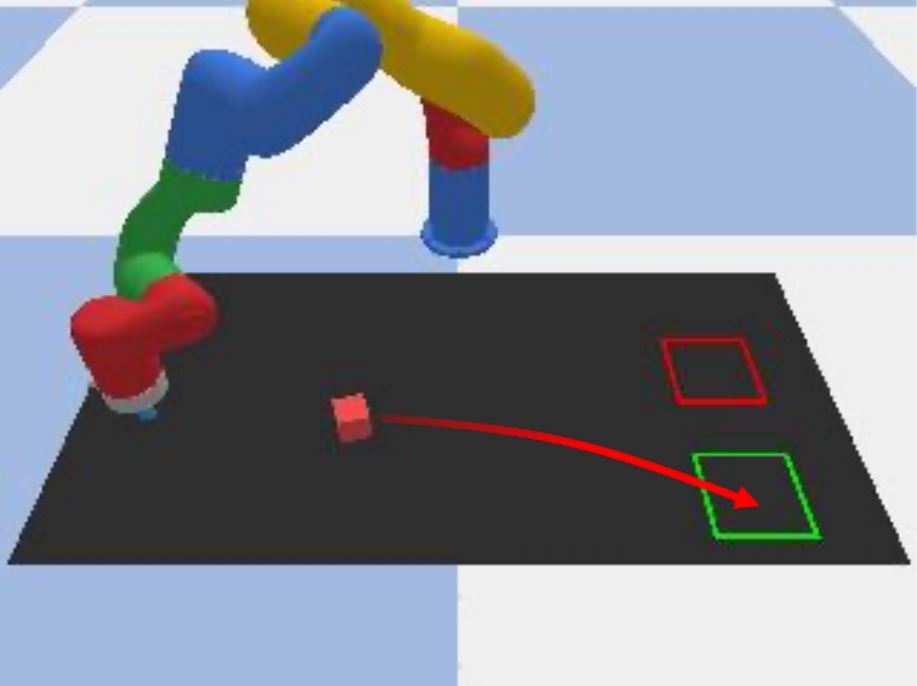}
        \caption{Block Pushing}
        \label{fig:bp_env_render}
    \end{subfigure}
    \caption{We evaluate our algorithm in the context of four shared autonomy environments including (a) a 2D Control task in which an agent navigates to one of two different goals, (b) Lunar Lander that tasks a drone with landing at a designated location, (c) a Lunar Reacher variant in which the objective is to reach a designated region in the environment, and (d) Block Pushing, in which the objective is to use a robot arm to push an object into one of two different goal regions.}
    \label{fig-diffusha:four_exp_env}
\end{figure*}

In a shared autonomy task, a copilot is asked to produce a shared action $a^s_t$, 
given the current state $s_t$ and the pilot's action $a_t^h$.
The goal is for the copilot to intervene in a manner that preserves the pilot's ``intention'' while correcting the action. 
However, formulating the pilot's ``intention'' is challenging, especially when the space of the goals is unknown or not well-defined.

Our algorithm assumes that the latent intent of the user has non-zero likelihood under the distribution over target behaviors that we learn to sample from using a diffusion model. In other words, if we have a finite set of demonstrations to train the diffusion model, the states or state-action pairs that a user would want to achieve need to exist in the demonstrations.
With our formulation to shared autonomy, the copilot manages the trade-off between respecting pilot's action (i.e., \emph{fidelity} to the pilot) and executing an action that is likely under the learned behavior distribution (i.e., \emph{conformity} to the target behaviors).
This trade-off bears similarity to the one discussed in Section~\ref{subsec-diffusha:fwd-and-rev} based on Figure~\ref{fig-diffusha:diffusion-visualization}.

Now, given a state $s_t$, $P_\textrm{user}$ in Figure~\ref{fig-diffusha:diffusion-visualization} is an (unknown) distribution over pilot actions and $P_\textrm{demo}$ is a distribution over target behaviors where each point represents a single action.
We can interpret that each cluster in $P_\textrm{demo}$
corresponds to a set of actions, each of which causes distinct transitions. Hence, each cluster can be thought of as a pilot's ``intention'', 
and the actions outside of the cluster as being undesirable. %
Considering the visualization of the different partially forward and reverse diffused distributions at the bottom of the figure, we see that as we increase $\fwr$ from $0.0$ to $1.0$, most of the pilot's actions begin to go inside of the set of ideal %
actions.
This captures the notion that conformity to the target behavior increases with $\fwr$.
On the other hand, once $\fwr$ exceeds $0.5$, 
the displacement of pilot actions (signified by the mixture of colors)
begin to enlarge, which is equivalent to the copilot producing an action that ignores the pilot's ``intention''. This captures the notion that the fidelity to the pilot decreases as we increase $\fwr$.

Algorithm \ref{alg:gen-fwd-rev} summarizes the procedure of applying partial forward and reverse diffusion to a pilot action $a^h_t$ and generating a shared action $a^s_t$. In Section~\ref{sec-diffusha:experiments}, we investigate the fidelity-conformity trade-off by modulating $\fwr$ empirically for various environments with different pilots.

\section{Experiments}\label{sec-diffusha:experiments}
\input{diffusha/tb-lunar-tasks}
\input{diffusha/fig-maze-over-gamma}
\input{diffusha/fig-lunar-over-gamma}
We evaluate our approach to shared autonomy by pairing our copilot with various pilots on a variety of simulated continuous control tasks (Fig.~\ref{fig-diffusha:four_exp_env}): 2D Control, Lunar Lander, Lunar Reacher and Block Pushing. Each of these tasks provide the opportunity for the pilot to execute one of several different behaviors that is not known to the copilot. We design each domain to include a randomly sampled target state the pilot intends to reach (herein referred to as the \textit{goal}). 
For example, in Lunar Reacher, a goal location is sampled randomly above ground. Across all tasks, we reveal the goal only to the pilot, by including it as part of the state. The copilot, on the other hand, never has access to the goal. This results in a scenario in which pilot's intention (i.e., the goal) is unknown to the copilot.

We seek to answer the following questions: (1) How much can our copilot assist a pilot? (2) Does our copilot generalize to different pilots? and (3) What is the effect of the forward diffusion ratio $\fwr$, and how can we interpret it?

\subsection{Continuous control domains}

\paragraph{2D Control} We build a simple 2D continuous control domain based on the maze-2D environment in D4RL~\parencite{fu2020d4rl} built on the MuJoCo simulator~\parencite{mujoco}, where the goal is located either at the lower-left or lower-right corners of the large open space (Fig.~\ref{fig:maze_env_render}).
The agent is represented as a point mass, and the actions are 2D forces applied to itself. The state consists of the agent's location and velocity, as well as the goal location. Episodes are terminated if a timeout of $300$ steps is reached.

\paragraph{Lunar Lander} Lunar Lander (Fig.~\ref{fig:ll-lander_env_render}) is a continuous control environment adopted from Open AI Gym \parencite{brockman16} that involves landing a spaceship on a landing pad. The actions consist of continuous left, right and upward forces that emulate thrusters on the spaceship.
The state contains the position, orientation, linear and rotational velocity of the spaceship, whether each leg touches the ground, and the landing pad location (provided only to the pilot).
An episode ends when the spaceship lands on the landing pad and becomes idle, crashes, flies out of bounds, or it reaches a timeout of $1,000$ steps.

\paragraph{Lunar Reacher} A variant of Lunar Lander adopted from previous work~\parencite{rsa, brockman16}, where the goal is not to land, but to reach a random target location above the ground (Fig.~\ref{fig:ll-reacher_env_render}). The setting otherwise matches that of Lunar Lander.

\paragraph{Block Pushing} A variant of the Simulated Pushing environment~\parencite{ibc}. The environment (Fig.~\ref{fig:bp_env_render}) consists of a simulated six-DoF robot xArm6 in PyBullet~\parencite{coumans2021} equipped with a small cylindrical end effector. The task is to push an object into one of two target zones in the robot's workspace. The episode terminates when the target reaches one of the two locations or the number of steps exceeds a timeout of $100$. The position and orientation of the block and end effector are randomly initialized at the start of each episode, while the target locations are fixed.

\subsection{Training}
To understand the pilot's intent, our copilot relies on a state-conditioned diffusion model $p_\theta(a_t \vert s_t)$.
This diffusion model is trained as follows. First, we collect expert demonstrations, each containing a sequence of goal-embedded state-action pairs for each domain, where the goal is randomly sampled at each episode.
We then use the resulting demonstrations to train a state-conditioned diffusion model using the DDPM loss.
To hide a goal from our copilot, we remove goal locations from each observation prior to training.

The following experiments use a single diffusion model trained separately for each task. We note that the copilot's behavior changes only according to $\fwr$ given a diffusion model.

To collect demonstrations, we first train an expert policy with soft actor-critic~\parencite{sac}
 for $3\text{M}$ timesteps in Lunar Lander and Lunar Reacher, and
$1\text{M}$ timesteps in Block Pushing.
We then roll out the policies in each environment to collect demonstrations of state-action pairs $D_\textrm{expert}$ for various goals.

\subsection{Surrogate pilots for evaluation}\label{subsec:pilot-and-copilot}
Our method does not require access to a pilot (surrogate or otherwise) when training the copilot. However, surrogate pilots are useful when we want to perform a large number of evaluations
in a reproducible manner.
Thus, we prepare two surrogate pilots consistent with previous work~\parencite{sha-via-deeprl,rsa}: a \emph{Laggy} pilot and a \emph{Noisy} pilot, both of which are corrupted versions of a single expert.
At each time step, the Laggy pilot repeats its previous action with probability $p_\textrm{laggy}$, and otherwise executes an action drawn from the expert's policy. With probability $p_\textrm{noisy}$, the Noisy pilot samples an action from a uniform distribution over the action space, and otherwise executes an action sampled from the expert policy. We evaluate our shared autonomy algorithm for pilots across a broad range of parameters $p_\textrm{noisy}$ and $p_\textrm{laggy}$. For the purpose of this thesis,
we only include results for a representative subset of the pilot parameters for Lunar Lander ($p_\text{laggy} = 0.85,~p_\text{noisy} = 0.3$), Lunar Reacher ($p_\text{laggy} = 0.85,~p_\text{noisy} = 0.6$), and Block Pushing ($p_\text{laggy} = p_\text{noisy} = 0.6$).

\subsection{Various pilots with our copilot}
We evaluate our copilot with various surrogate pilots in all four environments.
Table \ref{tb:lunar-lander-reacher} shows success and crash/out-of-bound rates when various pilots are paired with our copilot in Lunar Lander and Lunar Reacher.
We see that adopting our copilot significantly improves the success and crash rates for all but the expert policy (as expected).
We note that each task involves reaching or landing on a randomly sampled target that is not known to the copilot. Consequently, one can not expect the copilot to have significant effect on the success rate of the Random or Zero pilot. Nevertheless, the results show that our copilot improves their success rates with performance similar to that of full diffusion ($\fwr = 1.0$) and significantly decreases the rate at which the pilots crash.

\subsection{The effects of the forward diffusion ratio $\fwr$}
The forward diffusion ratio $\fwr$ determines how many steps of forward diffusion process to apply on pilot's action $a^h_t$. 
This value changes the balance of interpolating between source and target distributions, which in turn controls the trade-off between fidelity and conformity of an action.
In this section, we demonstrate this trade-off with different $\fwr$ values in various environments.
We deployed our copilot with various surrogate pilots as discussed in Section \ref{subsec:pilot-and-copilot} for each $\fwr$ value.

Figure~\ref{fig-diffusha:maze-vs-fwr} shows the effect of $\fwr$ for the  2D Control domain. Although the copilot is trained on demonstrations for both goals, the pilot's goal is fixed to the left. As $\fwr$ increases, we see that the number of trajectories that reach the correct goal increases. Qualitatively, we also see that the stochasticity of the trajectories lessens, and the paths towards the goal become smoother. However, when $\fwr$ is too high, the copilot ignores the pilot's actions. Without knowledge of the pilot's goal, the copilot ends up distributing trajectories evenly between the left and right goals, consistent with the set of trajectories on which the model was trained.

Figures~\ref{fig-diffusha:ll-land-assisted} summarizes the results on Lunar Lander and Lunar Reacher, respectively, by categorizing the trajectories into success, crash and float. Here, float denotes a trajectory where the spaceship 
does not crash nor go out of bounds, but nevertheless fails to complete the task within the time limit.

In all cases, we observe that the success rate follows a distinct pattern. First, the success rate monotonically increases with $\fwr$ as the copilot improves task performance. After reaching a peak success rate at a critical value of $\fwr$, the success rate gradually decreases.
The drop in the success rate reflects the copilot's infidelity to the user's intent. The pilot's original actions provide a signal of which goal or landing pad is correct. When $\fwr$ is too large, the copilot totally ignores the pilot's actions and instead chooses to land in a manner following the copilot's original training distribution.

In general, the crash rate monotonically decreases with $\fwr$, with the exception of Lunar Reacher. This demonstrates that the quality of the generated actions improves as $\fwr$ increases.
In Lunar Reacher, many of the demonstrations go straight to a target, which is often placed at the edge of the the environment. Consequently, if the agent overshoots and misses the target, it is very likely that it will go out of bounds, which is counted as a crash. This is likely why increasing $\fwr$ does not necessarily result in lower crash rates for Lunar Reacher.

\section{Real Human User Experiments}
\label{subsec:real-user-exp}
\input{diffusha/tb-lunar-humans}
\input{diffusha/fig-lunar-humans-qualitative}
We conducted a set of human user experiments involving the Lunar Lander and Lunar Reacher continuous control domains. For the experiments, we recruited $17$ participants ($9$ who identified as male, and $8$ as female, with an average age of $25$).\footnote{None of the participants were co-authors or otherwise involved in this research.} We asked each user to interact with one of two different co-pilots within each episode, one corresponding to direct teleoperation (i.e., no assistance) and the other being our diffusion-based shared autonomy assistant. The identity of the co-pilot was not disclosed to the participant. At the beginning of each experiment, we allowed the user to practice for $10$ episodes with each co-pilot. In the subsequent testing phase, the user controlled the system for another $10$ episodes with each of the co-pilots. We conducted this experiment for both Lunar Lander and Lunar Reacher and emphasize that the location of the goal (the landing pad for Lunar Lander and goal region for Lunar Reacher) was chosen randomly from the continuous space of goals for each episode.

We consider both the quantitative and qualitative performance of our shared autonomy algorithm. Quantitatively, Table~\ref{tb-diffusha:human-lunar-lander-reacher} compares the average performance of human pilots on Lunar Lander and Lunar Reacher in terms of success, crash/out-of-bounds, and float (i.e., neither crashing, going out-of-bounds, nor succeeding) with and without the assistance of our diffusion-based shared autonomy. We performed a Welch $t$-test for the success rate of each participant and find $p$-values of $p = 3.363 \times 10^{-7}$ for Lunar Lander and $p = 9.033 \times 10^{-3}$ for Lunar Reacher, based on which we can reject the null hypothesis that our diffusion-based co-pilot does not improve success rate.

Qualitatively, we asked participants to rate their agreement with five statements about whether they felt that each co-pilot behaved in a manner that was ``helpful'', ``consistent'', ``responsive'', ``collaborative'', and ``trustworthy'' using a five-point Likert scale. Figure~\ref{fig-diffusha:human-lunar-lander-reacher-survey} visualizes the survey results for all four combinations of tasks and co-pilots. The results reveal that human users rated our shared autonomy co-pilot higher than the no-assistance co-pilot (i.e., teleoperation) with regards to all five qualities for both environments.

\section{Real-Robot Experiments}
\input{diffusha/fig-realrobot-setup}
\input{diffusha/tb-realrobot}
We consider a real-world task (Fig.~\ref{fig-diffusha:real-robot-experiments}) in which a user commands a UR5 manipulator equipped with a pinch gripper to place a donut-shaped object (e.g., a disk with a hole in the center) held by the robot onto one of two posts in the robot's workspace (e.g., analogous to Towers of Hanoi, but with two poles and one disk).

Emulating remote operation, the user observes a real-time (i.e., little-to-no latency) video feed of the robot and its workspace from a third-person view. The user controls the three-dimensional velocity of the end-effector in Cartesian space using a game controller with two joysticks (velocity in the $x$- and $y$-directions using the right joystick, and the $z$-direction using the left joystick). The arm position is initialized to a random location.

As in the human user experiments discussed above, we presented the user with two unidentified co-pilots, one being our diffusion-based shared autonomy assistant and the other being direct teleoperation (i.e., no assistance). We allowed each user to practice with each co-pilot for three episodes. In the subsequent testing phase, participants controlled the arm using both interfaces for five episodes each. We made sure to use the same set of (random) initial arm positions for both test phases. We set the time limit to be $500$ steps (where each velocity command is considered a single step), and treat any episode that exceeds this limit as a failure.
After the testing phase, we asked each participant to rate their agreement with the same quantitative questions in the same manner as the Lunar Lander and Lunar Reacher experiments using a five-point Likert
scale.

We recruited $14$ participants ($10$ who
identified as male, $4$ who
identified as female, average age of $27.7$)\footnote{None of the participants were co-authors or otherwise involved in this research.} for human user experiments with the real robot.

Table~\ref{tb-diffusha:human-ur5} shows the success rate and episode length %
that users took to reach the goal. We found that providing users with the assistance of our diffusion-based shared autonomy algorithm both improved their success rate and decreased the time required to complete the task. The results of the qualitative evaluation reveal that users preferred commanding the manipulator with the assistance of our algorithm.

\section{Related Work} \label{sec-diffusha:related-work}
Shared autonomy has appeared in many problem domains, including remote telepresense~\cite{telepresence01, goertz63, rosenberg93}, assistive robotic manipulation \cite{assistive-manip-01, muelling17, assistive-manip-03}, and assistive navigation~\cite{assistive-navi-01, assistive-navi-02}.
In shared autonomy, one of the most persistent challenges has been correctly identifying the pilot's intentions or goals. 

Early work sidesteps this challenge by assuming a priori knowledge of the pilot's goals ~\cite{users-goal-is-known-01, users-goal-is-known-02}.
Recent work has managed to relax this assumption by treating the pilot's goal as a latent random variable which can be inferred from environmental observations and pilot actions~\cite{infer-user-goal-01, infer-user-goal-02, dragan13, infer-user-goal-04, javdani15, users-goal-is-known-02, muelling14, perez15, infer-user-goal-09}.
In spite of this forward progress, these methods still assume knowledge about some combination of the transition dynamics, the pilot's goals, or pilot's policy, making them difficult to deploy in many unstructured scenarios. 

\textcite{sha-via-deeprl}, and subsequently \textcite{optimizing-interventions-sha, rsa} introduce model-free deep RL to the shared autonomy setting. Because these methods are model free, knowledge of environment dynamics is no longer required, allowing one to train a policy that is not limited to a specific model class. Several follow up works have adapted deep RL to a variety of shared autonomy problems \cite{ave, Reddy2022FirstCU, disagreement-sub-policies, asha}.

To the best of our knowledge, all previous work on shared autonomy with deep RL either explicitly or implicitly assumes a human-in-the-loop setting. In particular, the main training loop typically contains a step to query a pilot (i.e., human user) to obtain its action. This constraint often makes training inefficient or impractical. \textcite{asha} addresses this issue by proposing a two-phase training scheme.
In the first phrase, an autonomous agent learns a task-conditioned policy that can be helpful for assisting a pilot. This is followed by the second phase, which incorporates sparse feedback from humans. Although such approach improves the efficiency of the training pipeline, it does not change the fact that these methods inherently rely on human feedback.

Diffusion models \parencite{thermo} have recently been applied to many problems including image generation, image editing, text-conditioned image generation and video generation. In the field of robotics, \textcite{janner2022diffuser} trained a diffusion model over trajectories, demonstrating that diffusion is capable of generating a diverse set of trajectories reaching a goal location. \textcite{anonymous2023imitating} applies diffusion models to imitation learning. Meanwhile, \textcite{diffusion-policy} show that the policy parameterized with diffusion can generate a multi-modal distributions over possible actions. As far as we are aware, ours is the first approach that uses diffusion models in a shared control scenario, where a copilot learns from expert demonstrations and provides guidance by denoising noisy pilot action.

The core idea of our approach is in the partial forward and reverse diffusion that enables us to generate an output that effectively interpolates between the original input and one from target distribution that we train our diffusion models on. Although we independently came up with this idea, others have explored a similar approach for image editing~\parencite{sdedit}. Related, many recent papers have consider running reverse diffusion from some intermediate step rather than pure Gaussian noise~\parencite{Lyu2022AcceleratingDM, sjc, dreamfusion, ddnm}, as we do in this work.

\section{Conclusion} \label{sec-diffusha:conclusion}
In this work, we presented a new approach to shared autonomy based on diffusion models.
Our approach only requires access to demonstrations that are representative of desired behavior, and does not assume access to or knowledge of the user's policy, reward feedback, or knowledge of the goal space or environment dynamics. Integral to our approach is its modulation of the forward and reverse diffusion processes in a manner that seeks to balance the user's desire to maintain control authority with the benefits of generating actions that are consistent with the distribution over desired behaviors.%
We evaluated our copilot on various continuous control environments and demonstrated that our diffusion-based copilot generalizes across a variety of pilots, improving their performance, while preserving their intention. We further presented an analysis of the effects of different degrees of partial diffusion on task performance.
One limitation of our approach is that there is no component that explicitly addresses the likely mismatch in state distributions between $P_\text{pilot}(s)$ and $P_\text{target}(s)$. Intuitively, it is very likely that the target state visitation distribution (i.e., expert demonstrations) is different from that of the pilot. However, our empirical results suggest that this is not a critical limitation, possibly because executing corrected actions tends to encourage the agent to visit states that are close to those visited as part of the expert demonstrations. One means of addressing this is to design a goal-conditioned policy that can navigate itself to an in-distribution state at test time. We leave this for a future work.

%% file: diffusha/alg1.tex
\begin{algorithm}[t]
    \caption{Shared autonomy as partial diffusion}\label{alg:gen-fwd-rev}
    \small
    \textbf{Input:} Observation $\bm{s}_t$ and pilot's action $\bm{a}^h_t$ \\
    \textbf{Output:} Shared action $\bm{a}^s_t$\\
    \textbf{Require:} A pretrained state-conditioned denoising model $\mu_{\theta^{*}}(\bm{a}, k \mid \bm{s})$, forward diffusion ratio $\fwr$, diffusion timestep $K$
    \begin{algorithmic}[1]
        \small
        \State Compute the switching timestep $k_\sw \leftarrow \text{ToInteger} (\fwr K)$
        \State Sample a Gaussian noise $\bm{\epsilon} \sim \mathcal{N}(\bm{0}, \bm{I})$
        \State $\tilde{\bm{a}}^h_{t, k_\sw} \leftarrow \sqrt{\bar{\alpha}_{k_\sw}} \bm{a}^h_t + \sqrt{1 - \bar{\alpha}_{k_\sw}} \bm{\epsilon}$ \Comment{Forward process}
        \State $\bm{a}^s_{t, k_\sw} \leftarrow \tilde{\bm{a}}^h_{t, k_\sw}$
        \For{$k$ in $k_\sw,\dots, 2$}
        \State Sample a noise vector $\bm{z} \sim \mathcal{N}(\bm{0}, \bm{I})$
        \State $\bm{a}^s_{t, k-1} \leftarrow \bm{\mu}_{\theta^{*}}(\bm{a}^s_{t, k}, k \mid \bm{s}_t) + \sigma_k \mathbf{z}$  \Comment{Reverse process}
        \EndFor
        \State $\bm{a}^s_{t, 0} \leftarrow \bm{\mu}_\theta(\bm{a}^s_{t, 1}, 1 \mid \bm{s}_t)$
        \State return $\bm{a}^s_{t, 0}$
    \end{algorithmic}
\end{algorithm}

%% file: diffusha/tb-lunar-tasks.tex
\begin{table*}[ht]
    \centering
    \caption{Success and crash/out-of-bounds (OOB) rates on Lunar Lander and Lunar Reacher for different pilots with ($\gamma = 0.4$) and without assistance. Each entry corresponds to $10$ episodes across $30$ random seeds. Note that the Zero and Random pilots have no knowledge of the goal.}
    \label{tb:lunar-lander-reacher}
    \begin{tabularx}{1.0\linewidth}{lYYYYYYYY}%
        \toprule
        & \multicolumn{4}{c}{\bf Lunar Lander} & \multicolumn{4}{c}{\bf Lunar Reacher}\\
        \midrule
        & \multicolumn{2}{c}{Success Rate $\uparrow$} & \multicolumn{2}{c}{Crash/OOB Rate $\downarrow$} & \multicolumn{2}{c}{Success Rate $\uparrow$} & \multicolumn{2}{c}{Crash/OOB Rate $\downarrow$}\\
        Pilot & w/o Copilot & w/ Copilot & w/o Copilot & w/ Copilot & w/o Copilot & w/ Copilot & w/o Copilot & w/ Copilot\\
        \midrule
        Noisy & 
        $ 20.67 \pm 4.50 $ &  $ \bm{68.00 \pm 5.35} $ &
        $\hphantom{0} 28.33 \pm 2.62 $ &  $ \hphantom{0}\bm{7.67 \pm 2.87} $ &
        $ 14.33 \pm 2.49 $ &  $ \bm{45.33 \pm 3.30} $ &
        $ \hphantom{0}77.33 \pm 3.09 $ &  $ \bm{38.00 \pm 2.94} $\\
        Laggy & 
        $ 21.33 \pm 2.05 $ &  $ \bm{75.00 \pm 3.56} $ & 
        $\hphantom{0} 76.67 \pm 2.49 $ &  $ \hphantom{0} \bm{9.67 \pm 3.86} $ &
        $ 30.67 \pm 5.56 $ &  $ \bm{55.33 \pm 6.13} $ &
        $ \hphantom{0}69.33 \pm 5.56 $ &  $ \bm{31.33 \pm 2.87} $\\
        \midrule[0.1pt]
        Zero &
        $ \hphantom{0}0.00 \pm 0.00 $ &  $ 27.00 \pm 0.82 $ &
        $ 100.00 \pm 0.00 $ &  $ 19.00 \pm 2.94 $ &
        $ \hphantom{0}0.00 \pm 0.00 $ &  $ 19.67 \pm 2.62 $ &
        $ 100.00 \pm 0.00 $ &  $ 58.33 \pm 3.09 $\\
        Random & 
        $ \hphantom{0}0.00 \pm 0.00 $ &  $ 25.00 \pm 4.32 $ &
        $ 100.00 \pm 0.00 $ &  $ 19.33 \pm 5.25 $ &
        $ \hphantom{0}4.33 \pm 1.89 $ &  $ 22.33 \pm 2.87 $ &
        $ \hphantom{0}95.33 \pm 1.70 $ &  $ 59.00 \pm 0.82 $\\
        Expert &
        $ 77.67 \pm 2.62 $ &  $ 78.67 \pm 2.87 $ &
        $\hphantom{0} 12.33 \pm 0.94 $ &  $ \hphantom{0}8.00 \pm 1.63 $ &
        $ 49.33 \pm 4.78 $ &  $ 55.00 \pm 2.16 $ &
        $ \hphantom{0}44.00 \pm 3.56 $ &  $ 31.67 \pm 2.87 $\\   
        \bottomrule
    \end{tabularx}
\end{table*}

%% file: diffusha/fig-maze-over-gamma.tex
\begin{figure}[!th]
    \centering
    \subfloat[$\gamma=0.0$]{\includegraphics[height=3.0cm]{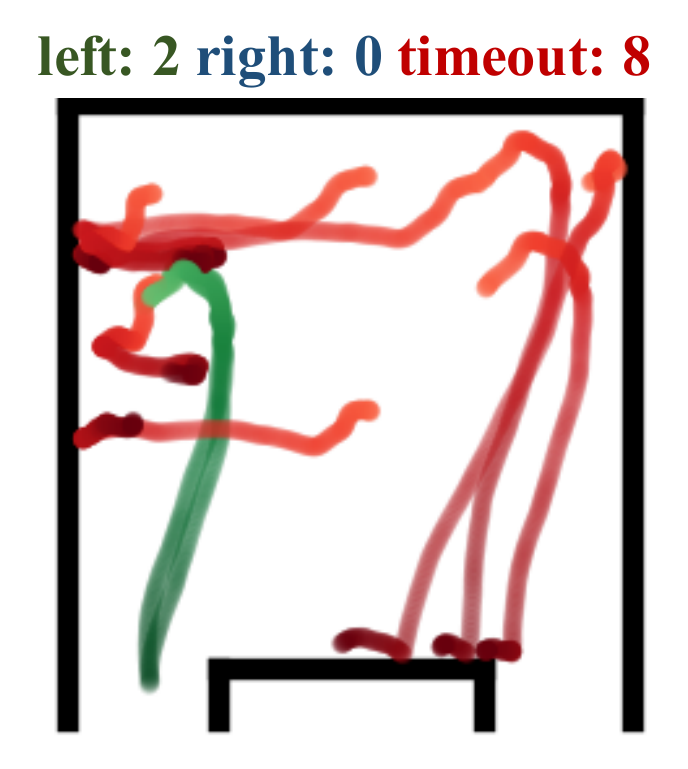}\label{fig:maze0}}
    \hfil
    \subfloat[$\gamma=0.1$]{\includegraphics[height=3.0cm]{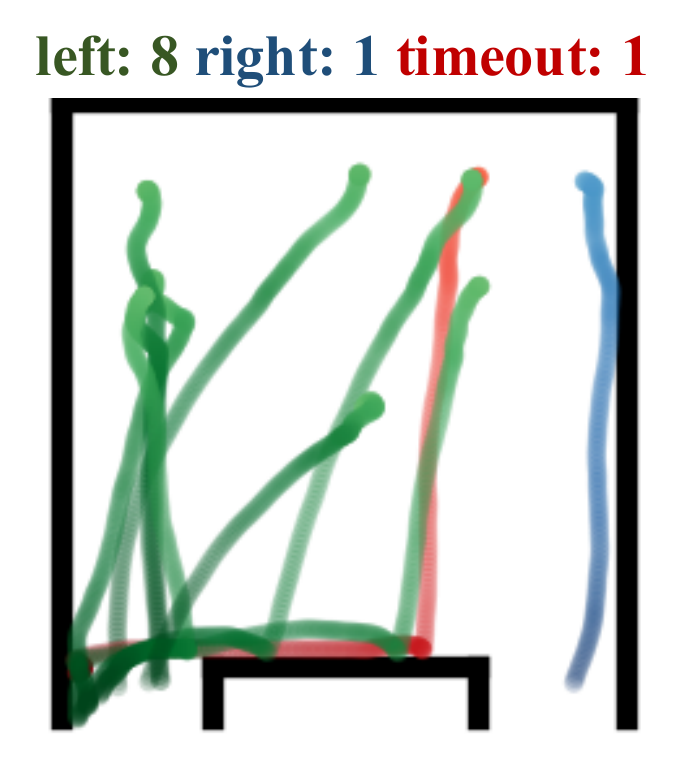}\label{fig:maze1}}
    \hfil
    \subfloat[$\gamma=0.2$]{\includegraphics[height=3.0cm]{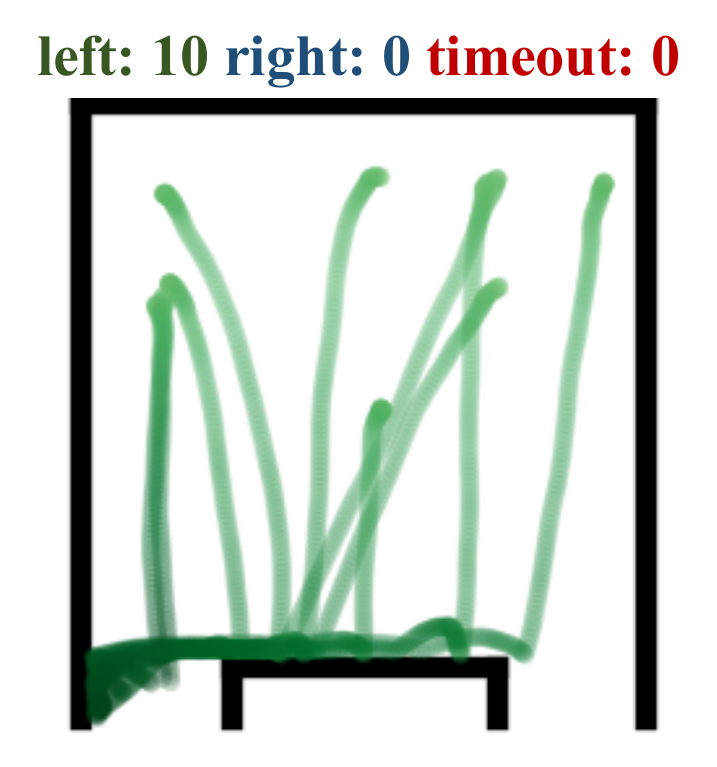}\label{fig:maze2}}\\
    \subfloat[$\gamma=0.4$]{\includegraphics[height=3.0cm]{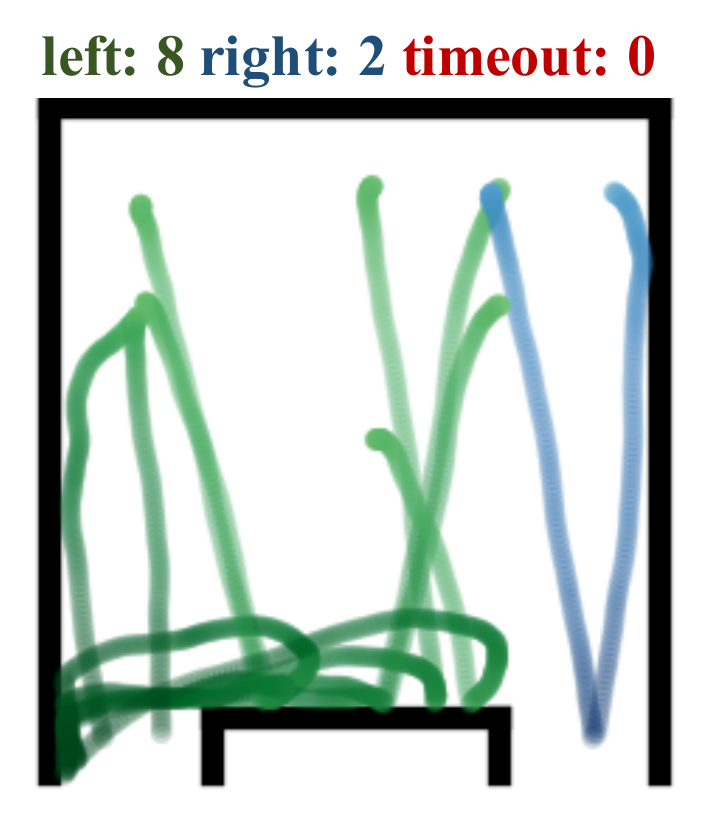}\label{fig:maze4}}
    \hfil
    \subfloat[$\gamma=0.6$]{\includegraphics[height=3.0cm]{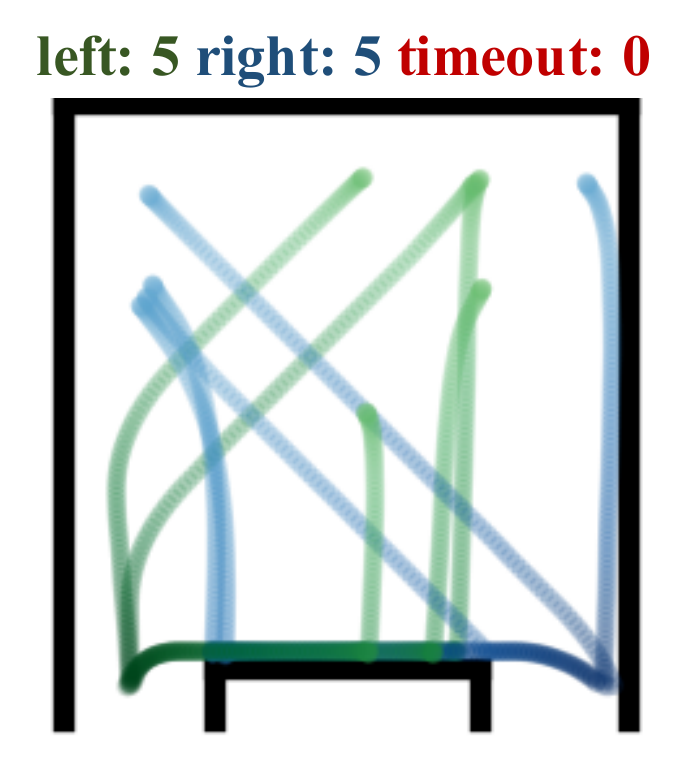}\label{fig:maze6}}
    \hfil
    \subfloat[$\gamma=0.8$]{\includegraphics[height=3.0cm]{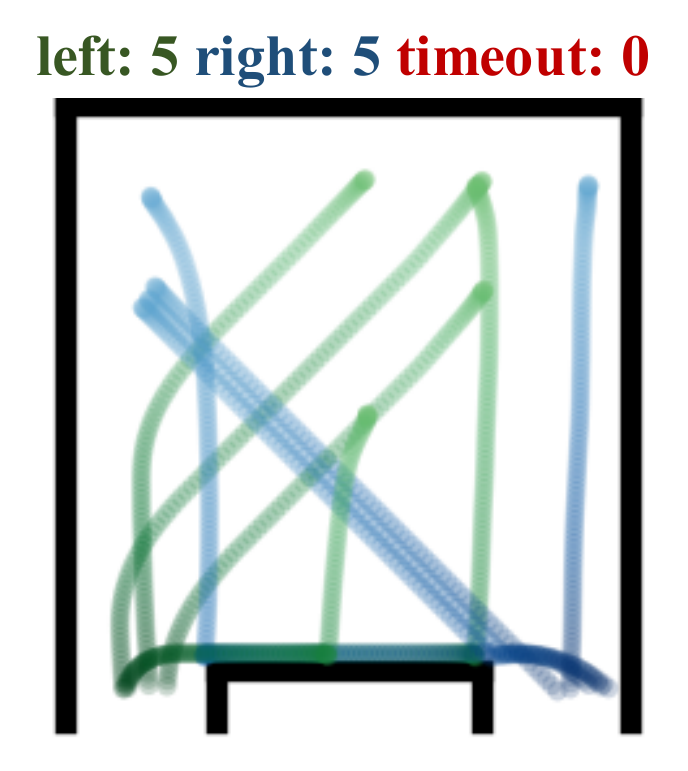}\label{fig:maze8}}
    \caption{A visualization of the resulting trajectories in the 2D Control environment for different settings for the forward diffusion ratio $\fwr$. The user's objective is to reach the left-hand goal. Without assistance \subref{fig:maze0} the user successfully reaches the goal two times, while the eight others timeout. As we increase $\fwr$, we see that \subref{fig:maze1}--\subref{fig:maze4} the user reaches the desired goal a vast majority of the time. As $\fwr$ gets closer to $1.0$, \subref{fig:maze6} \subref{fig:maze8} the assisted policy conforms to the expert policy, which avoids timeouts, but without knowledge of the user's goal distributes the trajectories evenly between the left and right goals.}%
    \label{fig-diffusha:maze-vs-fwr}%
\end{figure}

%% file: diffusha/fig-lunar-over-gamma.tex
\begin{figure}[!t]
    \centering
    \subfloat[Noisy Pilot]{\includegraphics[width=0.48\linewidth]{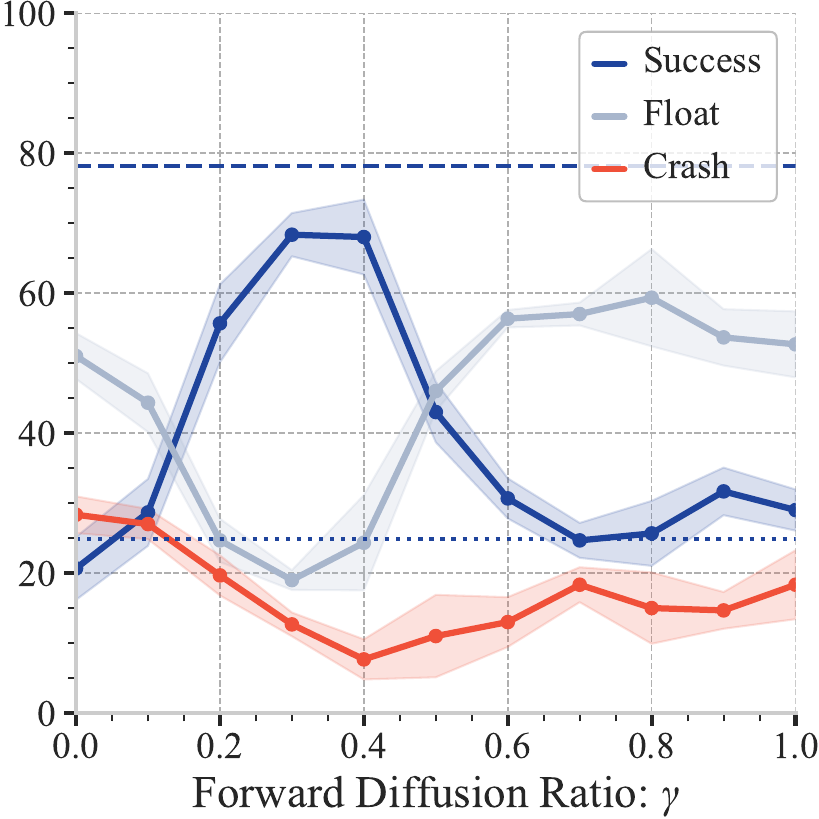}\label{fig:ll-land-assisted-noisy}}
    \hfil
    \subfloat[Laggy Pilot]{\includegraphics[width=0.48\linewidth]{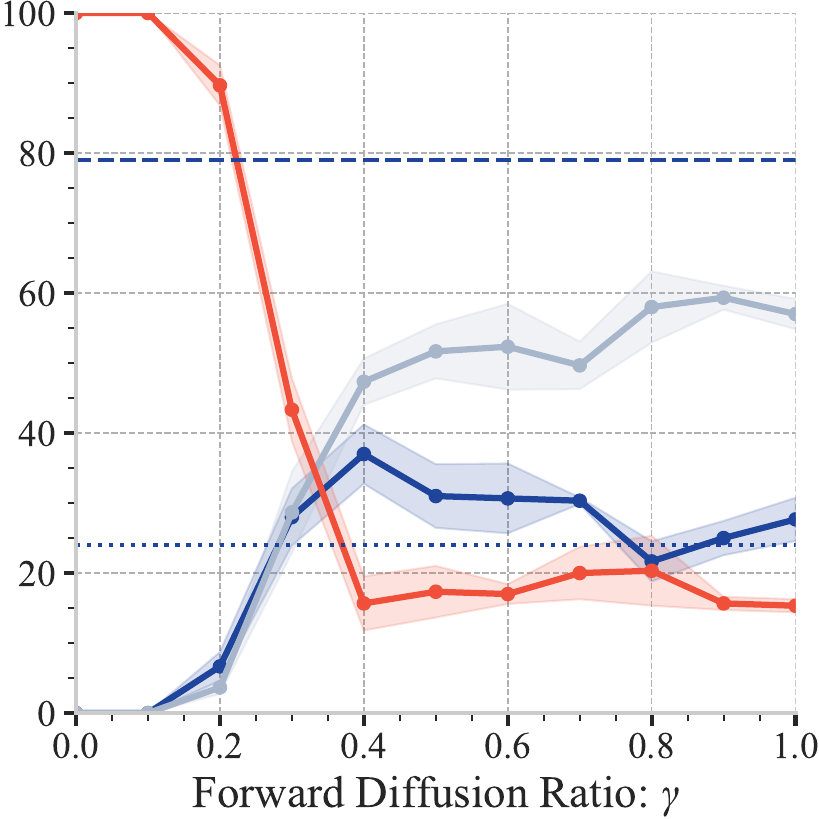}\label{fig:ll-land-assisted-laggy}}%
    \
    \subfloat[Noisy Pilot]{\includegraphics[width=0.48\linewidth]{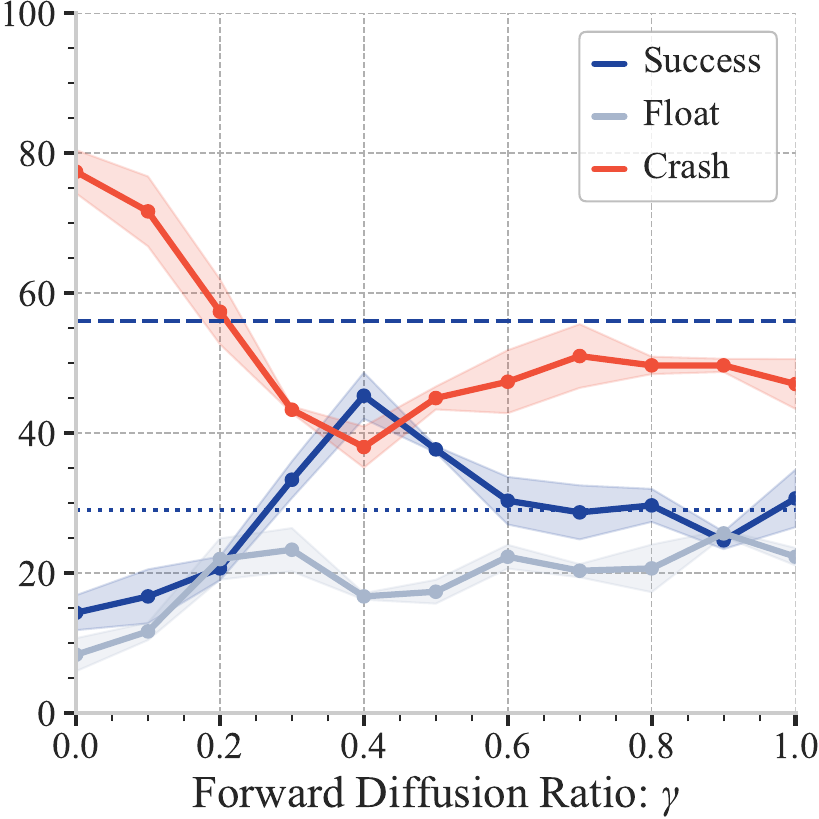}\label{fig:ll-reach-assisted-noisy}}
    \hfil
    \subfloat[Laggy Pilot]{\includegraphics[width=0.48\linewidth]{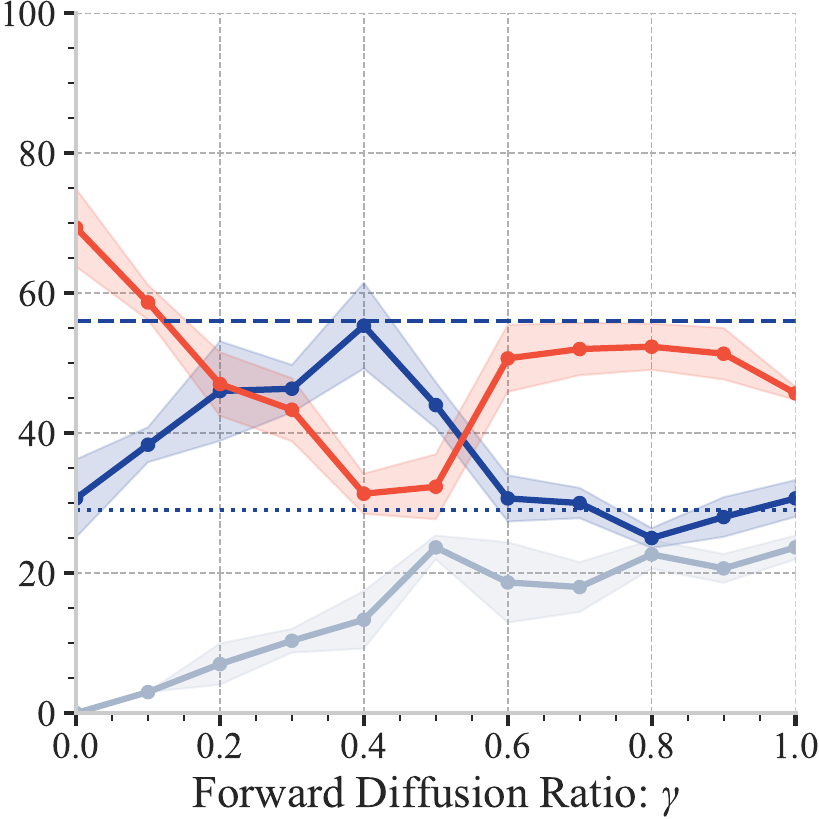}\label{fig:ll-reach-assisted-laggy}}%
    \caption{Success (higher is better), floating, and crash (lower is better) rates for Lunar Lander (top) and Lunar reacher (bottom) with \subref{fig:ll-land-assisted-noisy} \subref{fig:ll-reach-assisted-noisy} noisy and \subref{fig:ll-land-assisted-laggy} \subref{fig:ll-reach-assisted-laggy} laggy pilots. The dashed blue line denotes the success rate of an expert policy, while the dotted blue line denotes the success rate of our model with full-diffusion ($\fwr=1.0$).}\label{fig-diffusha:ll-land-assisted}%
\end{figure}

%% file: diffusha/tb-lunar-humans.tex
\begin{table}[!th]
    \centering
    \caption{Success, crash/out-of-bounds (OOB), and float rates on Lunar Lander and Lunar Reacher for human user experiments.}
    \label{tb-diffusha:human-lunar-lander-reacher}
    \begin{tabularx}{1.0\linewidth}{lYYYYYY}%
        \toprule
        & \multicolumn{3}{c}{\bf Lunar Lander} & \multicolumn{3}{c}{\bf Lunar Reacher}\\
        \midrule
        & Success Rate $\uparrow$ & Crash Rate $\downarrow$ & Float Rate & Success Rate $\uparrow$ & Crash Rate $\downarrow$ & Float Rate\\
        \midrule
        w/o Copilot & $\hphantom{0}1.76$ & $98.24$ & $0.00$ & $17.06$ & $82.94$ & $0.00$\\
        w/ Copilot & $32.35$ & $61.18$ & $6.47$ & $34.71$ & $65.29$ & $0.00$\\ 
        \bottomrule
    \end{tabularx}
\end{table}

%% file: diffusha/fig-lunar-humans-qualitative.tex
\begin{figure*}[!h]
    \centering
    \subfloat[Lunar Lander (w/o Co-pilot)]{\includegraphics[width=0.245\linewidth, trim = 1.3cm 0.1cm 1.3cm 1.3cm, clip]{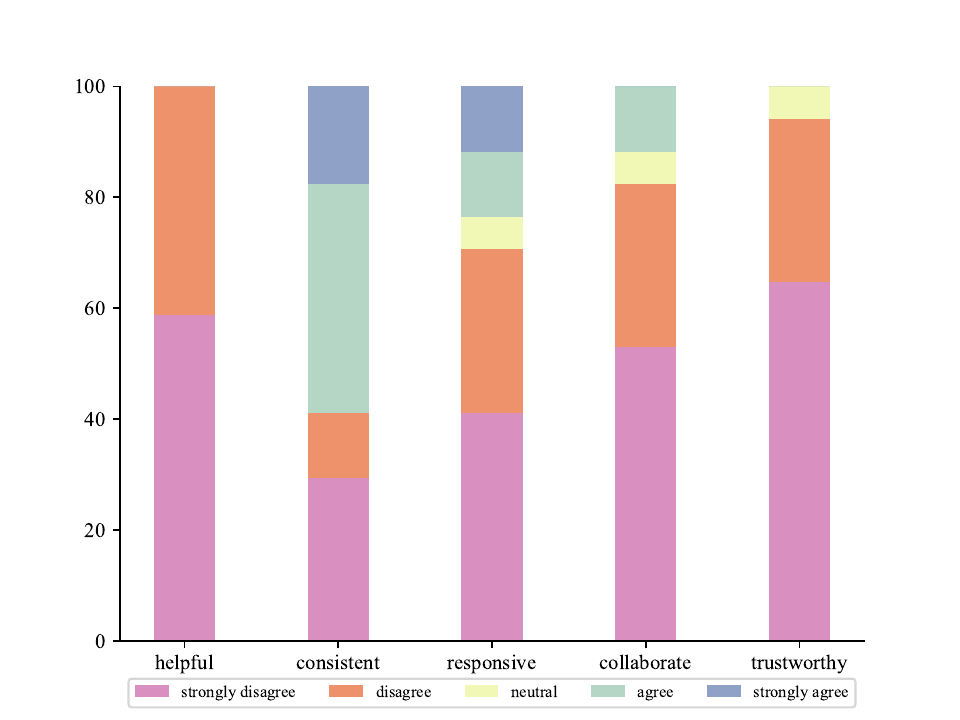}}\hfil
    \subfloat[Lunar Lander (w/ Co-pilot)]{\includegraphics[width=0.245\linewidth, trim = 1.3cm 0.1cm 1.3cm 1.3cm, clip]{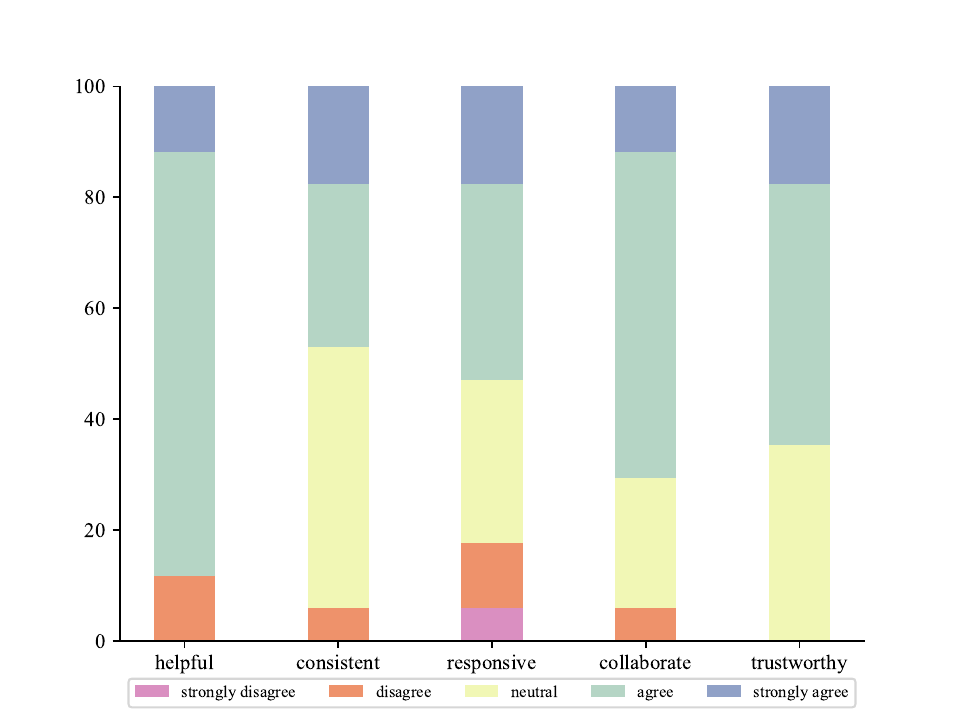}}
    \subfloat[Lunar Reacher (w/o Co-pilot)]{\includegraphics[width=0.245\linewidth, trim = 1.3cm 0.1cm 1.3cm 1.3cm, clip]{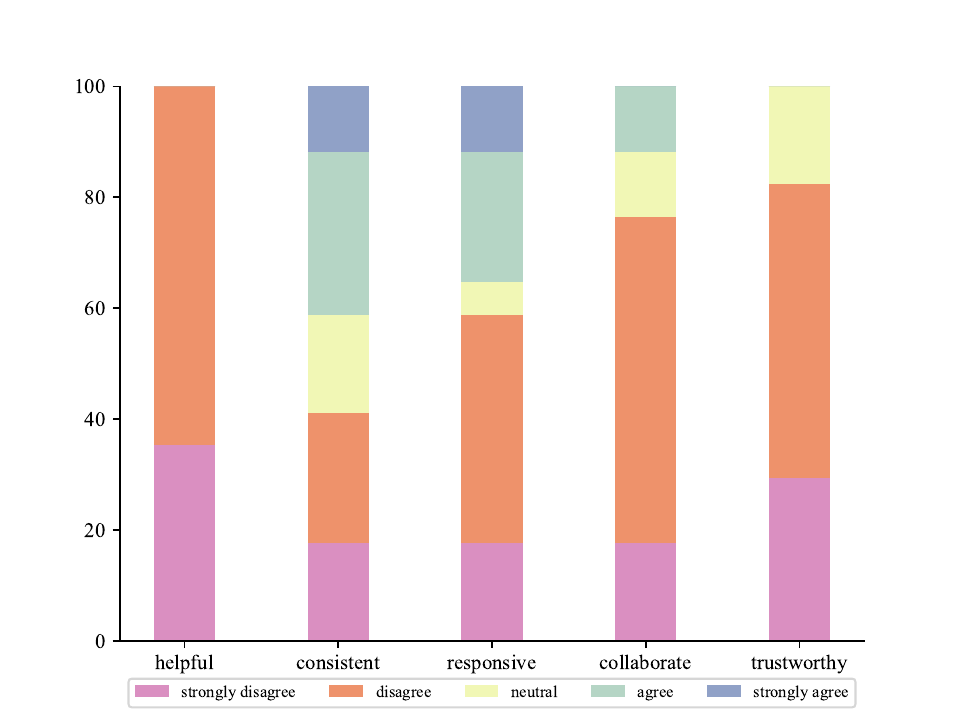}}\hfil
    \subfloat[Lunar Reacher (w/ Co-pilot)]{\includegraphics[width=0.245\linewidth, trim = 1.3cm 0.1cm 1.3cm 1.3cm, clip]{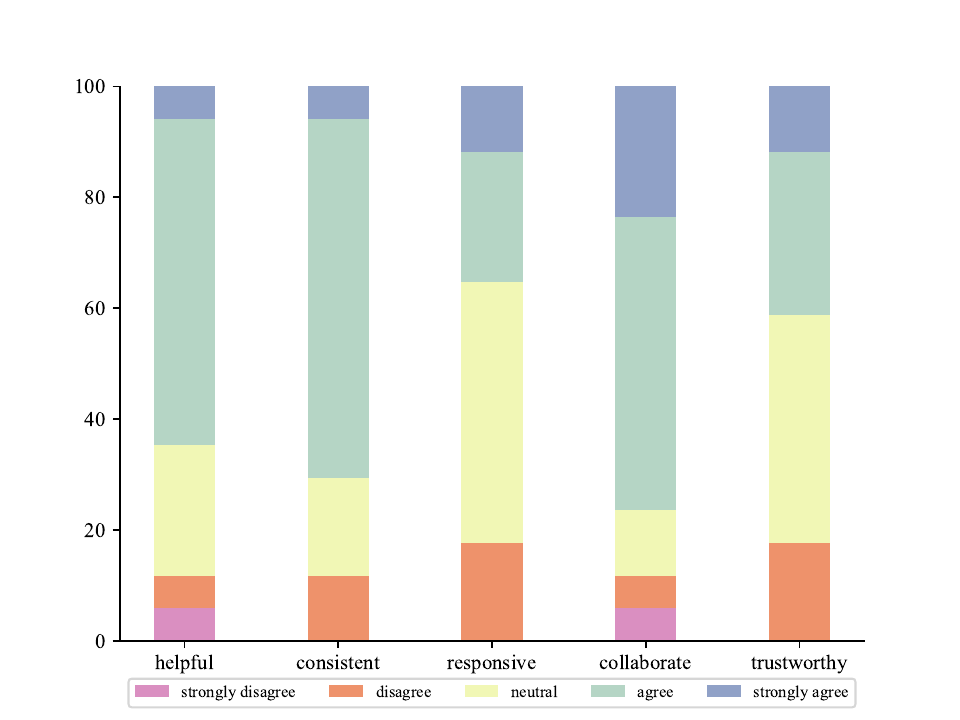}}
    \caption{Results of the human user qualitative surveys for Lunar Lander and Lunar Reacher without and with the assistance of our shared autonomy algorithm.}\label{fig-diffusha:human-lunar-lander-reacher-survey}
\end{figure*}

%% file: diffusha/fig-realrobot-setup.tex
\begin{figure}[!t]
    \centering
    \subfloat{\includegraphics[width=0.48\linewidth]{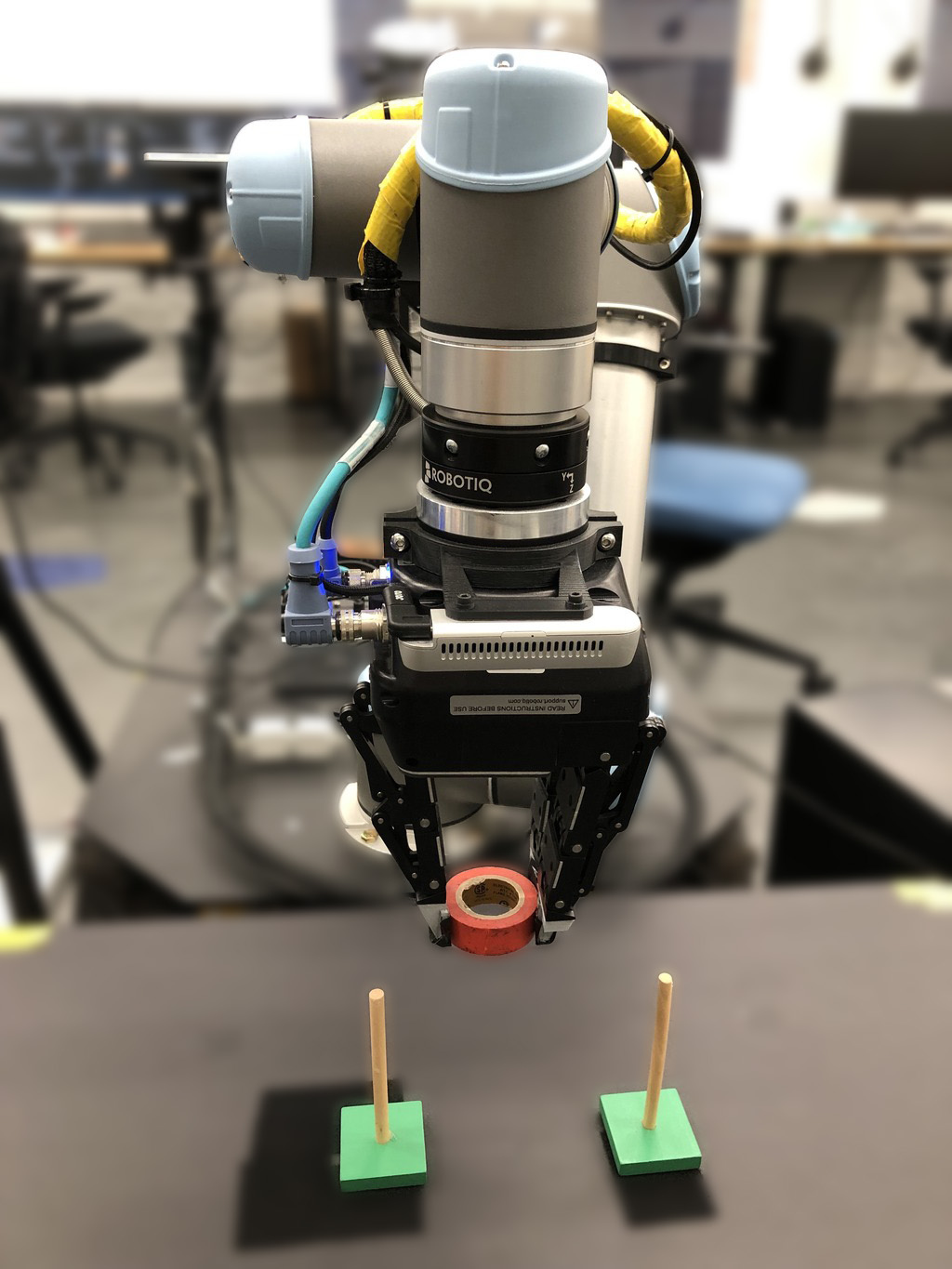}}\hfil
    \subfloat{\includegraphics[width=0.48\linewidth]{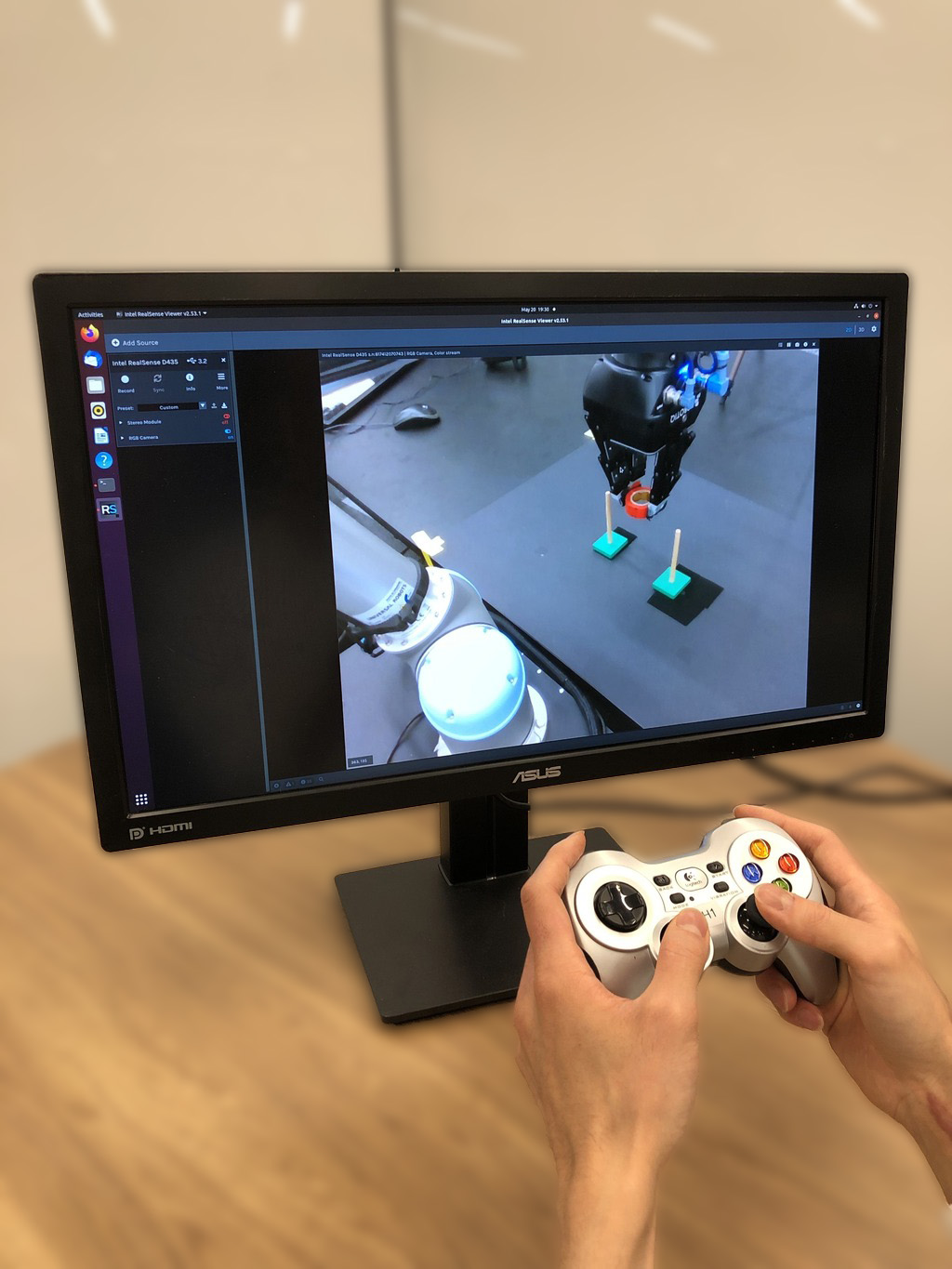}}
    \caption{The real-robot experiments involved (left) placing a disk on one of two posts. Human operators (right) controlled the 3D motion of the end-effector while observing images of the robot's workspace streamed in real-time.}\label{fig-diffusha:real-robot-experiments}
\end{figure}

%% file: diffusha/tb-realrobot.tex
\begin{table}[!t]
    \centering
    \caption{Success rate and episode length with and without copilot for the real-robot experiment with human users.}
    \label{tb-diffusha:human-ur5}
    \begin{tabularx}{0.95\linewidth}{lYY}
        \toprule
        & success rate $\uparrow$ & episode length $\downarrow$\\
        \midrule
        w/o Copilot & $0.89$ & $209.23 \pm 83.71$\\
        w/ Copilot & $1.00$ & $143.66 \pm 68.42$\\
        \bottomrule
    \end{tabularx}
\end{table}

%% file: ila/z_main.tex
\chapter{Unsupervised Domain Adaptation for a Control Policy}
\label{chap:ila}
In this chapter, we aim to address the issue of {\it domain gap} for control policies. In many practical settings, an environment where a control policy is deployed differs from where it was trained, creating a {\it gap} that confuses the policy.

There are different ways to train such a policy, representative ones are imitation learning and reinforcement learning.
Imitation learning assumes the existence of expert demonstrations, and aims to learn the policy from them.
Whereas reinforcement learning allows the agent to interact with the environment that provides reward signals to learn the behavior.
Here, we focus on addressing the domain gap for control policies trained with reinforcement learning, however, there is nothing in our method that prevents it from being applied to imitation learning based policies.

Reinforcement learning for control has achieved great success in a wide variety of challenging sensory-motor control tasks, including agile drone flight~\citep{Kaufmann2018drone,Kaufmann2020drone,Loquercio2021drone}, deformable object manipulation~\citep{Wu2019deformable}, and quadruped locomotion~\citep{Hwangbo2019agile,Miki2022wild,Lee2020terrain,margonlis2022rapid}. In comparison to their classical model-predictive control counterparts, approaches based on RL enable the use of more realistic forward dynamics model in the form of a physics simulator. Improvements in rigid-body simulator technologies~\citep{Todorov2012mujoco, Makoviychuk2021Isaac} allows RL algorithms to overcome their prohibitively-high sample complexity by first training in simulation and then deploying directly on the physical robot. %

Differences, however, still exist between what the controller experiences in the simulator and in the physical environment in the form of a \textit{sim-to-real gap}. In particular, the ability to produce visuomotor control policies that remain robust when perceptual conditions change during deployment, remains an open problem.

\begin{figure}[t!]
    \centering
    \includegraphics[width=\linewidth]{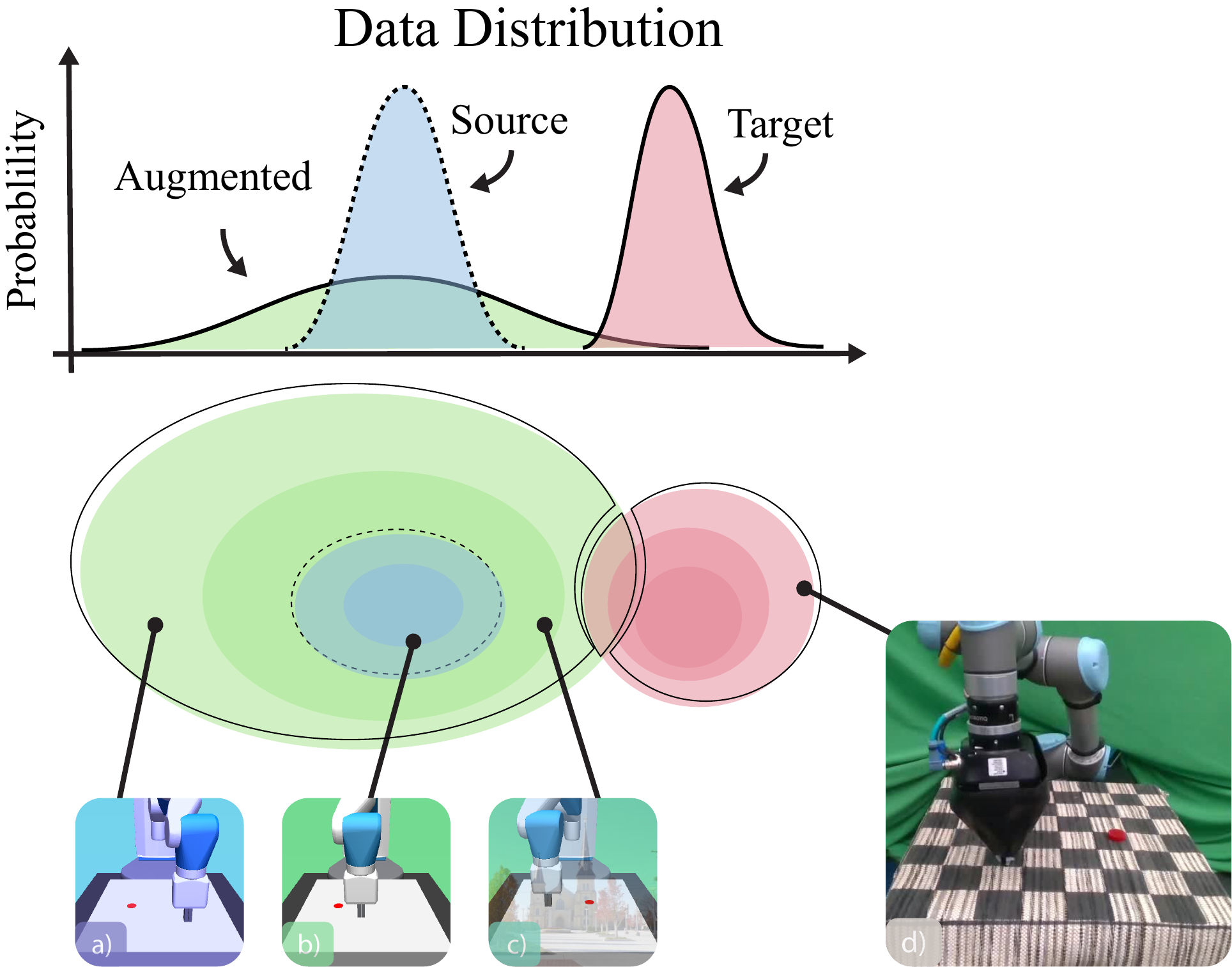}%
    \caption{Data augmentation improves the coverage of the training distribution at the expense of learning complexity and performance. 
    Insets: (a, c) augmented training data; (b) the original input image from the source domain; %
    and (d) the input image in the target domain. Data augmentation fails to provide sufficient coverage of the unknown deployment condition, so the learned controller fails to accomplish the task.}
    \label{fig:out-of-distribution}
\vspace{-1em}
\end{figure}

Consider the illustration in Figure~\ref{fig:out-of-distribution}.
A common approach to battle domain shift is to expose the agent to a large variety of data during training with the hope that the training distribution provides adequate coverage over what the agent will experience in the wild. When the simulator is extensible, one can use domain randomization to generate more diverse training data~\citep{tobin2017domain,active-domain-randomization}. Alternatively, one can use data augmentation mechanisms to decorate existing data~\citep{yarats2021drqv2,hansen2020deployment}. Both approaches aim to produce visual features that are invariant to perceptual changes irrelevant to the task. 
Such invariance does not come for free, however.
Additional training slows down the wall-clock speed of the training process, while domain randomization requires manual tuning~\citep{Andrychowicz2020dexterous} and relies on the assumption that the policy network has sufficient capacity to handle the increased support of the input distribution.
The added complexity can negatively affect model and policy performance~\citep{laskin2020reinforcement,hansen2020deployment}. Furthermore, hidden beneath is the assumption that one needs to know roughly what types of perceptual shift would occur during deployment.
Failure can happen when the target domain is not known \textit{a priori} and falls \textit{out-of-distribution}, resulting in a fumbling robot that is unable to self-correct.

Different from these prior approaches that produce invariance by memorizing what is irrelevant to the task during training, we consider a more challenging, but also more realistic scenario in which the specifics of the deployment is not known in advance. This requires the agent to truly generalize \textit{out-of-distribution}, without prior knowledge of the target environment. We further assume that reward supervision is unavailable, so fine-tuning via RL is out of the question.
This might seem to be an impossible task, but it does implicitly make the assumption that the \textit{same task} that the agent was optimized for during training remains well-defined in the target domain\todo{This may be unclear without an example}. This means that the agent has \textit{some} notion of \textit{what it knows} despite the sudden appearance of many unknowns that are not required for the task of interest.
Without further assumptions or loss of generality, this \textit{out-of-distribution} generalization problem can be formulated as \textit{unsupervised policy adaptation} between two MDPs that share the same latent dynamics and reward structure, but with distinct pixel observations.

In our work, \textbf{\underline I}nvariance through \textbf{\underline L}atent \textbf{\underline A}lignment (\textbf{ILA}) \citep{yoneda22ila}, we investigate ways to improve generalization under this challenging scenario. Rather than battling domain shift by baking perceptual invariance explicitly into the network during training, we demonstrate a way to produce feature invariance at the time of deployment by taking advantage of the fact that the task of interest remains the same, therefore what the agent experiences internally should remain the same as well. We collect latent features collected during training as examples of what the agent \textit{knows} about the task. On the target domain, the new latents are shifted from these prior distributions. Our goal in unsupervised policy adaptation is to match the distribution of these latent features on the target domain with those that appeared during training. This unsupervised learning objective, which we refer to as \textit{latent alignment}, does not require paired image data between the source and the target, and can be applied to any (pre-trained) agent without imposing specific requirements of how it is trained.

\section{Unsupervised Policy Adaptation}
\input{ila/graphical_model}  %
Unsupervised policy adaptation is a setting that involves two distinct domains --- a source domain and a target domain. These two domains share the same underlying MDP and task structure, but has different observation conditions. In the target domain, the agent only has access to observations \(o_t^\text{tgt}\) and its own actions \(a_t\), but not the corresponding rewards or the ground-truth state \(s_t\) (see Figure~\ref{fig:graphicalmodels}). A practical example is a robot that is trained with images in a clean, simulated environment that now has to work in-the-wild, in the presence of visual distractors and changes in the lighting condition or the mounting pose of the video camera. These variations could lead to significantly different image observations. As a result of this shift, deploying an agent trained in the source domain directly in the target domain (i.e., zero-shot transfer) generally results in poor performance.

Formally, we consider an infinite horizon Markov decision process (MDP)~\citep{Puterman2014mdp}  \(\mathcal{M}\) parameterized via the tuple \(\langle S, A, O, R, P, \gamma\rangle\), where \(S\) and \(A\) are the state and action spaces. \(P:S\times A\mapsto S\) is the transition function, \(R: S \times A\mapsto \mathbb R\) is the scalar reward, and \(\gamma\) is the discount factor. The agent receives a stream of observations $o \in O$. We assume a fully-observable setting where a single observation carries enough information to decide an appropriate action. In the source domain \(\mathcal M_\text{src}\) (see Figure~\ref{fig:source_mdp}), we can use reinforcement learning to produce an optimal policy $\pi: O \times A \mapsto [0, 1]$ that maximizes the expected discounted return \(\mathcal J = \mathbb{E}\left[ \sum_\infty \gamma^t R(s_t, a_t)\right]\). In the target domain (see Figure~\ref{fig:target_mdp}) however, the reward is not observable therefore we can not rely on reinforcement learning for fine-tuning. Nevertheless the task structure remains identical to that of the source domain. 
We assume that the policy $\pi$ consists of an encoder $F: O \mapsto Z$, where $Z$ is a compact latent space, and a policy head $\pi_z:Z \times A \mapsto [0,1]$ shown as {\color{darkgreen}green} arrows. The goal of unsupervised policy adaptation is to find ways to battle this distribution shift, so that the resulting, adapted policy can succeed on the task in \(\mathcal M_\text{tgt}\).

\section{Method}
\begin{figure*}[!t]
    \centering
    \includegraphics[width=\textwidth]{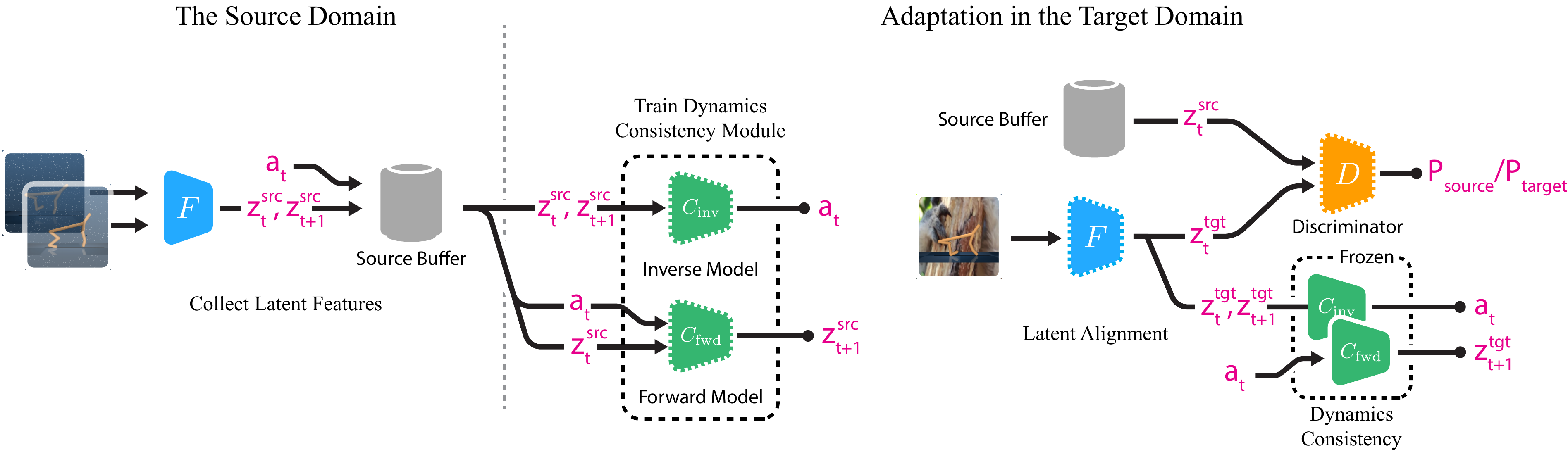}
    \caption{Training and adaptation phases of \lcila. The encoder $F$ takes an observation of target domain, and learns to fool the discriminator, while the discriminator $D$ predicts whether the input is an encoded target observation or a latent sample from source buffer. This adversarial training encourages the distribution of encoder outputs to be similar to the latent embedding sampled from the source buffer. $C_\textrm{fwd}$ and $C_\textrm{inv}$ are the forward and inverse dynamics networks that guides the encoder adaptation.}
    \label{fig:main-fig}
\end{figure*}

When we ask if the agent can perform in the target domain, we are effectively making the assumption that the task, and the underlying MDP has not changed. One way to factorize the problem is to divide the policy into two modules (see Figure~\ref{fig:source_mdp}). A \textit{policy head} \(\pi(a\vert z)\) that we keep \textit{frozen} during the adaptation process, and an \textit{encoder} \(F(z\vert o)\) that we adapt. Ideally, the latent feature \(z\) captures what the policy needs to know to accomplish the task. Hence we can formulate unsupervised policy adaptation as a distribution-matching objective that ``aligns'' the distribution \(P_\pi(z_\text{tgt})\), with the one in the source domain, \(P_\pi(z_\text{src})\).%

This overall \textit{sim-and-adaptation} pipeline starts in the source domain with collecting latent features \(z_\text{src}\) into a buffer. The agent carries these data into the deployment environment. Then during adaptation, it optimizes two objectives. The first is a minimax objective that focuses on individual latent features (\(D\) in Figure~\ref{fig:main-fig}). The second is an cooperative dynamics consistency objective that consists of both the forward and the inverse kinematics prediction error (\(C_\text{inv}\) and \(C_\text{fwd}\) in Figure~\ref{fig:main-fig}). The training procedure partially resembles a generative adversarial network (GAN)~\citep{Goodfellow2014gan} with two key distinctions. First, we do not reconstruct raw pixel observations but instead directly match the distribution of the latent features \(z\). This has the benefit that we do not require a generator that incur additional space and optimization overhead. Second, we find it helpful to pretrain the dynamics consistency module at the beginning of the adaptation process because it reduces the wall-clock time of the procedure.
We will elaborate the details of the procedure below.

\subsection{Collecting Latent Features in The Source Domain}

In the {\color{purple}source domain} we collect latent vectors \(z_t^\text{src}\) and actions \(a_t\) into a buffer
\begin{equation}
\mathcal B_\text{\src} = \{z_0^\text{src}, a_0; z_1^\text{src}, a_1,\dots\}\text{\quad where\quad}  z = F(o_t^\text{src}). %
\end{equation}
The trajectories, \(\tau_\text{src} = \{o_0, a_0; o_1, a_1, \dots\}\) are sampled from \(\mathcal M_\text{src}\) with an exploration policy \(\pi\) that can be different from the pretraind policy for the task. In our experiment we found that using a random policy \(\bar \pi\) that samples a uniformly random action is unexpectedly effective, with the added benefit that the same policy can be used in the target domain.

\subsection{Dynamics Consistency}

To match the joint distribution \(P(z_t^\text{tgt}, a_t, z_{t+1}^\text{tgt})\) with those on the source domain, we introduce a dynamics consistency loss which is the sum of the \(\ell^2\) error in the forward and inverse dynamics predictions
\begin{equation} \label{eqn:dynamics_consistency_loss}
    \begin{split}
        \mathcal{L}_\textrm{dyn}(z_t, z_{t+1}, a_t) &= \lVert C_{\text{fwd}}(z_t, a_t) - z_{t+1} \rVert^2\\%
        &+ \lVert C_{\text{inv}}(z_t, z_{t+1}) - a_t \rVert^2.
    \end{split}
\end{equation}
\(C_\text{fwd}(z_{t+1} \vert  z_t, a_t)\) is the forward kinematics model that predicts the next latent $z_{t+1}$ given the previous latent $z_t$ and $a_t$. \(C_\text{inv}(a_t \vert z_t, z_{t+1})\) is the inverse model that predicts the action $a_t$ associated with the transition from $z_t$ to $z_{t+1}$. We found it was not necessary to scale the two terms separately as it worked well enough.

To update the encoder \(F\) we sample transitions using a random policy \(\bar \pi\) from the {\color{blue!60}target domain}. We freeze the parameters of the two dynamics model when updating \(F\).
We found that a learning rate of \(10^{-6}\) worked sufficiently well for both, and we did not find it necessary to scale the two loss terms separately.

\subsection{Adversarial Loss}
In addition to the dynamics consistency loss, we also introduce an adversarial learning objective (see Figure~\ref{fig:main-fig}), where a discriminator $D$ tries to distinguish between embeddings from the source domain $\srceq{z^{\src}_t}$ and those from the target domain $\tgteq{z^{\tgt}_t}$. We update the parameters of the encoder such that latent embeddings on the target domain are indistinguishable from those of the source domain.
Using the earth-moving metric from~\citet{arjovsky2017wasserstein}, we express this distribution-matching objective as

\begin{equation}\label{eqn:adversarial_loss}
        \mathcal{J}_\textrm{adv} = \mathbb{E}_{z_t^\src\sim \mathcal B_\csrc} \bigl[ D\left(\srceq{z_t^\src})\right) \bigr]
      + \mathbb{E}_{P_{\bar{\pi}}(\mathcal{M}_\ctgt) }\bigl[1 - D\left(F(\tgteq{o_t^\tgt})\right) \bigr].
\end{equation}

The encoder tries to minimize this objective while the discriminator acts as an adversary and seeks to maximize it, resulting in a GAN-like minimax game.

\subsection{Putting things together}
We adapt our encoder by minimizing a loss that combines both the adversarial loss $\mathcal{J}_\textrm{adv}$ (Eqn.~\ref{eqn:adversarial_loss}) and the dynamics consistency loss $\mathcal{J}_\textrm{dyn}$ (Eqn.~\ref{eqn:dynamics_consistency_loss}). Specifically, we solve for the parameters of the encoder through the following objective
\begin{equation}\label{eqn:joint_objective}
    \min_{F}\left[\max_D \mathcal{J}_\textrm{adv} + \mathcal{J}_\textrm{dyn}\right].
\end{equation}
We did not find it necessary to add additional scaling factors to balance the loss terms. %
We train the dynamics consistency networks to convergence before running the adaptation loss, to improve the wall-time.

The proposed approach, invariance through latent alignment, does not affect the training procedure of the control agent in the source domain, so it can be applied to any pretrained agent. We note that this is an unsupervised adaptation procedure since it does not require reward supervision at test time, nor does it require paired source and target images.

\section{Experiments}

We want to understand the impact of test-time adaptation on an agent's ability to generalize out-of-distribution. This section will compare ILA with two state-of-the-art RL baselines that use data-augmentation: SVEA~\citep{svea} and DrQ-v2~\citep{yarats2021drqv2}.

Recall from the introduction that these methods vie for increased generalization capabilities by expanding the support of the training distribution. We expect the invariance produced this way to be less performant than unsupervised adaptation at test time that only needs to focus on one specific instance of perceptual variation. In our experiments, we will also compare against \textit{policy adaptation during deployment} (PAD; see~\citet{hansen2020pad}), a baseline that, like our method, adapts the policy without access to the reward at test time. 

\begin{figure*}[t]
    \begin{minipage}{0.34\textwidth}
    \includegraphics[width=\linewidth]{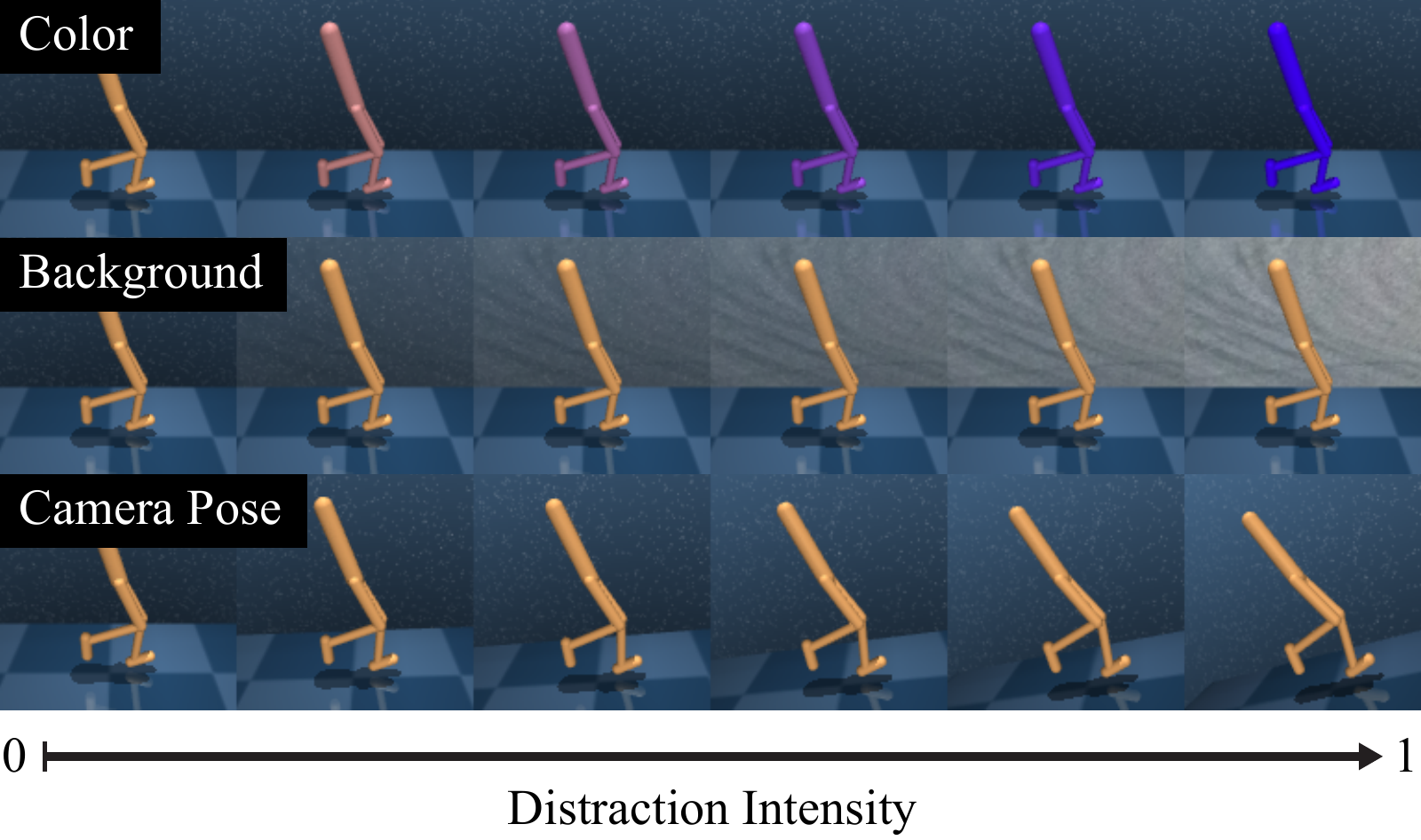}
    \caption{Samples from the modified \textit{distracting control suite}. Top row: color variations, middle row: background distractions, bottom row: camera pose variations.}
    \label{fig:distractions}
    \end{minipage}\hfill%
    \begin{minipage}{0.63\textwidth}
  \begin{subfigure}[t]{0.3\linewidth}
    \centering
    \includegraphics[height=80px]{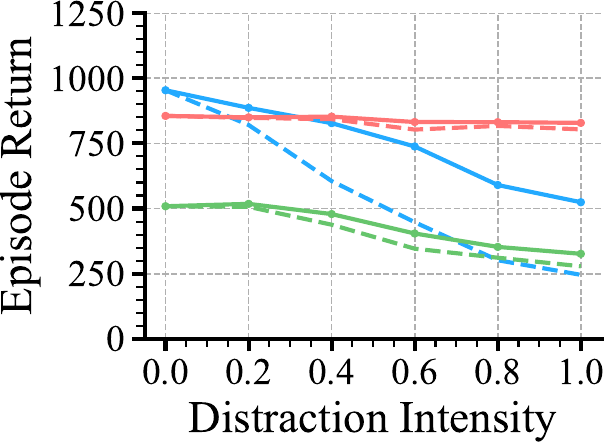}
    \caption{Color}
    \label{fig:intensity-colors}
  \end{subfigure}\hfill
  \begin{subfigure}[t]{0.3\linewidth}
    \centering
    \includegraphics[height=80px]{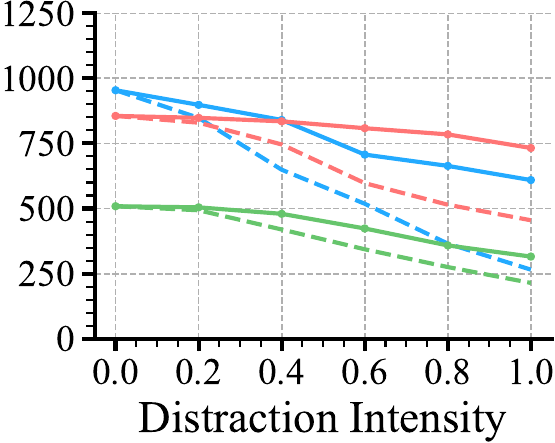}
    \caption{Background}
    \label{fig:intensity-background}
  \end{subfigure}\hfill
  \begin{subfigure}[t]{0.3\linewidth}
    \centering
    \includegraphics[height=80px]{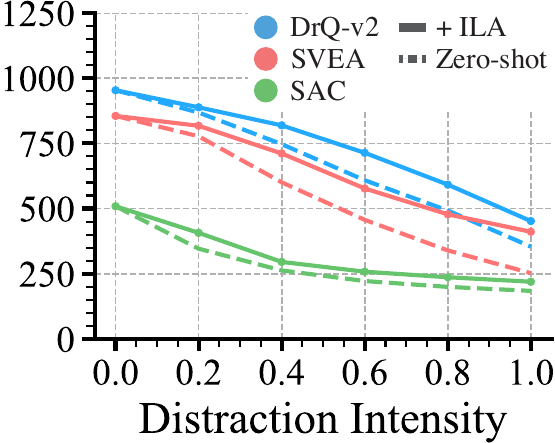}
    \caption{Camera Pose}
    \label{fig:intensity-camera}
  \end{subfigure}
    \vspace{0.5em}
    \caption{The gain of applying ILA to target domains with various distraction intensities. Dashed lines denote the performance of the baseline agent in the target environment (i.e., zero-shot transfer), while solid lines represent the performance gains of the agents with ILA.}%
    \label{fig:intensity-vs-reward}
  \end{minipage}
\end{figure*}

To further probe the generalization abilities of test-time adaptation, we conduct an experiment where we vary the intensity of environmental distractions, as illustrated in Fig. \ref{fig:distractions}. The results show that test-time adaptation significantly increases the policy performance during deployment. We will conclude with some general discussion and remarks regarding the design tradeoffs involved in test-time adaptation.

\paragraph{Setup} We conduct experiments on nine domains from the DeepMind Control Suite (DMC; see~\citet{deepmindcontrolsuite2018}) and treat it as the \emph{source} domain for training the RL agents. We use the Distracting Control Suite ~\citep{stone2021distracting} as the \emph{target} domain. Distracting control suite adds three types of distractions to DMC, including image background, random color texture, and changes to the camera pose. The intensity of these modes of distraction are calibrated. For details, refer to the accompanying report (see~\citet{stone2021distracting}) %

\paragraph{Modifications to Distracting Control Suite}
The default configuration of distracting control suite changes distractions at the start of every episode (e.g., different background images are used at every episode). However, we are interested in measuring an agent's ability to perform adaptation across several episodes on the \emph{specific} target environment. Thus, we modify distracting control suite to sample a distraction once in the beginning of learning, and then use the same distraction across all learning epochs. This also ensures consistent evaluation across algorithms. In accordance with this change, we also modify the intensity benchmark from distracting control suite. In our experiments, intensity measures the deviation of the target environment distraction from the source environment. For example, intensity may measure how far the distracting color is from the default. Finally, we modify the environments to only apply a single distraction during testing (rather than all three) in order to better understand the impact of each type of distraction on overall performance. Figure~\ref{fig:distractions} shows an example of distractions across intensities on Walker-walk domain.

\input{ila/table1}
\subsection{ILA on DeepMind Control Suite}
This section studies the impact of test-time adaptation on the DeepMind control suite. We begin by pretraining soft actor-critic (SAC)~\citep{sacapps}, SVEA, and DrQ-v2 in a non-distracting training-time environment. After training, we evaluate the learned policies on test environments with distractions of various intensities. This evaluation is zero-shot, i.e., there is no additional training in the test environment. 

Table~\ref{table:baseline-adaptation} presents the results for the different distracting control suite domains in the presence of background distractions with an intensity level of $1.0$. Specifically, we compare the test-time performance of SAC, SVEA, and DrQ-v2 in each domain with the episode rewards that we achieve when using {ILA} to adapt the encoder. The baseline algorithms employ image augmentation, which provides some robustness to variations at test time. Even then, however, we find that {ILA} improves the test-time generalization of all three baseline policies in most domains, often resulting in significant performance gains. In cases where {ILA} does not improve performance, the resulting reward is comparable to the baseline policy, i.e., {ILA} does not result in a performance degradation.

Figure~\ref{fig:intensity-vs-reward} visualizes the performance of the different methods, averaged over the set of distracting control suite domains, as a function of the intensity of the distractions. Since the baseline methods are trained with image augmentation, they do exhibit some robustness to distraction. However, we see this robustness rapidly diminishes as the distraction intensity increases. In particular, large changes to camera pose or the image background proved challenging for standard augmentation procedures. Comparatively, {ILA} makes the degradation of performance much slower. This supports our hypothesis that adaptation powered by unsupervised learning can significantly widen the generalization abilities of learning algorithms.

\subsection{Comparisons with PAD}
\input{ila/table-pad-comparison}
Similar to our approach, PAD pretrains the agent in a clean environment, and then adapts the agent via unsupervised objectives without assuming access to the target environment's reward function~\citep{hansen2020pad}. To evaluate the robustness of PAD to distractions, we consider distracting control suite with a fixed distraction intensity of $1.0$. Table~\ref{table:pad} compares the performance as the difference between the episode returns before and after adaptation along with the episode returns in the clean environment. It should be noted that PAD requires the policy to be trained along with an inverse dynamics prediction objective, whereas ILA does not. We include this additional auxiliary objective with soft actor-critic specifically for this experiment.

Across all environments, we see that PAD struggles to adapt to distractions at test time. We suspect this instability is caused by the large deviations in the latent variable distribution as a result of changes in the target environment. In particular, we posit that the signal from PAD's inverse dynamics predictor does not encourage the latent train and test distributions to match.%

\subsection{Sim-to-real Transfer}
\input{ila/fig-sim2real}
We are interested in the ability of {ILA} to bridge the gap between simulated and real world robotics environments. In the {\it reaching} task, the target position is given by a red disc placed on a table. The agent's objective is to control the arm so that the end-effector reaches this target location. Our goal is to train a policy in simulation, and then transfer the policy to a real UR-5 robot at test time. The same as with previous experiments, the test time agent receives no rewards. In both simulation and the real world, the policy's only input is an image from a camera placed in front of the robot and table. The action space is a 2D position controller that drives a small movement $(\delta x, \delta y)$ of the robotic gripper. See Figure~\ref{fig:sim2real} for the setup.

We carry out the experiment by first training a policy with SVEA~\citep{svea} in simulation. For adaptation, we collect random trajectories in the real environment and then use {ILA} to align the simulated and real world experiences. We evaluated the success rate of zero-shot and the adapted policies over \(20\) real episodes. We consider the episode to be success if the gripper's tip overlaps with the target in the front-view image.

Due to the challenging domain shift between simulator and real world, the zero-shot policy fails to adapt adequately, repeating the same action of moving the gripper to an edge of the table ad infinitum, regardless of the given goal location. This results in a final success rate of \(15\%\). %
On the same task, {ILA} is able to robustly against this domain shift, achieving a final success rate around \(90\%\).

\subsection{Further Discussion}
\input{ila/table-ablation}
\paragraph{Ablation Studies} In order to better understand the contribution of the different objectives to test-time generalization, we perform a series of ablations in which we omit either the dynamics consistency or the adversarial objectives.

In these experiments, we use a pretrained DrQ-v2 network for the algorithm's base policy, and then perform adaptation across all distractions with an intensity value of $1.0$. 
The results in Table \ref{table:ablation} show that the adversarial training is critical to adapt the latent representation in the target domain. Performing adaptation using only the dynamics consistency objective, i.e., $\argmin_F$ $\mathcal{J}_\textrm{dyn}$ (Eqn.~\ref{eqn:dynamics_consistency_loss}) results in a significant decrease in performance.
We theorize that the dynamics consistency objective helps to align latent transition manifolds only when the latent distributions in source and target domains are reasonably close. 
If the latent distributions significantly differ, however, the input to the pretrained dynamics networks is largely out-of-distribution, and thus the gradients from dynamics consistency loss may negatively affect convergence.

Compared to the adversarial objective, ablating the dynamics consistency objective
has surprisingly little effect on test-time generalization. 
It may be that the transition manifold in latent spaces
are preserved despite the distractions, which then diminishes the net effect of the dynamics consistency objective. 

\paragraph{Pre-Filling the Replay Buffer}
We implicitly make the assumption that a behavior policy $\bar{\pi}$ is available that can be used to generate trajectory data on both the source and the target domain with similar state visitation, and transition probabilities. To achieve good performance with the adapted policy, such distribution should also cover important states with higher reward. To our surprise, a simple scheme where we pre-fill both the source and target buffer using the uniformly random action policy works sufficiently well.

\section{Conclusion}
We introduced \emph{\lcila}, an unsupervised approach that matches the distribution of feature vectors in the latent space, to improve the test-time performance of a learned visuomotor control policy. Empirical results show that as discrepancies between the training and deployment environments become more intense, \lcila~has a large competitive edge over alternatives such as data augmentation techniques. The problem of test-time adaptation in visual reinforcement learning using unsupervised test-time trajectories is relatively new, but has thus far shown great relevance and promise in robotics~\citep{hansen2020pad,hansen2021soda}, where a sim-to-real pipeline has been at the fore-front of recent progress~\citep{Kaufmann2018drone,Wu2019deformable,margonlis2022rapid}.

Finally, although the adversarial distribution matching objective proposed here produce measurable improvements, better ways to align latent features are still needed. For a recent work in this direction, we refer the reader to \textit{adversarial support alignment}~\citep{tong2022support}.

%% file: ila/graphical_model.tex
\begin{figure}[!t]
\centering
\begin{subfigure}[t]{0.48\textwidth}
    \centering
    \scalebox{0.7}{%
    \usetikzlibrary{shapes.misc}
    \begin{tikzpicture}[
      node distance=2em, auto,
      lat/.style={draw=black, circle, minimum size=2.5em},
      det/.style={draw=black, rectangle, minimum size=2.5em},
      obs/.style={circle, draw=black, fill=black!20, minimum size=2.5em},
      gen/.style={->, -{Stealth[length=.5em, inset=0pt]}},
      inf/.style={dashed, ->, -{Stealth[length=.5em, inset=0pt]}},
    ]
    \node[obs, inner sep=.02em] (r1) {$r_1$};
    \node[obs, right=of r1, inner sep=.02em] (r2) {$r_2$};
    \node[obs, above=of r1] (o1) {$o^\textrm{src}_1$};
    \node[obs, above=of r2] (o2) {$o^\textrm{src}_2$};
    \node[lat, above=of o1] (x1) {$s_1$};
    \node[lat, above=of o2] (x2) {$s_2$};
    \node[right=of x2] (dots) {$\dots$};
    \path (x2) edge[gen] node {} (dots);
    \node at (x1) [lat, shift=(-3.5em :-3.5em)] (z1) {$z^\textrm{src}_1$};
    \path (o1) edge[red!80, gen, bend left=50, line width=2, dashed] node{} (z1);
    \node[obs, above=of x1] (a1) {$a_1$};
    \path (z1) edge[darkgreen, gen, line width=2] node{} (a1);
    \path (x1) edge[gen] node {} (z1);
    \path (x1) edge[gen] node {} (o1);
    \path (x1) edge[gen, bend right] node {} (r1);
    \path (x2) edge[gen] node {} (o2);
    \path (x2) edge[gen, bend right] node {} (r2);
    \path (x1) edge[gen] node {} (x2);
    \path (a1) edge[gen] node {} (x2);
    \end{tikzpicture}}
    \caption{Training on $\mathcal{M}_\textrm{src}$}
    \label{fig:source_mdp}
\end{subfigure}%
\begin{subfigure}[t]{0.48\textwidth}
    \centering
    \scalebox{0.7}{%
    \begin{tikzpicture}[
      node distance=2em, auto,
      lat/.style={draw=black, circle, minimum size=2.5em},
      det/.style={draw=black, rectangle, minimum size=2.5em},
      obs/.style={circle, draw=black, fill=black!20, minimum size=2.5em},
      gen/.style={->, -{Stealth[length=.5em, inset=0pt]}},
      inf/.style={dashed, ->, -{Stealth[length=.5em, inset=0pt]}},
    ]
    \node[lat, inner sep=.02em] (r1) {$r_1$};
    \node[lat, right=of r1, inner sep=.02em] (r2) {$r_2$};
    \node[obs, above=of r1] (to1) {$o_1^\textrm{tgt}$};
    \node[obs, right=of to1, above=of r2] (to2) {$o^\textrm{tgt}_2$};
    \node[lat, above=of to1] (s1) {$s_1$};
    \node[lat, above=of to2] (s2) {$s_2$};
    \path (s1) edge[gen, bend right] node {} (r1);
    \path (s2) edge[gen, bend right] node {} (r2);
    \node[obs, above=of s1] (a1) {$a_1$};
    \node[right=of s2] (dots) {$\dots$};
    \path (s2) edge[gen] node {} (dots);
    \node at (s1) [lat, shift=(-3.5em :-3.5em)] (z1) {$z^\textrm{tgt}_1$};
    \path (to1) edge[blue!50, gen, bend left=50, line width=2, dashed] node{} (z1);
    \path (s1) edge[gen] node {} (z1);
    \path (a1) edge[gen] node {} (s2);
    \path (s1) edge[gen] node {} (to1);
    \path (s1) edge[gen] node {} (s2);
    \path (s2) edge[gen] node {} (to2);
    \path (z1) edge[darkgreen, gen, line width=2] node{} (a1);
    \end{tikzpicture}}
    \caption{Adaptation on $\mathcal{M}_\textrm{tgt}$}
    \label{fig:target_mdp}
\end{subfigure}
\caption{
A Markov decision process gives rise to two different observations $o^\textrm{src}_t$ and $o^\textrm{tgt}_t$. $o^\textrm{src}_t$ is accessible during training, whereas $o^\textrm{tgt}_t$ is only accessible during deployment.
The {\color{red!80}red} dashed arrow indicates the learned inference network $F$, produced as part of the policy during training. During deployment, the latent features produced by $F$, $z^\textrm{tgt}_t$ ({\color{blue!50}blue} dashed arrow), experience a domain-shift. The reward further becomes unobservable. Our goal is to make the representation domain-invariant, such that the policy $\pi(a|o^\textrm{tgt})$ (frozen policy head in {\color{darkgreen}green}) can succeed in the target domain.}
\label{fig:graphicalmodels}
\end{figure}

%% file: ila/table1.tex
\todo{Fix Table 1 (centering)}
\begin{table*}[t!]
\centering
\caption{Episode return in the target (test) environments (mean and standard deviation) before (zero-shot) and after (+ILA) adaptation for SAC, SVEA, and DrQ-v2 with background distraction at an intensity setting of $1.0$. The performance of each baseline in the source (training) environments can be found in the Appendix.}\label{table:baseline-adaptation}{\small%
\setlength{\tabcolsep}{6pt}
\begin{tabularx}{\linewidth}{Xcc|cc|cc}
    \toprule
                       & \multicolumn{2}{c}{SAC}                                          & \multicolumn{2}{c}{SVEA}                                                       & \multicolumn{2}{c}{DrQ-v2}\\
    Domain              & Zero-shot                 & +ILA                                  & Zero-shot                             & +ILA                                  & Zero-shot                     & +ILA \\
    \midrule
\texttt{ball\_in\_cup-catch} & $115^{\scriptscriptstyle\pm 50} \hphantom{0}$  & $\bm{227^{\scriptscriptstyle\pm 222}}$              & $490^{\scriptscriptstyle\pm 376}$                          & $\bm{987^{\scriptscriptstyle\pm 27}} \hphantom{0}$    & $\hphantom{0}88^{\scriptscriptstyle\pm 39} \hphantom{0}$     & $\bm{386^{\scriptscriptstyle\pm 425}}$    \\
\texttt{cartpole-balance}    & $434^{\scriptscriptstyle\pm 275}$             & $\bm{585^{\scriptscriptstyle\pm 295}}$              & $446^{\scriptscriptstyle\pm 330}$                          & $\bm{627^{\scriptscriptstyle\pm 258}}$               & $273^{\scriptscriptstyle\pm 107}$                           & $\bm{322^{\scriptscriptstyle\pm 117}}$    \\
\texttt{cartpole-swingup}    & $182^{\scriptscriptstyle\pm 147}$             & $\bm{369^{\scriptscriptstyle\pm 243}}$              & $269^{\scriptscriptstyle\pm 365}$                          & $\bm{612^{\scriptscriptstyle\pm 213}}$               & $\hphantom{0}82^{\scriptscriptstyle\pm 35}\hphantom{0}$     & $\bm{247^{\scriptscriptstyle\pm 136}}$    \\
\texttt{cheetah-run}         & $169^{\scriptscriptstyle\pm 65} \hphantom{0}$  & $\bm{248^{\scriptscriptstyle\pm 53}}\hphantom{0}$   & $317^{\scriptscriptstyle\pm 137}$                          &  $\bm{378^{\scriptscriptstyle\pm 55}}\hphantom{0}$    & $100^{\scriptscriptstyle\pm 88}\hphantom{0}$                & $\bm{393^{\scriptscriptstyle\pm 125}}$    \\
\texttt{finger-spin}         & $113^{\scriptscriptstyle\pm 162}$             & $\bm{192^{\scriptscriptstyle\pm 196}}$              & $391^{\scriptscriptstyle\pm 467}$                          & $\bm{943^{\scriptscriptstyle\pm 54}}\hphantom{0}$    & $207^{\scriptscriptstyle\pm 328}$                           & $\bm{769^{\scriptscriptstyle\pm 206}}$    \\
\texttt{finger-turn\_easy}   & $\bm{163^{\scriptscriptstyle\pm 99}}\hphantom{0}$  & $146^{\scriptscriptstyle\pm 33} \hphantom{0}$   & $278^{\scriptscriptstyle\pm 180}$                          & $\bm{491^{\scriptscriptstyle\pm 343}}$  &  $268^{\scriptscriptstyle\pm 241}$   & $\bm{914^{\scriptscriptstyle\pm 44}}\hphantom{0}$  \\
\texttt{reacher-easy}        & $179^{\scriptscriptstyle\pm 65}\hphantom{0}$  & $\bm{381^{\scriptscriptstyle\pm 76}}\hphantom{0}$   & $\hphantom{0}75^{\scriptscriptstyle\pm 77} \hphantom{0}$    & $\bm{624^{\scriptscriptstyle\pm 305}}$               & $\hphantom{0}58^{\scriptscriptstyle\pm 32}\hphantom{0}$     & $\bm{685^{\scriptscriptstyle\pm 211}}$    \\
\texttt{walker-stand}        & $330^{\scriptscriptstyle\pm 118}$             & $\bm{364^{\scriptscriptstyle\pm 115}}$              & $917^{\scriptscriptstyle\pm 138}$                          & $\bm{999^{\scriptscriptstyle\pm 12}} \hphantom{0}$    & $630^{\scriptscriptstyle\pm 197}$                           & $\bm{868^{\scriptscriptstyle\pm 151}}$    \\
\texttt{walker-walk}         & $242^{\scriptscriptstyle\pm 142}$             & $\bm{291^{\scriptscriptstyle\pm 134}}$              & $866^{\scriptscriptstyle\pm 45}\hphantom{0}$               & $\bm{924^{\scriptscriptstyle\pm 45}}\hphantom{0}$    & $326^{\scriptscriptstyle\pm 195}$                           & $\bm{770^{\scriptscriptstyle\pm 140}}$ \\  
    \bottomrule
\end{tabularx}}
\end{table*}

%% file: ila/table-pad-comparison.tex
\begin{table}[t]
    \centering
    \caption{Comparison with PAD}
    \begin{tabularx}{0.8\linewidth}{Xcccc}
    \toprule
    Distraction & Zero-shot & +PAD & +ILA\\
    \midrule
    
    None          & $835^{\scriptscriptstyle\pm 230} $ & ---  & --- \\ 
    Background    & $213^{\scriptscriptstyle\pm 247} $ & $ 279^{\scriptscriptstyle\pm 271}$ & $ \bm{425^{\scriptscriptstyle\pm 292}}$ \\
    Colors        & $230^{\scriptscriptstyle\pm 263}$ & $ 271^{\scriptscriptstyle\pm 300}$ & $ \bm{402^{\scriptscriptstyle\pm 339}}$ \\
    Camera Pose   & $319^{\scriptscriptstyle\pm 265}$ & $ 326^{\scriptscriptstyle\pm 259}$ & $ \bm{412^{\scriptscriptstyle\pm 275}}$ \\
    \bottomrule
    \end{tabularx}
    \label{table:pad}
\end{table}

%% file: ila/fig-sim2real.tex
\begin{figure}[!t]
    \centering
    \begin{subfigure}[t]{0.34\linewidth}
    \includegraphics[width=\linewidth]{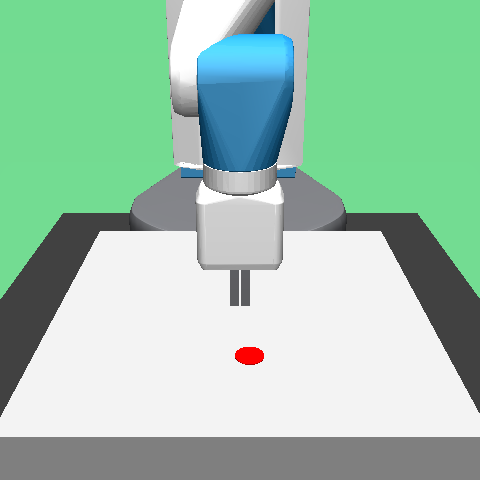}
    \end{subfigure}
    \hspace{2em}
    \begin{subfigure}[t]{0.34\linewidth}
    \includegraphics[width=\linewidth]{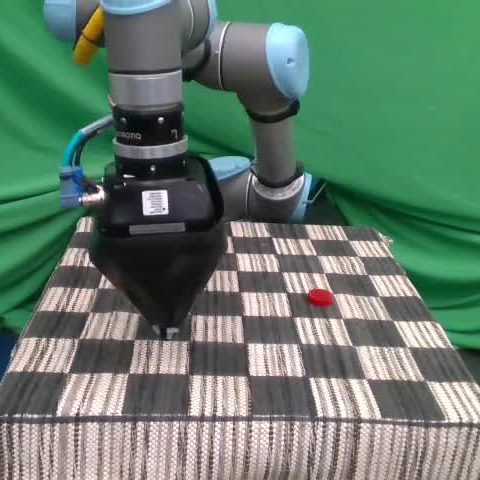}
    \end{subfigure}\hfill
    \caption{Simulated and real reach environment. The goal location is denoted as a red disk that the robot must reach. Using ILA, we can transfer a policy trained in simulation (left) onto a real UR-5 robot (right). This adaptation requires no paired data and no rewards on the deployment environment, instead employing ILA for unpaired adaptation.}\label{fig:sim2real}
\end{figure}

%% file: ila/table-ablation.tex
\begin{table}[t]
    \centering
    \caption{Ablations with variants of ILA that remove inverse/forward dynamics, or the adversarial objectives. DrQ-v2 is used as a pretrained policy. We compute episodic returns from nine domains and five random seeds, and the results correspond to an intensity value of $1$.} \label{table:ablation}
    {\small \begin{tabularx}{\linewidth}{Xccccc}
    \toprule
                 &           &      & +ILA & +ILA \\
    Distraction  & Zero-shot & +ILA & w/o dyn. & w/o adv. \\
    \midrule
    Background & $228 \pm 232$ & $602 \pm 300$ & $615 \pm 289$ & $176  \pm 221$ \\
    Colors  & $234 \pm 245$ & $536 \pm 320$ & $534 \pm 327$ & $117  \pm 96\hphantom{0}$ \\
    Camera Pose & $345 \pm 287$ & $417 \pm 284$ & $407 \pm 272$ & $208 \pm 235$ \\
    \bottomrule
    \end{tabularx}}
\end{table}